%% file: main.tex
\documentclass[sigconf]{acmart}
\AtBeginDocument{%
  \providecommand\BibTeX{{%
    \normalfont B\kern-0.5em{\scshape i\kern-0.25em b}\kern-0.8em\TeX}}}

\copyrightyear{2024}
\acmYear{2024}
\setcopyright{acmlicensed}\acmConference[KDD '24]{Proceedings of the 30th ACM SIGKDD Conference on Knowledge Discovery and Data Mining}{August 25--29, 2024}{Barcelona, Spain}
\acmBooktitle{Proceedings of the 30th ACM SIGKDD Conference on Knowledge Discovery and Data Mining (KDD '24), August 25--29, 2024, Barcelona, Spain}
\acmDOI{10.1145/3637528.3671928}
\acmISBN{979-8-4007-0490-1/24/08}

\newtheorem{myDef}{\textbf{Definition}}

\newtheorem{myPro}{\textbf{Problem}}
\newtheorem{lemma}{\textbf{Lemma}}

\input{math_commands.tex}

\usepackage{booktabs}
\usepackage{tabularx}
\usepackage{multirow}
\usepackage{verbatim}
\usepackage{wrapfig}
\usepackage{soul}
\usepackage{caption}
\usepackage{array}
\usepackage{xcolor} 
\usepackage{threeparttable}
\usepackage{subcaption}
\usepackage{tabularx}
\usepackage{adjustbox}
\usepackage{boldline}
\usepackage{algorithm}
\usepackage{xspace}
\usepackage{algpseudocode}
\usepackage{listings}
\usepackage{multicol} 
\definecolor{Pine}{RGB}{0,139,114}
\definecolor{Brick}{RGB}{182,50,28}
\definecolor{Cerulean}{RGB}{0,162,227}
\definecolor{dodgerblue}{RGB}{30,144,255}
\usepackage{hyperref}
\usepackage{changepage}

\newcommand{\boldres}[1]{{\textbf{#1}}}
\newcommand{\secondres}[1]{\underline{#1}}
\newcommand{\Kc}{\mathcal{K}} 

\newcommand{\methodn}{\texttt{Fredformer}}
\newcommand{\method}{\textsc{\methodn}\xspace}

\begin{document}

\title{
Fredformer: Frequency Debiased Transformer for Time Series Forecasting
}

\author{Xihao Piao*}
\affiliation{%
  \institution{SANKEN, Osaka University}
  \city{Osaka}
  \country{Japan}
}
\email{park88@sanken.osaka-u.ac.jp}

\author{Zheng Chen*}
\affiliation{%
  \institution{SANKEN, Osaka University}
  \city{Osaka}
  \country{Japan}
}
\email{chenz@sanken.osaka-u.ac.jp}

\author{Taichi Murayama}
\affiliation{%
  \institution{SANKEN, Osaka University}
  \city{Osaka}
  \country{Japan}
}
\email{taichi@sanken.osaka-u.ac.jp}

\author{Yasuko Matsubara}
\affiliation{%
   \institution{SANKEN, Osaka University}
   \city{Osaka}
   \country{Japan}
 }
\email{yasuko@sanken.osaka-u.ac.jp}

\author{Yasushi Sakurai}
\affiliation{%
    \institution{SANKEN, Osaka University}
    \city{Osaka}
    \country{Japan}
}
\email{yasushi@sanken.osaka-u.ac.jp}

\thanks{* Indicates corresponding authors} 

\begin{abstract}
The Transformer model has shown leading performance in time series forecasting.
Nevertheless, in some complex scenarios,
it tends to learn low-frequency features in the data and overlook high-frequency features, showing a frequency bias.
This bias prevents the model from accurately capturing important high-frequency data features.
In this paper, we undertake empirical analyses to understand this bias and discover that frequency bias results from the model disproportionately focusing on frequency features with higher energy.
Based on our analysis, we formulate this bias and propose \method, a Transformer-based framework designed to mitigate frequency bias by learning features equally across different frequency bands. 
This approach prevents the model from overlooking lower amplitude features important for accurate forecasting.
Extensive experiments show the effectiveness of our proposed approach, which can outperform other baselines in different real-world time-series datasets. 
Furthermore, we introduce a lightweight variant of the \method with an attention matrix approximation, which achieves comparable performance but with much fewer parameters and lower computation costs.
The code is available at: \url{https://github.com/chenzRG/Fredformer}

\end{abstract}

\ccsdesc[500]{Computing methodologies~Artificial intelligence}
\ccsdesc[500]{Computing methodologies~Neural networks}

\keywords{ Time series forecasting, Deep learning}

\maketitle

\section{Introduction}\label{sec: 1}

Time series data are ubiquitous in everyday life.  
Forecasting time series could provide insights for decision-making support, such as potential traffic congestion \citep{app_elec} or changes in stock market trends \citep{financialsupoort}. 
Accurate forecasting typically involves discerning various informative temporal variations in historical observations, e.g., trends, seasonality, and fluctuations, which are consistent in future time series \citep{Timesnet}.
Benefiting from the advancements in deep learning, the community has seen great progress, particularly with Transformer-based methods \citep{Informer, Etsformer, IJCAI2023Transformersurvey}.
Successful methods often \textit{tokenize time series} with multiresolution, such as time points \citep{Autoformer} or sub-series \citep{Crossformer}, and \textit{model their dependencies leveraging the self-attention mechanism}.
Several state-of-the-art (SOTA) baselines have been proposed, namely PatchTST \citep{PatchTST}, Crossformer \citep{Crossformer}, and iTransformer \citep{liu2023itransformer}, and demonstrate impressive performance.

\begin{figure*}
\centering
\includegraphics[width=1\textwidth]{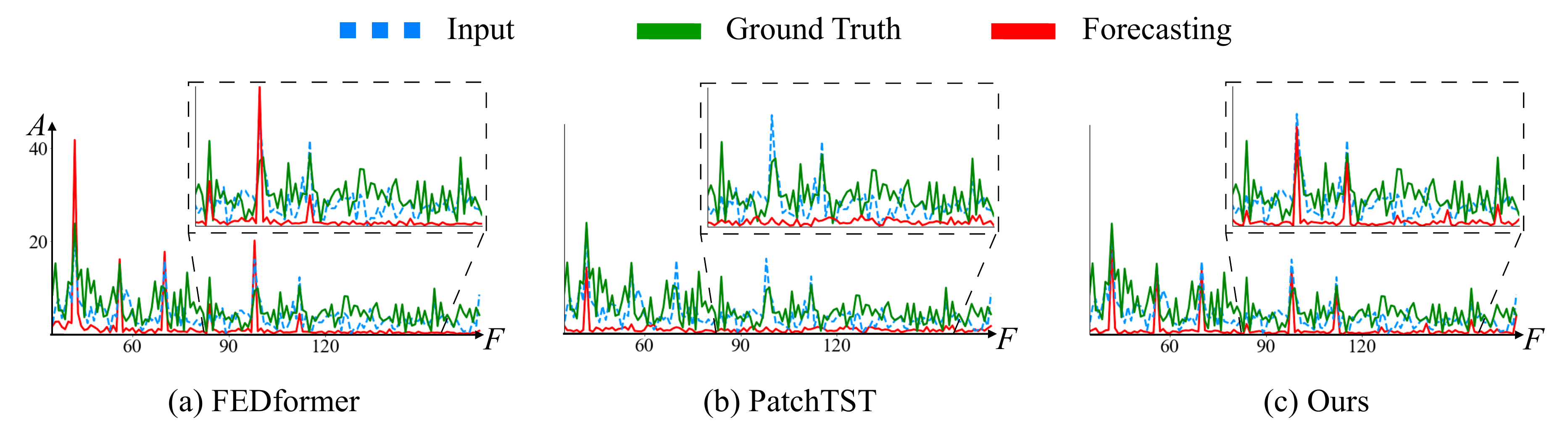}
\caption{
In contrast to a frequency modeling-based work FEDformer \cite{ICMLFedformer} and a SOTA work PatchTST \cite{PatchTST}, our model can accurately capture more significant mid-to-high frequency components.}
\vspace{0.5cm}
\label{fig: First_pic}
\end{figure*}

Despite their success, the effectiveness with which we can capture informative temporal variations remains a concern.
From a data perspective, a series of time observations is typically considered a complex set of signals or waves that varies over time \cite{time_frequency_decomposition_1998, almost_harmonics_2020}. 
Various temporal variations, manifested as different frequency waves, such as low-frequency long-term periodicity or high-frequency fluctuation, often co-occur and are intermixed in the real world \citep{long_and_short_SiGIR18, logtrans2019, Timesnet}. 
While \textit{tokenizing a time series} may provide fine-grained information for the model, the temporal variations in resulting tokens or sub-series are also entangled.
This issue may complicate the feature extraction and forecasting performance.
Existing works have proposed frequency decomposition to represent the time series and deployed Transformers on new representation to explicitly learn eventful frequency features \citep{Timesnet, Autoformer}.
Learning often incorporates feature selection strategies in the frequency domain, such as top-K or random-K \citep{Etsformer, ICMLFedformer}, to help Transformers better identify more relevant frequencies.
However, such heuristic selection may introduce sub-optimal frequency correlations into the model (seen in Figure \ref{fig: First_pic}(a)), inadvertently misleading the learning process. 

From a model perspective, researchers have recently noticed a \textit{learning bias issue that is common in the Transformer}.
That is, the self-attention mechanism often prioritizes low-frequency features at the expense of high-frequency features \citep{antioversmooth_2022_ICLR, howTransWork_2022_ICLR, NIPS2023scan, contranorm_2023_ICLR}.
This subtle issue may also appear in time series forecasting, potentially biasing model outcomes and leading to information losses.
Figure \ref{fig: First_pic}(b) shows an electricity case where the forecasting result successfully captures low-frequency features, neglecting some consistent mid-to-high frequencies.
In practice, such high frequencies represent short-term variations, e.g., periodicities over short durations, which serve as good indicators for forecasting \citep{app_elec, financialsupoort, Waterresearch2023}.
However, the low-frequencies typically carry a substantial portion of the energy in the spectrum and are dominant in time series. 
The amplitude of these low-frequency components far exceeds that of higher frequencies \citep{VLDB2002}, which provides the Transformer with more observations.
This may raise the possibility of frequency bias in time series forecasting, as the model might disproportionately learn from these dominant low-frequency components.

This work explores one direction of capturing informative, complex variations by frequency domain modeling for accurate time series forecasting.
We introduce \method, a \textbf{Fre}quency-\textbf{d}ebiased Trans\textbf{former} model.
\method follows the line of frequency decomposition but further investigates how to facilitate the uses of Transformers in learning frequency features.
To improve the effectiveness of our approach, we provide a comprehensive analysis of frequency bias in time series forecasting and a strategy for debiasing it.
Our main contributions lie in three folds.

\noindent \textbf{- Problem definition.} We undertake empirical studies to investigate how this bias is introduced into time series forecasting Transformers.
We observe that the main cause is the proportional difference between key frequency components.
Notably, these key components should be consistent in the historical and ground truth of the forecasting. 
We also investigate the objective and key designs that affect debiasing.\\
\textbf{- Algorithmic design.}
Our \method has three pivotal components: patching for the frequency band, sub-frequency-independent normalization to mitigate proportional differences, and channel-wise attention within each sub-frequency band for fairness learning of all frequencies and attention debiasing.\\
\textbf{- Applicability.}
\method undertakes Nyström approximation to reduce the computational complexity of the attention maps, thus achieving a lightweight model with competitive performance. 
This opens new opportunities for efficient time series forecasting. 

\noindent \textbf{Remark.}
This is the first paper to study the frequency bias issue in time series forecasting. 
Extensive experimental results on eight datasets show the effectiveness of \method, which achieves superior performance with 60 top-1 and 20 top-2 cases out of 80.

\section{Preliminary Analysis}\label{sec: PreliminaryAnalysis}

We present two cases to show $\rm(\hspace{.18em}i\hspace{.18em})$ how frequency attributes of time series data introduce bias into forecasting with the Transformer model and $\rm(\hspace{.08em}ii\hspace{.08em})$ an empirical analysis of the potential debiasing strategy.
This section introduces the notation and a metric for the case studies in Sec. \ref{subsec:Preliminary}. 
The case analyses are detailed in Sec. \ref{subsec: CaseStudies}.

\subsection{Preliminary}\label{subsec:Preliminary}

\noindent \textbf{Time Series Forecasting.} 
Let $\mathbf{X}= {\{\vx_1^{(c)}, \ldots, \vx_L^{(c)}\}}_{c=1}^{C}$ denote a multivariate time series consisting of $C$ channels, where each channel records an independent $L$ length look-back window.
For simplicity, we omit channel index $c$ in subsequent discussions.
The forecasting task is to predict $H$ time steps in the future data $\hat{\mathbf{X}}$:
\[
\hat{\mathbf{X}}_{L+1:L+H} = f(\mathbf{X}_{1:L})
\]
\noindent where $f(\cdot)$ denotes the forecasting function, which is a Transformer-based model in this work.
Our objective is to mitigate the learning bias in the Transformer and enhance the forecasting outcome $\mathbf{X}'$, that is, to minimize the error between $\mathbf{X}'$ and $\hat{\mathbf{X}}$.

\begin{figure}[t]
\begin{center}
\includegraphics[width=1\linewidth]{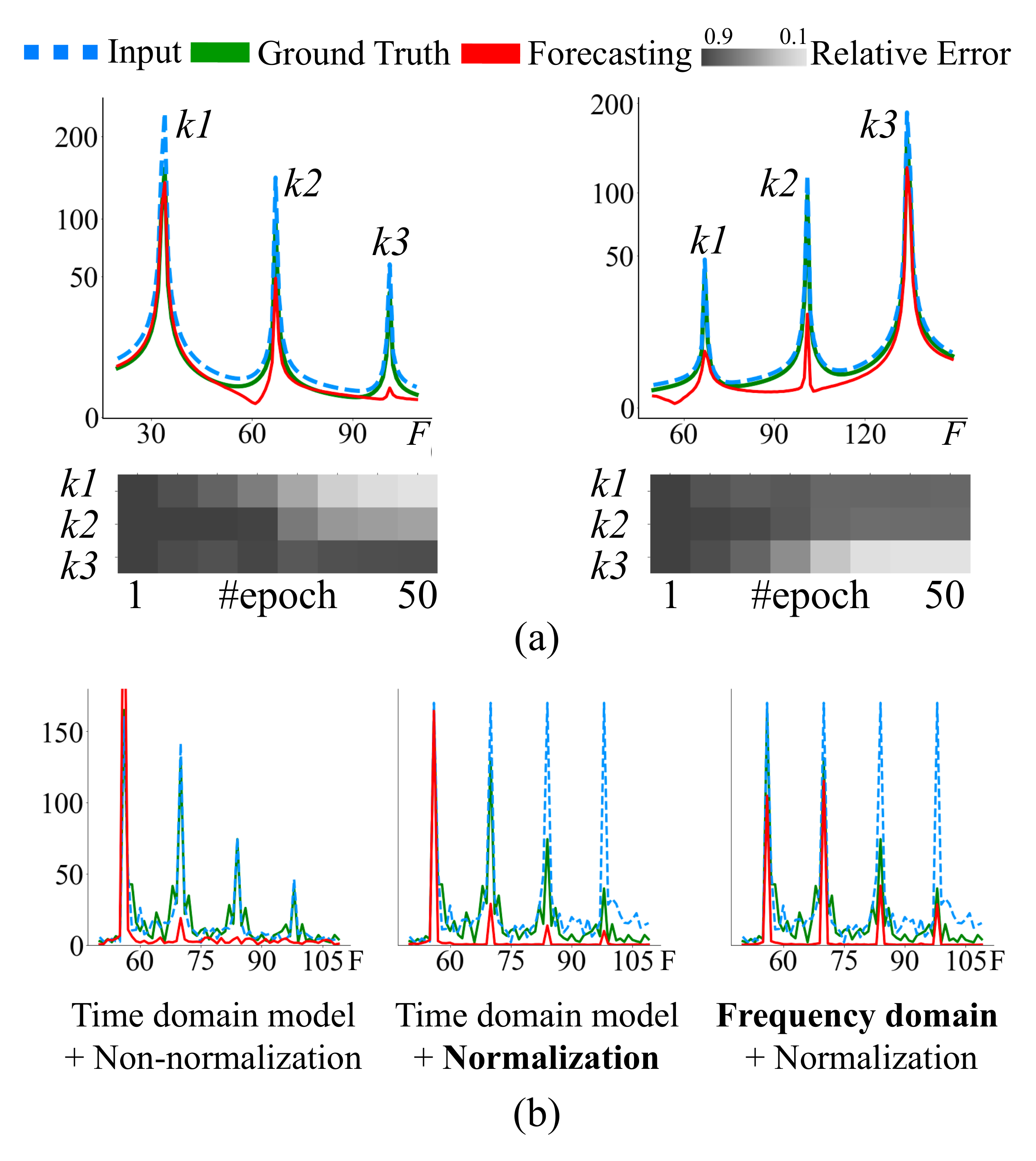}
\end{center}
\caption{
Figure (a) shows the learning dynamics and results for two synthetic datasets, employing line graphs to illustrate amplitudes in the frequency domain and heatmaps to represent training epoch errors. Figure (b) explores the influence of amplitude and domain on learning by comparing Transformers in the time and frequency domains, both with and without frequency local normalization.}
\label{fig: case1}
\end{figure}

\noindent \textbf{Discrete Fourier Transform (DFT).}
We use DFT to analyze the frequency content of $\mathbf{X}$, $\hat{\mathbf{X}}$, and $\mathbf{X}'$.
For example, given the input sequence $ \{\vx_1, ..., \vx_L\} $, the DFT can be formulated as
\[
\va_k = \frac{1}{L}\sum_{l=1}^{L}{\vx_l \cdot f_k}, \quad k = 1,\ldots,L
\]
where $f_k =e^{-i2\pi k/L}$ denotes the $k$-th frequency component.
The DFT coefficients $\mathbf{A} = \{\va_1,\va_2,\ldots,\va_L\}$ represent the amplitude information of these frequencies.
As illustrated in Figure \ref{fig: case1} (b, left), four components are observed to have higher amplitudes in the historical observations ($\mathbf{X}$) and the forecasting data ($\hat{\mathbf{X}}$).
We refer to such consistent components as '\textit{key components}' (defined in Sec. \ref{subsec: BiasDefinition}).
Here, the inverse DFT (i.e., IDFT) is $ \vx_l = \sum_{k=1}^{L} \va_k \cdot f_k^{-1}$, which reconstructs the time series data from the DFT coefficients.

\noindent \textbf{Frequency Bias Metric.}
Inspired by the work of \citep{relative_error, frequencyPrinciple_2020},
this study employs a Fourier analytic metric of relative error $\Delta_{k}$ to determine the frequency bias.
Given the model outputs $\mathbf{A}'$ and the ground truth $\hat{\mathbf{A}}$, the mean-square error (MSE) for the $k$-th component is calculated as follows:
$\text{MSE}_{k}=|{\va'}_k-\hat{\va}_k|$, where $|\cdot|$ denotes the L2 norm of a complex number.
Then, the relative error is applied to mitigate scale differences. In other words, the error may become larger as the proportion of amplitude increases.
\[
\Delta_{k}=|{\va'}_k-\hat{\va}_k|/|\hat{\va}_k|
\]

This metric is used in case study analyses and the experiments detailed in Section \ref{subsec: MainResults}.

\subsection{{Case Studies}}\label{subsec: CaseStudies}

We first generate single-channel time series data with a total length of $10000$ timestamps and then employ a Transformer model \cite{PatchTST} to forecast the data. The details are in Appendix \ref{app:case_study_details}.
For the first case study \textbf{(Case 1)}, we generate two datasets with three key frequency components ($\{k1, k2, k3\}$).
Each dataset contains a different proportion of these three components, as illustrated in the DFT visualization in Figure \ref{fig: case1}. 
On the left side of the figure, their amplitudes are arranged as $\va_{k1} < \va_{k2} < \va_{k3}$, whereas on the right side, the arrangement is $\va_{k1} > \va_{k2} > \va_{k3}$.
We maintain these proportions so that they are consistent between the observed {\color{Cerulean} $\mathbf{A}$} and the ground truth {\color{Pine} $\hat{\mathbf{A}}$} (i.e., {\color{Cerulean} $\mathbf{A}$}$ \approx$ {\color{Pine} $\hat{\mathbf{A}}$}).
Then, we assess the bias for different $k$ in the Transformer outputs {\color{Brick} $\mathbf{A}'$}.
Meanwhile, we track how $\Delta_{k}$ changes during the model training to show the learning bias, using heatmap values to represent the numerical values of $\Delta_{k}$.

Here, we generate a dataset with four key frequency components for the second case study \textbf{(Case 2)}. 
This study analyzes different modeling strategies to investigate their flexibility for debiasing.

\subsubsection{\textbf{Investigating the Frequency Bias of Transformer} \textbf{(Case 1)}}\label{subsec: CaseStudies.1}
As shown in Figure \ref{fig: case1}(a) (left), after 50 epochs of training, the model successfully captures the amplitude of low-frequency component $k1$ but fails to capture $k2$ and $k3$.
Meanwhile, the heatmap values show that the model predominantly focuses on learning the $k1$ component.
In other words, the relative error decreases to around 0.01 (red codes) during the training.
But, it lacks optimization for $k3$, resulting in a high relative error of almost 0.95.
These observations indicate that \textit{signals in the time domain can be represented by a series of frequency waves}, typically dominated by low-frequency components \cite{Digital_Signal_Processing,long_and_short_SiGIR18,long_and_short_KDD19}.
When the Transformer is deployed on this mixed-frequency collection, the dominant proportion of frequencies experiences a learning bias. 
A similar result is also evident in the control experiment in the right Subfigure.
Here, we introduce synthetic data with higher amplitudes in the mid and high-frequency ranges (resulting in $\va_{k1} < \va_{k2} < \va_{k3}$).
In response, the model shifts its focus towards the key component $k3$, leading to $\Delta_{k1} > \Delta_{k2} > \Delta_{k3}$.
This learning bias aligns with recent theoretical analyses of the Transformer model \citep{NIPS2023scan,antioversmooth_2022_ICLR, howTransWork_2022_ICLR}.
In addition, Sec. \ref{subsec: BiasDefinition} provides a formal definition of this frequency bias.

\subsubsection{\textbf{Debiasing the Frequency Learning for Transformer (Case 2).}}\label{subsec: CaseStudies.2}
Based on the above discussion, we initially use the same experimental settings for a new dataset, as shown in Figure \ref{fig: case1}(b) (Left).
We then perform two feasibility analyses for debiasing by (1) mitigating the influence of high proportionality and (2) providing the transformer with fine-grained frequency information.

\noindent \textit{(1) Frequency normalization:}
We first decompose the frequency domain and normalize the amplitudes of the frequencies to eliminate their proportional differences. 
Specifically, we apply the DFT, normalize the amplitudes, and then use the IDFT to convert the frequency representation back into the time domain before inputting it into the Transformer, formulated as
$ \mathbf{X}' = (\text{IDFT}(\mathbf{A}_{norm}))$.

As depicted in Figure \ref{fig: case1}(b) (middle and right), the four input components are adjusted so that they have the same amplitude value, shown by a blue dashed line.
The middle subfigure shows that frequency normalization enhances the forecasting performance for the latter three frequencies, but relative errors remain high.

\noindent \textit{(2) Frequency domain modeling:}
We further directly deploy the Transformer on the frequency domain to model the DFT matrix.
Subsequently, we apply the IDFT to return the forecasting outcome to the time domain.
Here, the purpose is to provide the transformer with more refined and disentangled frequency features.
Formally, $\mathbf{X}' = \text{IDFT}((\mathbf{A}_{norm}))$.
As shown in Figure \ref{fig: case1}(b) (right), there is a marked improvement in forecasting accuracy for the latter three frequency components. 
Notably, the bias in the second frequency component (60-75 Hz) is effectively eliminated.
These findings suggest the potential for direct \textit{\textbf{frequency domain modeling}} with \textit{\textbf{proportion mitigation}} in achieving the debiasing.

\section{Frequency Bias Formulation}
\label{sec: Formulation}

This section defines the frequency bias in Sec.\ref{subsec: BiasDefinition}, then describes the research problem in Sec.\ref{subsec: ProblemStatement}.

\subsection{Frequency Bias Definitions}\label{subsec: BiasDefinition}

Given the aforementioned empirical analyses, which demonstrate that a frequency bias exists in key frequency components, we first define these key components in terms of two properties: 1) a key component should have a relatively high amplitude within the spectrum, and 2) it should be consistent in historical observations and future time series, as well as robust to time shifts \citep{textbook1_broughton_discrete_2011, Digital_Signal_Processing}.

\begin{myDef}
\label{def:keycomponents}
\textbf{Key Frequency Components.}
Given a frequency spectrum $\mathbf{A}$ with length $L$, $\mathbf{A}$ can be segmented into $N$ sub-frequency bands $\{\vw_1,\vw_2,\ldots,\vw_N\}$ by a sliding window, where $\vw_n \in \mathbb{R}^{1 \times s}$.
The maximum amplitude in the $n$-th window is determined as follows:
\begin{equation} \label{eq:10}
\max(\vw_n) = \max {|\va_k| : \va_k \in \vw_n} \quad \text{for } n = 1, 2, \ldots, N
\end{equation}
where $\vw_n$ denotes $s$ amplitudes in the $n$-th window.
If $\va_k$ is a key component in the $i$-th window, then:

$$
\va_k = \max(\vw_n) \, \text{and} \, \, \hat{\va}_k = \max(\hat{\vw}_n)
$$

$\tilde{\mathbf{A}}$ is a collection of all key components. 
Notably, $\tilde{\mathbf{A}}$ should be present in historical $\mathbf{A}$ and ground truth $\hat{\mathbf{A}}$ for accurate forecasting.
\end{myDef}

\begin{myDef}
\label{def:frequencybias}
\textbf{Frequency Bias in Transformer.}
Given that a time series $\mathbf{X}$ contains $N$ key frequency components amplitudes $\tilde{\mathbf{A}} = \{ \tilde{\va}_1,\ldots,\tilde{\va}_N\}$, 
for the $k$-th component $\tilde{\va}_k \in \tilde{\mathbf{A}}$, we have $P(\tilde{\va}_k) = \frac{|\tilde{\va}_k|}{\sum^N_{n=1} |\tilde{\va}_{n}|} $, which refers to the proportion of $\tilde{\va}_k$ in the total sum of amplitudes of $\tilde{\mathbf{A}}$.
Frequency bias can be defined as relative error $\Delta_k$. 
Here, a larger proportion $P(\tilde{\va}_k)$
leads to a smaller $\Delta_k$ and exhibits a higher ranking:
\begin{equation}\label{eq:increasing}
    - |\Delta_k|  \varpropto P(\tilde{\va}_k)
\end{equation}

Eventually, the Transformer pays more attention to high-ranked components during the training, as seen in Figure \ref{fig: case1} (a) heatmaps.
\end{myDef}

\subsection{Problem Statement}\label{subsec: ProblemStatement}
Based on the discussions in Sec.~\ref{sec: PreliminaryAnalysis},
we argue that if the Transformer assigns attention to all key frequency components $\tilde{\mathbf{A}}$ equally during learning,
then the frequency bias could be mitigated.

\begin{myPro}
\label{problem}
\textbf{Debiasing Frequency Learning for Transformer.} 
Given a Transformer output $f_{Trans} (X)$, where $X$ contains several key frequency component $\tilde{\va}_k$, our goal is to debias $f_{Trans}$ and improve performance by making the relative error $\Delta\tilde{\va}_k$ independent of $P(\tilde{\va}_k)$:
\begin{equation}\label{eq:constant}
    - |\Delta_k|  \not\varpropto ¥P(\tilde{\va}_k)
\end{equation}
thereby ensuring a balanced response by the Transformer to different key frequency components.
\end{myPro}

\section{Fredformer}\label{sec: Fredformer}

Here, we discuss how to tackle the problem formulated in Sec.\ref{subsec: ProblemStatement} and propose \method, a \textbf{Fre}quency \textbf{d}ebiased Trans\textbf{former} model for accurate time series forecasting.

\begin{figure}[t]
\includegraphics[width=1\linewidth]{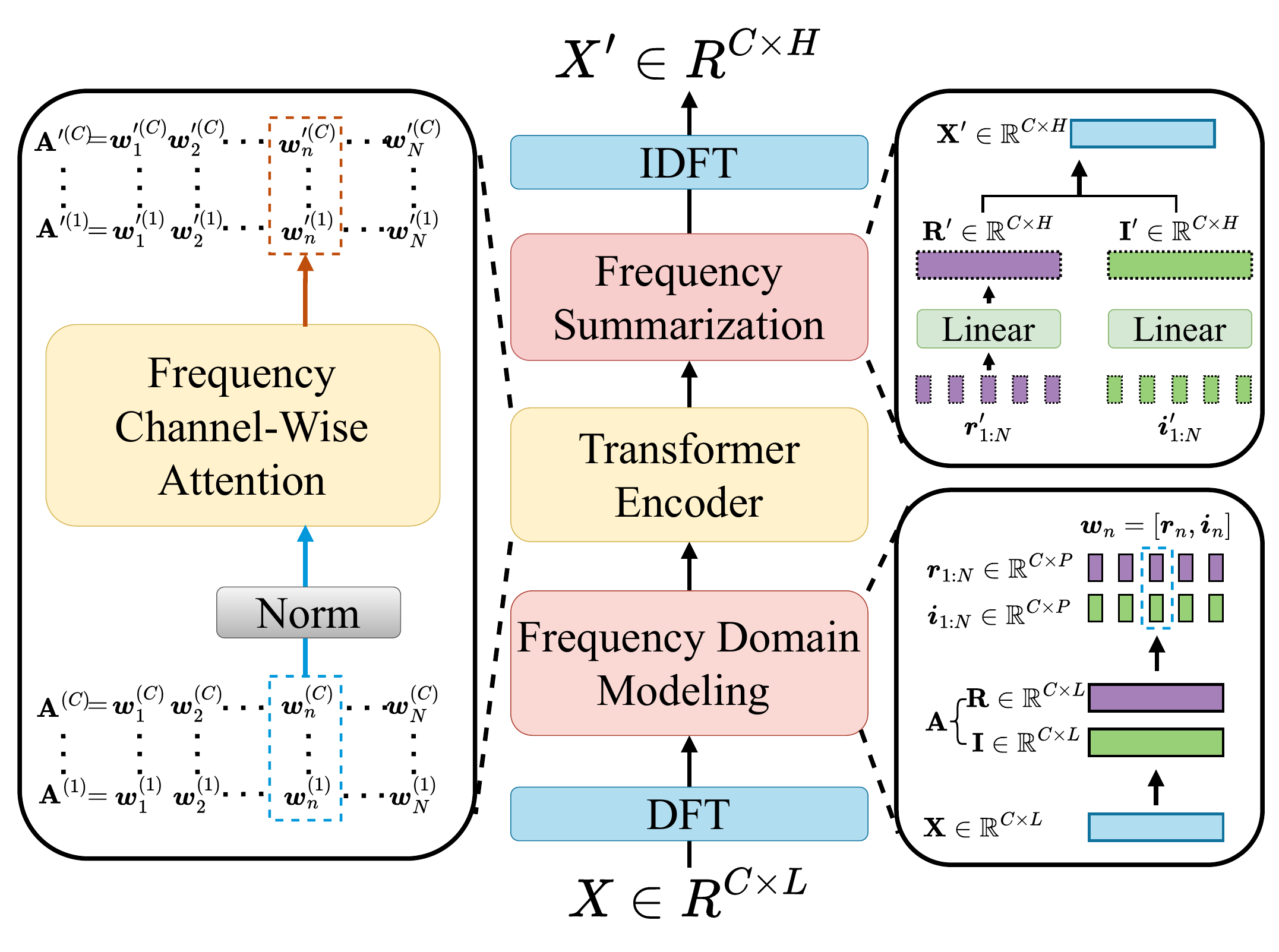}
\caption{ Overview of our framework. \method employs DFT to transform input sequences into the frequency domain, normalizes locally, and segments into patches before employing channel-wise attention, yielding final predictions through a frequency-wise summarizing layer and IDFT.
}
\label{fig: model}
\end{figure}

\noindent\textbf{Architecture Overview.}
\method consists of four principal components: $\rm(\hspace{.18em}i\hspace{.18em})$ a DFT-to-IDFT backbone, $\rm(\hspace{.18em}ii\hspace{.18em})$ frequency domain refinement, $\rm(\hspace{.18em}iii\hspace{.18em})$ local frequency independent learning, and $\rm(\hspace{.18em}iv\hspace{.18em})$ global semantic frequency summarization.
Figure \ref{fig: model} shows an architectural overview. 
The DFT-to-IDFT backbone breaks down the input time series $\mathbf{X}$ into its frequency components using DFT and learns a debiased representation of \textit{key frequency components} by modules $\rm(\hspace{.18em}ii\hspace{.18em})$ $\rm(\hspace{.18em}iii\hspace{.18em})$and $\rm(\hspace{.18em}iv\hspace{.18em})$.
Based on the discussion in Sec. \ref{subsec: CaseStudies.2} (2), where we noted the significant potential of frequency modeling for debiasing, we first refine the overall frequency spectrum into sub-frequencies, which we achieve through a patching operation on DFT coefficients.
Patches from different channels within the same sub-frequency band are embedded as tokens. 
That is, each sub-frequency band is encoded independently, which avoids the influence of other frequency components, as discussed in Section \ref{subsec: CaseStudies.2} (1).
We deploy the Transformer to extract local frequency features for each sub-band across all channels.
This mitigates the higher proportion crux defined in Def. \ref{def:frequencybias}.
Finally, we summarize all the frequency information, which serves as IDFT for forecasting.
A detailed workflow of \method is in Appendix \ref{appendix:algorithm1}.
Below, we provide a description of each module.

\subsection{Backbone} 
Given $ \mathbf{X}$, 
we first use DFT to decompose $\mathbf{X}$ into frequency coefficients $\mathbf{A}$\footnote{$\mathbf{A}$ consists of two coefficient matrices: a real part $\mathbf{R}\in \mathbb{R}^{C \times L}$ and an imaginary matrix $\mathbf{I} \in \mathbb{R}^{C \times L}$. Since all operations are conducted synchronously for these two matrices, we will refer to them as $\mathbf{A}$ in our subsequent discussions.} for all channels.
We then extract the debiased frequency features by using a Transformer encoder to $\mathbf{A} \in \mathbb{R}^{C \times L}$.
The frequency outputs are subsequently reconstructed to the time domain signal $\mathbf{X}'$ by IDFT.
$$
    {\mathbf{X}}' = \text{IDFT}(f_{Trans}(\mathbf{A})), \quad \mathbf{A} = \text{DFT}(\mathbf{X}))
$$

\vspace{0.3cm}

\subsection{Frequency Refinement and Normalization}
\label{sec:Frequency_Refinement_and_Normalization}

From the observations described in Sec. \ref{subsec: CaseStudies.2}, we conclude that if there are significant proportional differences between different $\tilde{\mathbf{A}}_k$ values in the input data, it will lead to the model overly focusing on components with larger amplitudes. 
To address this issue, we propose frequency refinement and normalization.
Specifically, a non-overlapping patching operation is applied to $\mathbf{A}$ along the $C$-axis (i.e., channel), resulting in a sequence of local sub-frequencies as follows:
$$
\mathbf{W} = \{\mathbf{W}_1,\mathbf{W}_2,\ldots,\mathbf{W}_N \} = \text{Patching}(\mathbf{A}), \quad \mathbf{W}_{n} \in \mathbb{R}^{C \times S}
$$
where $N$ is the total number of the patches, while $S$ represents the length of each patch.
Mitigating information redundancy over fine-grained frequency bands, such as neighboring 1 Hz and 2 Hz, allows the model to learn the local features in each sub-frequency.
Parameter $S$ is adaptable to the requirements of real-world scenarios, for example, an hourly sampling of daily recordings or the alpha waveform typically occurring at 8-12 Hz \citep{wake2}.

Since patching operation allows the model to manage each $\mathbf{M}_n$ independently, we further normalize each $\mathbf{W}_n$ along the $N$-axis:
$$
{\mathbf{W}^*_n} = \sigma({\mathbf{W}_n}) \quad n = 1, 2, \ldots, N
$$
where $\sigma(\cdot)$ denotes the normalization, and it further projects the numerical value of each $\tilde{\mathbf{A}}_k$ into a range of 0-1.
This operation eliminates proportionate differences in the maximum values within sub-frequency bands, thereby maintaining an equal $\Delta$ across all key components $\tilde{\mathbf{A}}$.
\noindent \begin{lemma}
\label{lemma:norm}
\textbf{Frequency-wise Local Normalization}: Given frequency patches $\forall\, \mathbf{W}_n,\, \mathbf{W}_m \in \mathbf{W}$ for $\max(\mathbf{W}_n) > \max(\mathbf{W}_m)$ and $\sigma(\cdot)$, the normalization strategy is defined by:
    $$
        \mathbf{W}^* = \{ \sigma(\mathbf{W}_1), \dots , \sigma(\mathbf{W}_N)\}
    $$
    This ensures that within each localized frequency patch $\mathbf{W}_n$, the amplitude differences between key frequency components are minimized, promoting equal attention to all key frequencies by the model:
    $$\label{eq:local_norm}
    \max(\mathbf{W}_n^*) = \max(\mathbf{W}_m^*)
    $$
\end{lemma}

Some studies also introduce patching operations in the time domain and perform normalization within these time domain patches \citep{PatchTST}.
However, according to Parseval's theorem \citep{parseval}, normalization within time domain patches is equivalent to normalizing across all frequencies.
This could not address the issue of amplitude bias among key frequency components.
A more detailed description can be found in Appendix \ref{app:lemma}.

\subsection{Frequency Local Independent Modeling}
Given the normalized $\mathbf{W}^*$, we deploy frequency local independent Transformer encoders to learn the importance of each $\mathbf{W}_n^*$ independently.
For $\mathbf{W}^{(1:C)}_n = \{\vw^{(1)}_n,\vw^{(2)}_n,\ldots,\vw^{(C)}_n\}_{n=1}^{N}$, a Transformer encoder $f_{Trans}(\cdot)$ accepts each ${\vw^*}^{(c)}_n$ as an input token:
$$
    \{\mathbf{W'}^{(1:C)}_n\} = f_{Trans}(\mathbf{W^*}^{(1:C)}_n)
$$ 
where $\mathbf{W'}^{(1:C)}_n$ is encoded by a channel-wise self-attention encoder:
$$
\begin{aligned}
\text{Attention}(\mathbf{Q}_n, \mathbf{K}_n, \mathbf{V}_n) 
= \\
& \hspace*{-6em} \text{Softmax}\left( \frac{\mathbf{W^*}_{n}^{(1:C)} \mathbf{W}^{q}_n  (\mathbf{W^*}_{n}^{(1:C)}\mathbf{W}^{k}_n)^T}{\sqrt{d}} \right) \mathbf{W^*}_{n}^{(1:C)} \mathbf{W}^{v}_n 
\end{aligned}
$$
where $\mathbf{W}^{q}_n, \mathbf{W}^{k}_n, \mathbf{W}^{v}_n \in \mathbb{R}^{S \times M}$ are the weight matrices for generating the query matrix $\mathbf{Q}_n$, key matrix $ \mathbf{K}_n$, and value matrix $\mathbf{V}_n$. 
$\sqrt{d}$ denotes a scaling operation.
The attention module also includes normalization and a feed-forward layer with residual connections \citep{dosovitskiy2021an},
and $\text{Attention}(\mathbf{Q}_n, \mathbf{K}_n,
\mathbf{V}_n) \in \mathbb{R}^{C\times M}$ weights the correlations among $C$ channels for the $n$-th sub-frequency band $\mathbf{M}_n$.
This design ensures that the features of each sub-frequency are calculated independently, preventing learning bias.

\begin{lemma}
\label{lemma:channel-wise}
Given $\ \mathbf{W^*}_n^{(1:C)} = \{\vw^{*(1)}_n, \vw^{*(2)}_n, \ldots, \vw^{*(C)}_n\}_{n=1}^{N}$, if $\ \mathbf{W'}_n = f_{Trans}(\mathbf{W^*}_n^{(1:C)})$,
then by modeling the relationships of identical frequencies $\vw^c_n$ across different channels, for the $k$-th key component $\tilde{\va}_k$ presents in ${\vw}_n^{(c)}$, we have $-|\Delta_k^{(c)}|  \varpropto \{|\Delta_k^{(c)}|\}{_{c=1}^{C}}$.
The Transformer encoders will focus on channel-wise correlations instead of the $\{|\Delta_k^{(c)}|\}{_{k=1}^{K}}$, i.e., debiasing $- |\Delta_k^{(c)}|  \not\varpropto P(\tilde{\va}_k)$.
\end{lemma}

Lemma \ref{lemma:channel-wise}, which indicates a lower $P(\tilde{\va}_k)$ does not necessarily lead to an increase in $|\Delta \tilde{\va}_k|$, thus avoiding disproportionate attention to frequency components.
Channel-wise attention is proposed in the work of \citep{Crossformer, liu2023itransformer}.
We include these studies as the baselines and the results in Sec. \ref{subsec: MainResults}.
This work has different modeling purposes: we deploy self-attention on the aligned local features, i.e., in the same frequency bands across channels, for frequency debiasing.

\begin{table*}[t]
  \caption{Multivariate forecasting results with prediction lengths $S\in\{96, 192, 336, 720\}$ for all datasets and fixed look-back length $T=96$.
  The \boldres{best} and \secondres{second best} results are highlighted.
  The full results of four selected datasets* will be shown in Figure \ref{tab:part_baseline_results}. Results are averaged from all prediction lengths. Full results for all datasets are listed in Appendix \ref{appendix:fullresult}.}
  \renewcommand{\arraystretch}{0.85} 
  \centering
  \resizebox{2\columnwidth}{!}{
  \begin{threeparttable}
  \begin{small}
  \renewcommand{\multirowsetup}{\centering}
  \setlength{\tabcolsep}{1.45pt}
  \begin{tabular}{c|cc|cc|cc|cc|cc|cc|cc|cc|cc|cc|cc|cc}
    \toprule
    {\multirow{2}{*}{Models}} & 
    \multicolumn{2}{c}{\rotatebox{0}{\scalebox{0.75}{\textbf{\method}}}} &
    \multicolumn{2}{c}{\rotatebox{0}{\scalebox{0.75}{iTransformer}}} &
    \multicolumn{2}{c}{\rotatebox{0}{\scalebox{0.8}{RLinear}}} &
    \multicolumn{2}{c}{\rotatebox{0}{\scalebox{0.8}{PatchTST}}} &
    \multicolumn{2}{c}{\rotatebox{0}{\scalebox{0.8}{Crossformer}}} &
    \multicolumn{2}{c}{\rotatebox{0}{\scalebox{0.8}{TiDE}}} &
    \multicolumn{2}{c}{\rotatebox{0}{\scalebox{0.8}{{TimesNet}}}} &
    \multicolumn{2}{c}{\rotatebox{0}{\scalebox{0.8}{DLinear}}} &
    \multicolumn{2}{c}{\rotatebox{0}{\scalebox{0.8}{SCINet}}} &
    \multicolumn{2}{c}{\rotatebox{0}{\scalebox{0.8}{FEDformer}}} &
    \multicolumn{2}{c}{\rotatebox{0}{\scalebox{0.8}{Stationary}}} &
    \multicolumn{2}{c}{\rotatebox{0}{\scalebox{0.8}{Autoformer}}}  \\
     &
    \multicolumn{2}{c}{\scalebox{0.8}{\textbf{(Ours)}}} &
    \multicolumn{2}{c}{\scalebox{0.8}{\textbf{(\citep{liu2023itransformer})}}} &
    \multicolumn{2}{c}{\scalebox{0.8}{\citep{li2023RLinear}}} & 
    \multicolumn{2}{c}{\scalebox{0.8}{\citep{PatchTST}}} & 
    \multicolumn{2}{c}{\scalebox{0.8}{\citep{Crossformer}}} & 
    \multicolumn{2}{c}{\scalebox{0.8}{\citep{das2023TiDE}}} & 
    \multicolumn{2}{c}{\scalebox{0.8}{\citep{Timesnet}}} & 
    \multicolumn{2}{c}{\scalebox{0.8}{\citep{DLinear}}} &
    \multicolumn{2}{c}{\scalebox{0.8}{\citep{liu2022scinet}}} & 
    \multicolumn{2}{c}{\scalebox{0.8}{\citep{ICMLFedformer}}} &
    \multicolumn{2}{c}{\scalebox{0.8}{\citep{Non-stationary}}} &
    \multicolumn{2}{c}{\scalebox{0.8}{\citep{Autoformer}}} \\
    \cmidrule(lr){2-3} \cmidrule(lr){4-5}\cmidrule(lr){6-7} \cmidrule(lr){8-9}\cmidrule(lr){10-11}\cmidrule(lr){12-13} \cmidrule(lr){14-15} \cmidrule(lr){16-17} \cmidrule(lr){18-19} \cmidrule(lr){20-21} \cmidrule(lr){22-23} \cmidrule(lr){24-25}
    
    {Metric}  & \scalebox{0.8}{MSE} & \scalebox{0.8}{MAE} & \scalebox{0.8}{MSE} & \scalebox{0.8}{MAE}  & \scalebox{0.8}{MSE} & \scalebox{0.8}{MAE}  & \scalebox{0.8}{MSE} & \scalebox{0.8}{MAE}  & \scalebox{0.8}{MSE} & \scalebox{0.8}{MAE}  & \scalebox{0.8}{MSE} & \scalebox{0.8}{MAE}  & \scalebox{0.8}{MSE} & \scalebox{0.8}{MAE} & \scalebox{0.8}{MSE} & \scalebox{0.8}{MAE} & \scalebox{0.8}{MSE} & \scalebox{0.8}{MAE} & \scalebox{0.8}{MSE} & \scalebox{0.8}{MAE} & \scalebox{0.8}{MSE} & \scalebox{0.8}{MAE} & \scalebox{0.8}{MSE} & \scalebox{0.8}{MAE} \\
    
    \toprule
    
    \scalebox{0.95}{ECL} & \boldres{\scalebox{0.8}{0.175}} & \boldres{\scalebox{0.8}{0.269}} & \secondres{\scalebox{0.8}{0.178}} & \secondres{\scalebox{0.8}{0.270}} &\scalebox{0.8}{0.219} &\scalebox{0.8}{0.298} & \scalebox{0.8}{0.216} & \scalebox{0.8}{0.304} & \scalebox{0.8}{0.244} & \scalebox{0.8}{0.334}  & \scalebox{0.8}{0.251} & \scalebox{0.8}{0.344} &\scalebox{0.8}{0.192} &\scalebox{0.8}{0.295} &\scalebox{0.8}{0.212} &\scalebox{0.8}{0.300} & \scalebox{0.8}{0.268} & \scalebox{0.8}{0.365} &\scalebox{0.8}{0.214} &\scalebox{0.8}{0.327} &{\scalebox{0.8}{0.193}} &{\scalebox{0.8}{0.296}} &\scalebox{0.8}{0.227} &\scalebox{0.8}{0.338} \\
    
    \midrule
    
    \scalebox{0.95}{ETTh1} & \boldres{\scalebox{0.8}{0.435}} & \boldres{\scalebox{0.8}{0.426}} & \scalebox{0.8}{0.454} & \scalebox{0.8}{0.447} & \scalebox{0.8}{0.446} & \secondres{\scalebox{0.8}{0.434}} & \scalebox{0.8}{0.469} & \scalebox{0.8}{0.454} & \scalebox{0.8}{{0.529}} & \scalebox{0.8}{{0.522}} & \scalebox{0.8}{{0.541}} & \scalebox{0.8}{{0.507}} & \scalebox{0.8}{{0.548}} & \scalebox{0.8}{{0.450}} & \scalebox{0.8}{{0.456}} & \scalebox{0.8}{{0.452}} & \scalebox{0.8}{{0.747}} & \scalebox{0.8}{{0.647}} & \secondres{\scalebox{0.8}{{0.440}}} & \scalebox{0.8}{{0.460}} & \scalebox{0.8}{{0.570}} & \scalebox{0.8}{{0.537}} & \scalebox{0.8}{{0.496}} & \scalebox{0.8}{{0.487}} \\
    
    \midrule
    
    \scalebox{0.95}{ETTh2}* & \boldres{\scalebox{0.8}{0.365}} & \boldres{\scalebox{0.8}{0.393}} & {\scalebox{0.8}{0.383}} & {\scalebox{0.8}{0.407}} & \secondres{\scalebox{0.8}{{0.374}}} & \secondres{\scalebox{0.8}{0.398}} & \scalebox{0.8}{0.387} & \scalebox{0.8}{0.407} & \scalebox{0.8}{{0.942}} & \scalebox{0.8}{{0.684}} & \scalebox{0.8}{{0.611}} & \scalebox{0.8}{{0.550}} & \scalebox{0.8}{{0.414}} & \scalebox{0.8}{{0.427}} & \scalebox{0.8}{{0.559}} & \scalebox{0.8}{{0.515}} & \scalebox{0.8}{{0.954}} & \scalebox{0.8}{{0.723}} & \scalebox{0.8}{{0.437}} & \scalebox{0.8}{{0.449}} & \scalebox{0.8}{{0.526}} & \scalebox{0.8}{{0.516}} & \scalebox{0.8}{{0.450}} & \scalebox{0.8}{{0.459}} \\

    \midrule
    
    \scalebox{0.95}{ETTm1}* & \boldres{\scalebox{0.8}{0.384}} & \boldres{\scalebox{0.8}{0.395}} & {\scalebox{0.8}{0.407}} & {\scalebox{0.8}{0.410}} & \scalebox{0.8}{{0.414}} & \scalebox{0.8}{0.407} & \secondres{\scalebox{0.8}{0.387}} & \secondres{\scalebox{0.8}{0.400}} & \scalebox{0.8}{{0.513}} & \scalebox{0.8}{{0.496}} & \scalebox{0.8}{{0.419}} & \scalebox{0.8}{{0.419}} & \scalebox{0.8}{{0.400}} & \scalebox{0.8}{{0.406}} & \scalebox{0.8}{{0.403}} & \scalebox{0.8}{{0.407}} & \scalebox{0.8}{{0.485}} & \scalebox{0.8}{{0.481}} & \scalebox{0.8}{{0.448}} & \scalebox{0.8}{{0.452}} & \scalebox{0.8}{{0.481}} & \scalebox{0.8}{{0.456}} & \scalebox{0.8}{{0.588}} & \scalebox{0.8}{{0.517}} \\

    \midrule
    
    \scalebox{0.95}{ETTm2} & \boldres{\scalebox{0.8}{0.279}} & \boldres{\scalebox{0.8}{0.324}} & {\scalebox{0.8}{0.288}} & {\scalebox{0.8}{0.332}} & \scalebox{0.8}{{0.286}} & \scalebox{0.8}{0.327} & \secondres{\scalebox{0.8}{0.281}} & \secondres{\scalebox{0.8}{0.326}} & \scalebox{0.8}{{0.757}} & \scalebox{0.8}{{0.610}} & \scalebox{0.8}{{0.358}} & \scalebox{0.8}{{0.404}} & \scalebox{0.8}{{0.291}} & \scalebox{0.8}{{0.333}} & \scalebox{0.8}{{0.350}} & \scalebox{0.8}{{0.401}} & \scalebox{0.8}{{0.571}} & \scalebox{0.8}{{0.537}} & \scalebox{0.8}{{0.305}} & \scalebox{0.8}{{0.349}} & \scalebox{0.8}{{0.306}} & \scalebox{0.8}{{0.347}} & \scalebox{0.8}{{0.327}} & \scalebox{0.8}{{0.371}} \\

    \midrule 
    
    \scalebox{0.95}{Traffic} & \secondres{\scalebox{0.8}{0.431}} & \secondres{\scalebox{0.8}{0.287}} & \boldres{\scalebox{0.8}{0.428}} & \boldres{\scalebox{0.8}{0.282}} &\scalebox{0.8}{0.626} &\scalebox{0.8}{0.378} & \scalebox{0.8}{0.555} & \scalebox{0.8}{0.362} & \scalebox{0.8}{0.550} & \scalebox{0.8}{0.304} & \scalebox{0.8}{0.760} & \scalebox{0.8}{0.473} &{\scalebox{0.8}{0.620}} &{\scalebox{0.8}{0.336}} &\scalebox{0.8}{0.625} &\scalebox{0.8}{0.383} & \scalebox{0.8}{0.804} & \scalebox{0.8}{0.509} &{\scalebox{0.8}{0.610}} &\scalebox{0.8}{0.376} &\scalebox{0.8}{0.624} &{\scalebox{0.8}{0.340}} &\scalebox{0.8}{0.628} &\scalebox{0.8}{0.379} \\
    
    \midrule
    
    \scalebox{0.95}{Weather}* & \boldres{\scalebox{0.8}{0.246}} & \boldres{\scalebox{0.8}{0.272}} & \secondres{\scalebox{0.8}{0.258}} & \secondres{\scalebox{0.8}{0.279}} &\scalebox{0.8}{0.272} &\scalebox{0.8}{0.291} & \scalebox{0.8}{0.259} & \scalebox{0.8}{0.281} & \scalebox{0.8}{0.259} & \scalebox{0.8}{0.315}  & \scalebox{0.8}{0.271} & \scalebox{0.8}{0.320} &{\scalebox{0.8}{0.259}} &{\scalebox{0.8}{0.287}} &\scalebox{0.8}{0.265} &\scalebox{0.8}{0.317} & \scalebox{0.8}{0.292} & \scalebox{0.8}{0.363} &\scalebox{0.8}{0.309} &\scalebox{0.8}{0.360} &\scalebox{0.8}{0.288} &\scalebox{0.8}{0.314} &\scalebox{0.8}{0.338} &\scalebox{0.8}{0.382} \\
    
    \midrule
    
    \scalebox{0.95}{Solar-Energy}* & \boldres{\scalebox{0.8}{0.226}} & \boldres{\scalebox{0.8}{0.261}} &\secondres{\scalebox{0.8}{0.233}} &\secondres{\scalebox{0.8}{0.262}} &\scalebox{0.8}{0.369} &\scalebox{0.8}{0.356} &\scalebox{0.8}{0.270} &\scalebox{0.8}{0.307} &\scalebox{0.8}{0.641} &\scalebox{0.8}{0.639} &\scalebox{0.8}{0.347} &\scalebox{0.8}{0.417} &\scalebox{0.8}{0.301} &\scalebox{0.8}{0.319} &\scalebox{0.8}{0.330} &\scalebox{0.8}{0.401} &\scalebox{0.8}{0.282} &\scalebox{0.8}{0.375} &\scalebox{0.8}{0.291} &\scalebox{0.8}{0.381} &\scalebox{0.8}{0.261} &\scalebox{0.8}{0.381} &\scalebox{0.8}{0.885} &\scalebox{0.8}{0.711} \\

    \bottomrule
  \end{tabular}
    \end{small}
  \end{threeparttable}
}
    \label{tab:main_result}
\end{table*}

\begin{table*}[t]
  \caption{Full results of four selected datasets, with the \boldres{best} and \secondres{second best} results are highlighted. We compare extensive competitive models under different prediction lengths following the setting of iTransformer \citep{liu2023itransformer}. The input sequence length is set to 96 for all baselines. Avg means the average results from all four prediction lengths.}\label{tab:part_baseline_results}
  \vskip -0.0in
  \vspace{3pt}
  \renewcommand{\arraystretch}{0.85} 
  \centering
  \resizebox{2\columnwidth}{!}{
  \renewcommand{\multirowsetup}{\centering}
  \setlength{\tabcolsep}{1pt}
  \begin{tabular}{c|c|cc|cc|cc|cc|cc|cc|cc|cc|cc|cc|cc|cc}
    \toprule
    \multicolumn{2}{c}{\multirow{2}{*}{Models}} & 
    \multicolumn{2}{c}{\rotatebox{0}{\scalebox{0.8}{\textbf{\method}}}} &
    \multicolumn{2}{c}{\rotatebox{0}{\scalebox{0.8}{iTransformer}}} &
    \multicolumn{2}{c}{\rotatebox{0}{\scalebox{0.8}{RLinear}}} &
    \multicolumn{2}{c}{\rotatebox{0}{\scalebox{0.8}{PatchTST}}} &
    \multicolumn{2}{c}{\rotatebox{0}{\scalebox{0.8}{Crossformer}}}  &
    \multicolumn{2}{c}{\rotatebox{0}{\scalebox{0.8}{TiDE}}} &
    \multicolumn{2}{c}{\rotatebox{0}{\scalebox{0.8}{{TimesNet}}}} &
    \multicolumn{2}{c}{\rotatebox{0}{\scalebox{0.8}{DLinear}}}&
    \multicolumn{2}{c}{\rotatebox{0}{\scalebox{0.8}{SCINet}}} &
    \multicolumn{2}{c}{\rotatebox{0}{\scalebox{0.8}{FEDformer}}} &
    \multicolumn{2}{c}{\rotatebox{0}{\scalebox{0.8}{Stationary}}} &
    \multicolumn{2}{c}{\rotatebox{0}{\scalebox{0.8}{Autoformer}}} \\

    \multicolumn{2}{c}{} &
    \multicolumn{2}{c}{\scalebox{0.8}{\textbf{(Ours)}}} & 
    \multicolumn{2}{c}{\scalebox{0.8}{\citeyearpar{liu2023itransformer}}} & 
    \multicolumn{2}{c}{\scalebox{0.8}{\citeyearpar{li2023RLinear}}} & 
    \multicolumn{2}{c}{\scalebox{0.8}{\citeyearpar{PatchTST}}} & 
    \multicolumn{2}{c}{\scalebox{0.8}{\citeyearpar{Crossformer}}}  & 
    \multicolumn{2}{c}{\scalebox{0.8}{\citeyearpar{das2023TiDE}}} & 
    \multicolumn{2}{c}{\scalebox{0.8}{\citeyearpar{Timesnet}}} & 
    \multicolumn{2}{c}{\scalebox{0.8}{\citeyearpar{DLinear}}}& 
    \multicolumn{2}{c}{\scalebox{0.8}{\citeyearpar{liu2022scinet}}} &
    \multicolumn{2}{c}{\scalebox{0.8}{\citeyearpar{ICMLFedformer}}} &
    \multicolumn{2}{c}{\scalebox{0.8}{\citeyearpar{Non-stationary}}} &
    \multicolumn{2}{c}{\scalebox{0.8}{\citeyearpar{Autoformer}}} \\
    
    \cmidrule(lr){3-4} \cmidrule(lr){5-6}\cmidrule(lr){7-8} \cmidrule(lr){9-10}\cmidrule(lr){11-12}\cmidrule(lr){13-14} \cmidrule(lr){15-16} \cmidrule(lr){17-18} \cmidrule(lr){19-20} \cmidrule(lr){21-22} \cmidrule(lr){23-24}\cmidrule(lr){25-26}
    
    \multicolumn{2}{c}{Metric}  & \scalebox{0.78}{MSE} & \scalebox{0.78}{MAE} & \scalebox{0.78}{MSE} & \scalebox{0.78}{MAE}  & \scalebox{0.78}{MSE} & \scalebox{0.78}{MAE}  & \scalebox{0.78}{MSE} & \scalebox{0.78}{MAE} & \scalebox{0.78}{MSE} & \scalebox{0.78}{MAE}  & \scalebox{0.78}{MSE} & \scalebox{0.78}{MAE}  & \scalebox{0.78}{MSE} & \scalebox{0.78}{MAE} & \scalebox{0.78}{MSE} & \scalebox{0.78}{MAE} & \scalebox{0.78}{MSE} & \scalebox{0.78}{MAE} & \scalebox{0.78}{MSE} & \scalebox{0.78}{MAE} & \scalebox{0.78}{MSE} & \scalebox{0.78}{MAE} & \scalebox{0.78}{MSE} & \scalebox{0.78}{MAE} \\
    
    \toprule

    \multirow{5}{*}{\rotatebox{90}{\scalebox{0.95}{ETTh2}}}
    &  \scalebox{0.78}{96} & \secondres{\scalebox{0.78}{0.293}} & \secondres{\scalebox{0.78}{0.342}} & \scalebox{0.78}{0.297} & {\scalebox{0.78}{0.349}} & \boldres{\scalebox{0.78}{0.288}} & \boldres{\scalebox{0.78}{0.338}} & {\scalebox{0.78}{0.302}} & \scalebox{0.78}{0.348} & \scalebox{0.78}{0.745} & \scalebox{0.78}{0.584} &\scalebox{0.78}{0.400} & \scalebox{0.78}{0.440}  & {\scalebox{0.78}{0.340}} & {\scalebox{0.78}{0.374}} &{\scalebox{0.78}{0.333}} &{\scalebox{0.78}{0.387}} & \scalebox{0.78}{0.707} & \scalebox{0.78}{0.621}  &\scalebox{0.78}{0.358} &\scalebox{0.78}{0.397} &\scalebox{0.78}{0.476} &\scalebox{0.78}{0.458} &\scalebox{0.78}{0.346} &\scalebox{0.78}{0.388} \\ 
    
    & \scalebox{0.78}{192} & \boldres{\scalebox{0.78}{0.371}} & \boldres{\scalebox{0.78}{0.389}} & \scalebox{0.78}{0.380} & \scalebox{0.78}{0.400}& \secondres{\scalebox{0.78}{0.374}} & \secondres{\scalebox{0.78}{0.390}} &{\scalebox{0.78}{0.388}} & {\scalebox{0.78}{0.400}} & \scalebox{0.78}{0.877} & \scalebox{0.78}{0.656} & \scalebox{0.78}{0.528} & \scalebox{0.78}{0.509} & {\scalebox{0.78}{0.402}} & {\scalebox{0.78}{0.414}} &\scalebox{0.78}{0.477} &\scalebox{0.78}{0.476} & \scalebox{0.78}{0.860} & \scalebox{0.78}{0.689} &{\scalebox{0.78}{0.429}} &{\scalebox{0.78}{0.439}} &\scalebox{0.78}{0.512} &\scalebox{0.78}{0.493} &\scalebox{0.78}{0.456} &\scalebox{0.78}{0.452} \\ 
    
    & \scalebox{0.78}{336} & \boldres{\scalebox{0.78}{0.382}} & \boldres{\scalebox{0.78}{0.409}} & {\scalebox{0.78}{0.428}} & \scalebox{0.78}{0.432} & \secondres{\scalebox{0.78}{0.415}} & \secondres{\scalebox{0.78}{0.426}} & \scalebox{0.78}{0.426} & {\scalebox{0.78}{0.433}}& \scalebox{0.78}{1.043} & \scalebox{0.78}{0.731} & \scalebox{0.78}{0.643} & \scalebox{0.78}{0.571}  & {\scalebox{0.78}{0.452}} & {\scalebox{0.78}{0.452}} &\scalebox{0.78}{0.594} &\scalebox{0.78}{0.541} & \scalebox{0.78}{1.000} &\scalebox{0.78}{0.744} &\scalebox{0.78}{0.496} &\scalebox{0.78}{0.487} &\scalebox{0.78}{0.552} &\scalebox{0.78}{0.551} &{\scalebox{0.78}{0.482}} &\scalebox{0.78}{0.486}\\ 
    
    & \scalebox{0.78}{720} & \boldres{\scalebox{0.78}{0.415}} & \boldres{\scalebox{0.78}{0.434}} & \scalebox{0.78}{0.427} & \scalebox{0.78}{0.445} & \secondres{\scalebox{0.78}{0.420}} & \secondres{\scalebox{0.78}{0.440}} & {\scalebox{0.78}{0.431}} & {\scalebox{0.78}{0.446}} & \scalebox{0.78}{1.104} & \scalebox{0.78}{0.763} & \scalebox{0.78}{0.874} & \scalebox{0.78}{0.679} & {\scalebox{0.78}{0.462}} & {\scalebox{0.78}{0.468}} &\scalebox{0.78}{0.831} &\scalebox{0.78}{0.657} & \scalebox{0.78}{1.249} & \scalebox{0.78}{0.838} &{\scalebox{0.78}{0.463}} &{\scalebox{0.78}{0.474}} &\scalebox{0.78}{0.562} &\scalebox{0.78}{0.560} &\scalebox{0.78}{0.515} &\scalebox{0.78}{0.511} \\ 

    \midrule

    & \scalebox{0.78}{Avg} & \boldres{\scalebox{0.78}{0.365}} & \boldres{\scalebox{0.78}{0.393}} & \scalebox{0.78}{0.383} & \scalebox{0.78}{0.407} & \secondres{\scalebox{0.78}{0.374}} & \secondres{\scalebox{0.78}{0.398}} & {\scalebox{0.78}{0.387}} & {\scalebox{0.78}{0.407}} & \scalebox{0.78}{0.942} & \scalebox{0.78}{0.684} & \scalebox{0.78}{0.611} & \scalebox{0.78}{0.550}  &{\scalebox{0.78}{0.414}} &{\scalebox{0.78}{0.427}} &\scalebox{0.78}{0.559} &\scalebox{0.78}{0.515} & \scalebox{0.78}{0.954} & \scalebox{0.78}{0.723} &\scalebox{0.78}{{0.437}} &\scalebox{0.78}{{0.449}} &\scalebox{0.78}{0.526} &\scalebox{0.78}{0.516} &\scalebox{0.78}{0.450} &\scalebox{0.78}{0.459} \\ 
    
    \cmidrule(lr){2-26}

    \multirow{5}{*}{\rotatebox{90}{\scalebox{0.95}{ETTm1}}}
    &  \scalebox{0.78}{96} & \boldres{\scalebox{0.78}{0.326}} & \boldres{\scalebox{0.78}{0.361}} & \scalebox{0.78}{0.334} & \scalebox{0.78}{0.368} & \scalebox{0.78}{0.355} & \scalebox{0.78}{0.376} & \secondres{\scalebox{0.78}{0.329}} & \secondres{\scalebox{0.78}{0.367}} & \scalebox{0.78}{0.404} & \scalebox{0.78}{0.426} & \scalebox{0.78}{0.364} & \scalebox{0.78}{0.387} &{\scalebox{0.78}{0.338}} &{\scalebox{0.78}{0.375}} &{\scalebox{0.78}{0.345}} &{\scalebox{0.78}{0.372}} & \scalebox{0.78}{0.418} & \scalebox{0.78}{0.438} &\scalebox{0.78}{0.379} &\scalebox{0.78}{0.419} &\scalebox{0.78}{0.386} &\scalebox{0.78}{0.398} &\scalebox{0.78}{0.505} &\scalebox{0.78}{0.475} \\ 
    
    & \scalebox{0.78}{192} & \boldres{\scalebox{0.78}{0.363}} & \boldres{\scalebox{0.78}{0.380}} & \scalebox{0.78}{0.377} & \scalebox{0.78}{0.391} & \scalebox{0.78}{0.391} & \scalebox{0.78}{0.392} & \secondres{\scalebox{0.78}{0.367}} & \secondres{\scalebox{0.78}{0.385}} & \scalebox{0.78}{0.450} & \scalebox{0.78}{0.451} &\scalebox{0.78}{0.398} & \scalebox{0.78}{0.404} &\scalebox{0.78}{0.374} & \scalebox{0.78}{0.387} &{\scalebox{0.78}{0.380}} &{\scalebox{0.78}{0.389}} & \scalebox{0.78}{0.439} & \scalebox{0.78}{0.450}  &\scalebox{0.78}{0.426} &\scalebox{0.78}{0.441} &\scalebox{0.78}{0.459} &\scalebox{0.78}{0.444} &\scalebox{0.78}{0.553} &\scalebox{0.78}{0.496} \\ 
    
    & \scalebox{0.78}{336} &\boldres{ \scalebox{0.78}{0.395}} & \boldres{\scalebox{0.78}{0.403}} & \scalebox{0.78}{0.426} & \scalebox{0.78}{0.420} & \scalebox{0.78}{0.424} & \scalebox{0.78}{0.415} & \secondres{\scalebox{0.78}{0.399}} & \secondres{\scalebox{0.78}{0.410}} & \scalebox{0.78}{0.532}  &\scalebox{0.78}{0.515} & \scalebox{0.78}{0.428} & \scalebox{0.78}{0.425} & \scalebox{0.78}{0.410} & \scalebox{0.78}{0.411}  &{\scalebox{0.78}{0.413}} &{\scalebox{0.78}{0.413}} & \scalebox{0.78}{0.490} & \scalebox{0.78}{0.485}  &\scalebox{0.78}{0.445} &\scalebox{0.78}{0.459} &\scalebox{0.78}{0.495} &\scalebox{0.78}{0.464} &\scalebox{0.78}{0.621} &\scalebox{0.78}{0.537} \\ 
    
    & \scalebox{0.78}{720} & \boldres{\scalebox{0.78}{0.453}} & \boldres{\scalebox{0.78}{0.438}} & \scalebox{0.78}{0.491} & \scalebox{0.78}{0.459} & \scalebox{0.78}{0.487} & \scalebox{0.78}{0.450} & \secondres{\scalebox{0.78}{0.454}} & \secondres{\scalebox{0.78}{0.439}} & \scalebox{0.78}{0.666} & \scalebox{0.78}{0.589} & \scalebox{0.78}{0.487} & \scalebox{0.78}{0.461} &{\scalebox{0.78}{0.478}} & \scalebox{0.78}{0.450} & \scalebox{0.78}{0.474} &{\scalebox{0.78}{0.453}} & \scalebox{0.78}{0.595} & \scalebox{0.78}{0.550}  &\scalebox{0.78}{0.543} &\scalebox{0.78}{0.490} &\scalebox{0.78}{0.585} &\scalebox{0.78}{0.516} &\scalebox{0.78}{0.671} &\scalebox{0.78}{0.561} \\ 
    
    \cmidrule(lr){2-26}
    
    & \scalebox{0.78}{Avg} & \boldres{\scalebox{0.78}{0.384}} & \boldres{\scalebox{0.78}{0.395}} & \scalebox{0.78}{0.407} & \scalebox{0.78}{0.410} & \scalebox{0.78}{0.414} & \scalebox{0.78}{0.407} & \secondres{\scalebox{0.78}{0.387}} & \secondres{\scalebox{0.78}{0.400}} & \scalebox{0.78}{0.513} & \scalebox{0.78}{0.496} & \scalebox{0.78}{0.419} & \scalebox{0.78}{0.419} & \scalebox{0.78}{0.400} & \scalebox{0.78}{0.406}  &{\scalebox{0.78}{0.403}} &{\scalebox{0.78}{0.407}} & \scalebox{0.78}{0.485} & \scalebox{0.78}{0.481}  &\scalebox{0.78}{0.448} &\scalebox{0.78}{0.452} &\scalebox{0.78}{0.481} &\scalebox{0.78}{0.456} &\scalebox{0.78}{0.588} &\scalebox{0.78}{0.517} \\ 

    \midrule
    
    \multirow{5}{*}{\rotatebox{90}{\scalebox{0.95}{Weather}}} 
    &  \scalebox{0.78}{96} & \secondres{\scalebox{0.78}{0.163}} & \boldres{\scalebox{0.78}{0.207}} & \scalebox{0.78}{0.174} & \secondres{\scalebox{0.78}{0.214}} & \scalebox{0.78}{0.192} & \scalebox{0.78}{0.232} & \scalebox{0.78}{0.177} & \scalebox{0.78}{0.218} & \boldres{\scalebox{0.78}{0.158}} & \scalebox{0.78}{0.230}  & \scalebox{0.78}{0.202} & \scalebox{0.78}{0.261} &\scalebox{0.78}{0.172} &{\scalebox{0.78}{0.220}} & \scalebox{0.78}{0.196} &\scalebox{0.78}{0.255} & \scalebox{0.78}{0.221} & \scalebox{0.78}{0.306} & \scalebox{0.78}{0.217} &\scalebox{0.78}{0.296} & {\scalebox{0.78}{0.173}} &{\scalebox{0.78}{0.223}} & \scalebox{0.78}{0.266} &\scalebox{0.78}{0.336} \\ 
    
    & \scalebox{0.78}{192} & \secondres{\scalebox{0.78}{0.211}} & \boldres{\scalebox{0.78}{0.251}} & \scalebox{0.78}{0.221} & \secondres{\scalebox{0.78}{0.254}} & \scalebox{0.78}{0.240} & \scalebox{0.78}{0.271} & \scalebox{0.78}{0.225} & \scalebox{0.78}{0.259} & \boldres{\scalebox{0.78}{0.206}} & \scalebox{0.78}{0.277} & \scalebox{0.78}{0.242} & \scalebox{0.78}{0.298} &\scalebox{0.78}{0.219} &\scalebox{0.78}{0.261} & \scalebox{0.78}{0.237} &\scalebox{0.78}{0.296} & \scalebox{0.78}{0.261} & \scalebox{0.78}{0.340} & \scalebox{0.78}{0.276} &\scalebox{0.78}{0.336} & \scalebox{0.78}{0.245} &\scalebox{0.78}{0.285} & \scalebox{0.78}{0.307} &\scalebox{0.78}{0.367} \\ 
    
    & \scalebox{0.78}{336} & \boldres{\scalebox{0.78}{0.267}} & \boldres{\scalebox{0.78}{0.292}} & \scalebox{0.78}{0.278} & \secondres{\scalebox{0.78}{0.296}} & \scalebox{0.78}{0.292} & \scalebox{0.78}{0.307} & \scalebox{0.78}{0.278} & \scalebox{0.78}{0.297} & \secondres{\scalebox{0.78}{0.272}} & \scalebox{0.78}{0.335} & \scalebox{0.78}{0.287} & \scalebox{0.78}{0.335} &{\scalebox{0.78}{0.280}} &{\scalebox{0.78}{0.306}} & \scalebox{0.78}{0.283} &\scalebox{0.78}{0.335} & \scalebox{0.78}{0.309} & \scalebox{0.78}{0.378} & \scalebox{0.78}{0.339} &\scalebox{0.78}{0.380} & \scalebox{0.78}{0.321} &\scalebox{0.78}{0.338} & \scalebox{0.78}{0.359} &\scalebox{0.78}{0.395}\\ 
    
    & \scalebox{0.78}{720} & \boldres{\scalebox{0.78}{0.343}} & \boldres{\scalebox{0.78}{0.341}} & \scalebox{0.78}{0.358} & \secondres{\scalebox{0.78}{0.349}} & \scalebox{0.78}{0.364} & \scalebox{0.78}{0.353} & \scalebox{0.78}{0.354} & \scalebox{0.78}{0.348} & \scalebox{0.78}{0.398} & \scalebox{0.78}{0.418} & \secondres{\scalebox{0.78}{0.351}} & \scalebox{0.78}{0.386} &\scalebox{0.78}{0.365} &{\scalebox{0.78}{0.359}} & {\scalebox{0.78}{0.345}} &{\scalebox{0.78}{0.381}} & \scalebox{0.78}{0.377} & \scalebox{0.78}{0.427} & \scalebox{0.78}{0.403} &\scalebox{0.78}{0.428} & \scalebox{0.78}{0.414} &\scalebox{0.78}{0.410} & \scalebox{0.78}{0.419} &\scalebox{0.78}{0.428} \\ 
    
    \cmidrule(lr){2-26}
    
    & \scalebox{0.78}{Avg} & \boldres{\scalebox{0.78}{0.246}} & \boldres{\scalebox{0.78}{0.272}} & \secondres{\scalebox{0.78}{0.258}} & \secondres{\scalebox{0.78}{0.279}} & \scalebox{0.78}{0.272} & \scalebox{0.78}{0.291} & \scalebox{0.78}{0.259} & \scalebox{0.78}{0.281} & \scalebox{0.78}{0.259} & \scalebox{0.78}{0.315} & \scalebox{0.78}{0.271} & \scalebox{0.78}{0.320} &{\scalebox{0.78}{0.259}} &{\scalebox{0.78}{0.287}} &\scalebox{0.78}{0.265} &\scalebox{0.78}{0.317} & \scalebox{0.78}{0.292} & \scalebox{0.78}{0.363} &\scalebox{0.78}{0.309} &\scalebox{0.78}{0.360} &\scalebox{0.78}{0.288} &\scalebox{0.78}{0.314} &\scalebox{0.78}{0.338} &\scalebox{0.78}{0.382} \\ 
    
    \midrule
    
    \multirow{5}{*}{\rotatebox{90}{\scalebox{0.95}{Solar-Energy}}} 
    &  \scalebox{0.78}{96} & \boldres{\scalebox{0.78}{0.185}} & \boldres{\scalebox{0.78}{0.233}} &\secondres{\scalebox{0.78}{0.203}} & \secondres{\scalebox{0.78}{0.237}} & \scalebox{0.78}{0.322} & \scalebox{0.78}{0.339} & \scalebox{0.78}{0.234} & \scalebox{0.78}{0.286} &\scalebox{0.78}{0.310} &\scalebox{0.78}{0.331} &\scalebox{0.78}{0.312} &\scalebox{0.78}{0.399} &\scalebox{0.78}{0.250} &\scalebox{0.78}{0.292} &\scalebox{0.78}{0.290} &\scalebox{0.78}{0.378} &\scalebox{0.78}{0.237} &\scalebox{0.78}{0.344} &\scalebox{0.78}{0.242} &\scalebox{0.78}{0.342} &\scalebox{0.78}{0.215} &\scalebox{0.78}{0.249} &\scalebox{0.78}{0.884} &\scalebox{0.78}{0.711}\\ 
    
    & \scalebox{0.78}{192} & \boldres{\scalebox{0.78}{0.227}} & \boldres{\scalebox{0.78}{0.253}} &\secondres{\scalebox{0.78}{0.233}} &\secondres{\scalebox{0.78}{0.261}} & \scalebox{0.78}{0.359} & \scalebox{0.78}{0.356}& \scalebox{0.78}{0.267} & \scalebox{0.78}{0.310} &\scalebox{0.78}{0.734} &\scalebox{0.78}{0.725} &\scalebox{0.78}{0.339} &\scalebox{0.78}{0.416} &\scalebox{0.78}{0.296} &\scalebox{0.78}{0.318} &\scalebox{0.78}{0.320} &\scalebox{0.78}{0.398} &\scalebox{0.78}{0.280} &\scalebox{0.78}{0.380} &\scalebox{0.78}{0.285} &\scalebox{0.78}{0.380} &\scalebox{0.78}{0.254} &\scalebox{0.78}{0.272} &\scalebox{0.78}{0.834} &\scalebox{0.78}{0.692} \\ 
    
    & \scalebox{0.78}{336} & \boldres{\scalebox{0.78}{0.246}} & \secondres{\scalebox{0.78}{0.284}} &\secondres{\scalebox{0.78}{0.248}} &\boldres{\scalebox{0.78}{0.273}} & \scalebox{0.78}{0.397} & \scalebox{0.78}{0.369}& \scalebox{0.78}{0.290} & \scalebox{0.78}{0.315} &\scalebox{0.78}{0.750} &\scalebox{0.78}{0.735} &\scalebox{0.78}{0.368} &\scalebox{0.78}{0.430} &\scalebox{0.78}{0.319} &\scalebox{0.78}{0.330} &\scalebox{0.78}{0.353} &\scalebox{0.78}{0.415} &\scalebox{0.78}{0.304} &\scalebox{0.78}{0.389} &\scalebox{0.78}{0.282} &\scalebox{0.78}{0.376} &\scalebox{0.78}{0.290} &\scalebox{0.78}{0.296} &\scalebox{0.78}{0.941} &\scalebox{0.78}{0.723} \\ 
    
    & \scalebox{0.78}{720} & \boldres{\scalebox{0.78}{0.247}} & \secondres{\scalebox{0.78}{0.276}} &\secondres{\scalebox{0.78}{0.249}} &\boldres{\scalebox{0.78}{0.275}} & \scalebox{0.78}{0.397} & \scalebox{0.78}{0.356} &\scalebox{0.78}{0.289} &\scalebox{0.78}{0.317} &\scalebox{0.78}{0.769} &\scalebox{0.78}{0.765} &\scalebox{0.78}{0.370} &\scalebox{0.78}{0.425} &\scalebox{0.78}{0.338} &\scalebox{0.78}{0.337} &\scalebox{0.78}{0.356} &\scalebox{0.78}{0.413} &\scalebox{0.78}{0.308} &\scalebox{0.78}{0.388} &\scalebox{0.78}{0.357} &\scalebox{0.78}{0.427} &\scalebox{0.78}{0.285} &\scalebox{0.78}{0.295} &\scalebox{0.78}{0.882} &\scalebox{0.78}{0.717} \\ 
    
    \cmidrule(lr){2-26}
    
    & \scalebox{0.78}{Avg} & \boldres{\scalebox{0.78}{0.226}} & \boldres{\scalebox{0.78}{0.261}} &\secondres{\scalebox{0.78}{0.233}} &\secondres{\scalebox{0.78}{0.262}} & \scalebox{0.78}{0.369} & \scalebox{0.78}{0.356} &\scalebox{0.78}{0.270} &\scalebox{0.78}{0.307} &\scalebox{0.78}{0.641} &\scalebox{0.78}{0.639} &\scalebox{0.78}{0.347} &\scalebox{0.78}{0.417} &\scalebox{0.78}{0.301} &\scalebox{0.78}{0.319} &\scalebox{0.78}{0.330} &\scalebox{0.78}{0.401} &\scalebox{0.78}{0.282} &\scalebox{0.78}{0.375} &\scalebox{0.78}{0.291} &\scalebox{0.78}{0.381} &\scalebox{0.78}{0.261} &\scalebox{0.78}{0.381} &\scalebox{0.78}{0.885} &\scalebox{0.78}{0.711} \\ 
    
    \midrule
    
     \multicolumn{2}{c|}{\scalebox{0.78}{{$1^{\text{st}}$ Count}}} & \scalebox{0.78}{\boldres{17}} & \scalebox{0.78}{\boldres{17}} & \scalebox{0.78}{0} & \scalebox{0.78}{\secondres{2}} & \scalebox{0.78}{1} & \scalebox{0.78}{1} & \scalebox{0.78}{0} & \scalebox{0.78}{0} & \scalebox{0.78}{\secondres{2}} & \scalebox{0.78}{0} & \scalebox{0.78}{0} & \scalebox{0.78}{0} & \scalebox{0.78}{0} & \scalebox{0.78}{0} & \scalebox{0.78}{0} & \scalebox{0.78}{0} & \scalebox{0.78}{0} & \scalebox{0.78}{0} & \scalebox{0.78}{0} & \scalebox{0.78}{0} & \scalebox{0.78}{0} & \scalebox{0.78}{0} & \scalebox{0.78}{0} & \scalebox{0.78}{0} \\ 
    \bottomrule
  \end{tabular}
  }
\end{table*}

\subsection{Frequency-wise Summarization}
Given the learned features of the sub-frequencies $\mathbf{W}' = \{{\vw}'_1, {\vw}'_2, \ldots, {\vw}'_N\}$ of the historical time series $\mathbf{X}$, the frequency-wise summarizing operation contains linear projections and IDFT:
$$
    \mathbf{X}' = \text{IDFT}(\mathbf{A}')
    \quad \mathbf{A}' = \text{Linear}(\mathbf{W}')
$$
where $\mathbf{X}' \in \mathbb{R}^{C \times H}$ is the final output of the framework.

\begin{table}[t]
    \caption{Benchmark dataset summary}
    \centering
    \resizebox{0.95\linewidth}{!}{
    \begin{tabular}{c|ccccccccc}
    \toprule
        Datasets  &  Weather  &  Electricity & ETTh1 & ETTh2 & ETTm1 & ETTm2 & Solar & Traffic \\ 
        \#Channel & 21 & 321 & 7 & 7 & 7 & 7 & 137 & 862 \\ 
        \#Timesteps  & 52969 & 26304 & 17420 & 17420 & 69680 & 69680 & 52179  & 17544 \\ \bottomrule
    \end{tabular}}
    \label{tab:database}
\end{table}

\section{EXPERIMENTS}\label{sec: Experiments}

\subsection{PROTOCOLS}\label{subsec: protocols}

\noindent\textbf{- Datasets.} 
We conduct extensive experiments on eight real-world benchmark datasets: Weather, four ETT datasets (ETTh1, ETTh2, ETTm1, ETTm2), Electricity (ECL), Traffic, and the Solar-Energy dataset \citep{long_and_short_SiGIR18}, with all datasets being published in \citep{liu2023itransformer}\footnote{https://github.com/thuml/iTransformer}. 
The information these datasets provide is summarized in Table \ref{tab:database}.
And the full results of four selected datasets* will be shown in Figure \ref{tab:part_baseline_results}.
and further details are available in Appendix \ref{appendix:datasets}.

\noindent\textbf{- Baselines.}  
We select 11 SOTA baseline studies.
Since we are focusing on Transformer, we first add seven proposed Transformer-based baselines, including 
iTransformer \citep{liu2023itransformer}, PatchTST \citep{PatchTST}, 
Crossformer \citep{Crossformer}, 
Stationary \citep{Non-stationary}, 
Fedformer \citep{ICMLFedformer}, 
Pyraformer \citep{pyraformer2022},
Autoformer \citep{Autoformer}.
We also add 2 MLP-based and 2 TCN-based methods, including
RLinear \cite{li2023RLinear}, 
DLinear \citep{DLinear}, 
TiDE \citep{das2023TiDE}, 
TimesNet \citep{Timesnet}.

\begin{table}[H]
    \centering
    \caption{The average forecasting accuracy (MSE) on ETTh1 dataset under 4 patch length settings.}
    \resizebox{0.4\textwidth}{!}{
    \begin{tabular}{c|c|c|c|c} 
        Patch length & 8 & 16 & 32 & Non \\ 
        \hlineB{2.5}
        MSE           & \textbf{0.417}        & 0.425         & 0.440       &  0.449    \\
    \end{tabular}
    }
    \label{tab:patchlength}
\end{table}


\begin{table}[H]
    \centering
    \caption{Averaged results for each setting in the ablation study. "No-CW" refers to removing channel-wise attention, and "No-FR" refers to removing frequency refinement.}
    \resizebox{0.49\textwidth}{!}{
    \begin{tabular}{c|cc|cc|cc}
    \multirow{2}{*}{Setting} & \multicolumn{2}{c|}{\textbf{Full}} & \multicolumn{2}{c|}{\textbf{No-CW}} & \multicolumn{2}{c}{\textbf{No-FR}} \\ 
                             & MSE   & MAE   & MSE    & MAE    & MSE    & MAE    \\ 
                             \hlineB{2.5}
    ETTm1                    & \textbf{0.384} & \textbf{0.396} & 0.418  & 0.419  & 0.539  & 0.485  \\
    Weather                  & \textbf{0.246} & \textbf{0.273} & 0.262  & 0.290  & 0.293  & 0.322
    \end{tabular}
    }
    \label{tab:ablationstudy}
\end{table}

\noindent\textbf{- Setup and Evaluation.} 
All baselines use the same prediction length with $H \in  \{96, 192, 336, 720\}$ for all datasets.
The look-back window $L$ = 96 was used in our setting for fair comparisons, referring to \citep{ICMLFedformer, liu2023itransformer}.
We used MSE and MAE as the forecasting metrics.
We further analyzed the forecasting results between the model outputs and the ground truth in the time and frequency domains.
Using heatmaps, we tracked the way in which $\Delta_{k}$ changes during training to show the debiased results of \method compared with various SOTA baselines.


\subsection{Results}\label{subsec: MainResults}

\textbf{Forecasting Results.}
Table~\ref{tab:main_result} shows the average forecasting performance across four prediction lengths.
The best results are highlighted in red, and the second-best in blue. 
With a default look-back window of $L = 96$, our approach realizes leading performance levels on most datasets, securing 14 top-1 and 2 top-2 positions across two metrics over eight datasets. 
More detailed results for 4 of the eight datasets are shown in Table~\ref{tab:part_baseline_results}, where our method achieves 34 top-1 and 6 top-2 rankings out of 40 possible outcomes across the four prediction lengths. 
More comprehensive results regarding the different prediction length settings on all datasets and the impact of extending the look-back window are detailed in Appendix~\ref{appendix:lookbackwindow} and~\ref{appendix:fullresult}.

\begin{figure*}[htb]
\centering
\includegraphics[width=\linewidth]{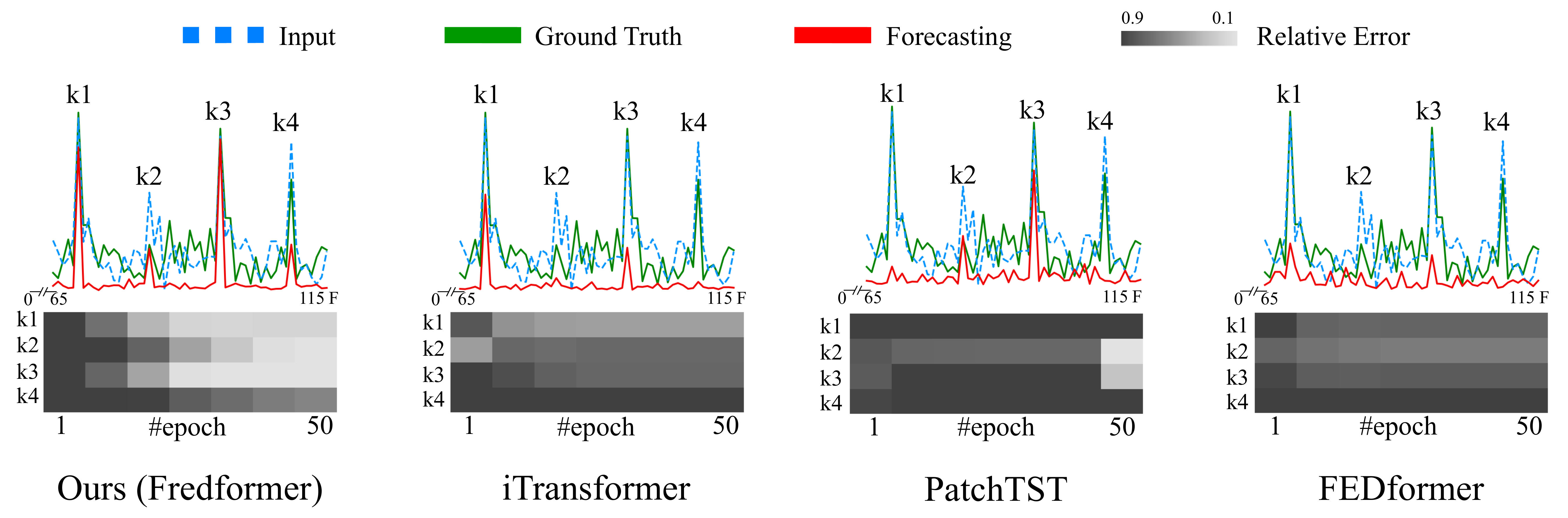}
\caption{Visualizations of the learning dynamics and results for \method and baselines on the ETTh1 dataset, employing line graphs to illustrate amplitudes in the frequency domain and heatmaps to represent training epoch errors.}
\label{fig: result_fre}
\end{figure*}

\noindent\textbf{Frequency Bias Evaluation.}
Figure~\ref{fig: result_fre} is a case study visualization in the frequency domain, i.e., the DFT plot.
The input, forecast output, and ground truth data series are shown in blue, red, and green, respectively.
Similar to Section~\ref{subsec: CaseStudies}, the heat map shows the relative error for four selected mid-to-high frequency components over increasing epochs.
After training, \method accurately identifies $k1$, $k2$, and $k3$, with uniformly decreasing relative errors. Despite a larger learning error for $k4$, $\Delta_{k4}$ consistently diminishes. 
This performance contrasts with all the baselines, demonstrating a lack of effectiveness in capturing these frequency components, with unequal reductions in relative errors.
In contrast, PatchTST demonstrates a sudden improvement in component accuracy ($k2$,$k3$) during the final stages of training.
FEDformer fails to capture these frequency components, possibly because its strategy of selecting and learning weights for only a random set of $k$ components overlooks all unselected components.
Notably, iTransformer overlooks mid-to-high frequency features, partially learning components $k1$ and $k3$ while ignoring $k2$ and $k4$, indicating a clear frequency bias. 
This may stem from its use of channel-wise attention alongside global normalization in the time domain, as discussed in Lemma~\ref{lemma:norm} and further supported by our ablation study~\ref{sec:ablation}. 
This highlights the effectiveness of frequency refinement and normalization.

\subsection{Ablation Study}\label{subsec: AblationStudy}
\label{sec:ablation}

\textbf{Channel-wise Attention and Frequency Refinement.}
We evaluate the effectiveness of channel-wise attention and frequency refinement.
To this end, we remove each component by ablation and compare it with the original \method.
Table~\ref{tab:ablationstudy} shows that our method consistently outperforms others in all experiments, highlighting the importance of integrating channel-wise attention with frequency local normalization in our design. Interestingly, employing frequency local normalization alone yields better accuracy than channel-wise attention alone. 
This suggests that minimizing proportional differences in amplitudes across various key frequency components is crucial for enhancing accuracy.

\noindent\textbf{Effect of Patch Length.}
\label{sec:patchlength}
This ablation evaluates the impact of patch length using the ETTh1 dataset. 
We conduct four experiments with $S = [8, 16, 32, 48]$ patch lengths and corresponding patch numbers $N = [6, 3, 2, 1]$. 
In this context, $N = 1$ means frequency normalization and channel-wise attention are applied to the entire spectrum without a patching operation. 
Table \ref{tab:patchlength} shows the forecasting accuracy for each setting.
As the patch length increases, the granularity of the frequency features extracted by the model becomes coarser, decreasing forecasting accuracy.


\begin{figure}[t]
\includegraphics[width=0.98\linewidth]{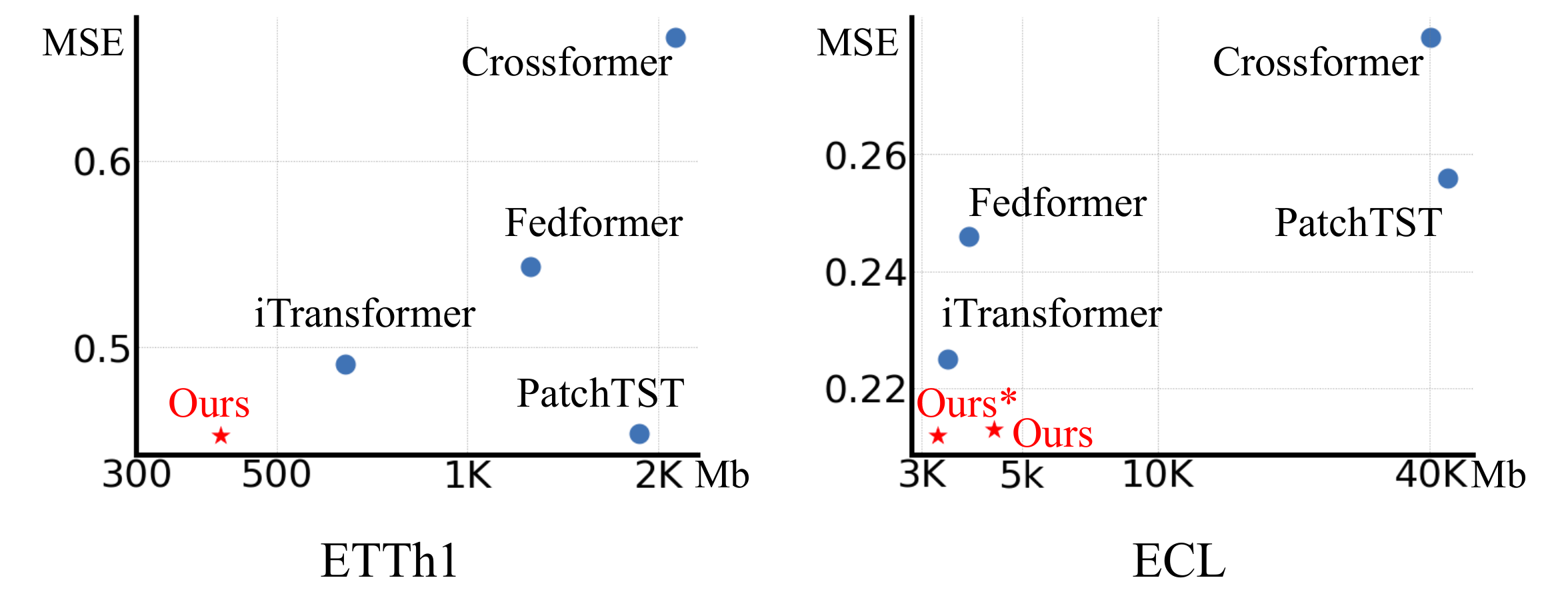}
\caption{This figure compares prediction accuracy and computational complexity (VRAM usage) among Transformer-based methods, \method (Ours), and its optimized variant, Nyström-\method(Ours*).}
\label{fig: result_nystrom}
\end{figure}

\begin{table}[t]
    \centering
    \begin{tabularx}{\linewidth}{c|ccc}
        Method & FEDformer & PatchTST & Crossformer \\ 
        \cmidrule(lr){2-4}
        & iTransformer & \textbf{Ours} & \textbf{Ours*(Nyström)} \\
        \hlineB{1.5}
        \noalign{\vskip 0.1cm}
        Complexity & $O(LC)$ & $O\left(\frac{L^2}{P^2}C\right)$ & $O\left(\frac{L^2}{P^2}C\right)$ \\
        \cmidrule(lr){2-4}
        & $O(C^2)$ & $O\left(\frac{L}{P}C^2\right)$ & $O\left(\frac{L}{P}C\right)$ \\
    \end{tabularx}
    \caption{The theoretical computational complexity of Transformer-based methods.}
\end{table}

\subsection{Discussion of Applicability}\label{subsec: Nystrom}

Beyond algorithmic considerations, we further discuss the practical deployment of \method in real-world scenarios, with the primary challenge being memory consumption during model training. 
The $O(n^2)$ complexity of self-attention limits the use of longer historical time series for forecasting, generating the need for innovations to reduce computational demands \citep{logtrans2019, Informer, PatchTST}. 
Through patching operations, we decrease the complexity from $O(LC^2)$ to $O(\frac{L}{P}C^2)$. 
However, our channel-wise attention increases the computational costs with the number of channels, potentially limiting practical applicability with many channels.
To address this, we propose a lightweight \method, inspired by NyströmFormer \citep{xiong2021nystromformer}, which applies a matrix approximation to the attention map.
This design allows us to further reduce our complexity to $O( \frac {L}{P}C )$ without the need to modify the feature extraction (attention computation) or the data stream structure within the Transformer, unlike with previous methods \citep{Informer,pyraformer2022, Autoformer, ICMLFedformer}.
Figure \ref{fig: result_nystrom} shows a tradeoff between the model efficiency (VRAM usage) and accuracy in our method and the baselines. 
The plain \method achieves high accuracy with low computational costs with fewer channels, such as ETTh1 with 7 channels. 
However, as shown in the ECL dataset (321 channels), the computational costs increase while maintaining high accuracy as the channel number increases.
Here, Nyström-\method further reduces computational requirements without compromising accuracy (the right sub-figure), showing that our model can realize computational efficiency and forecasting accuracy.
Further details and derivations are provided in Appendix \ref{append:Nyström}.




\section{Related Works}\label{sec: 5}

\textbf{Transformer for Time Series Forecasting.}
Forecasting is important in time series analysis \citep{tsdata_1, hamilton2020time}.
Transformer has significantly progressed in time series forecasting\citep{PatchTST, Crossformer, pdformer}. 
Earlier attempts focused on improving the computational efficiency of Transformers for time series forecasting tasks\citep{longformer, Informer, pyraformer2022}.
Several studies have used Transformers to model inherent temporal dependencies in the time domain of time series\citep{logtrans2019, Informer, pyraformer2022, PatchTST, liu2023itransformer}. 
Various studies have integrated frequency decomposition and spectrum analysis with the Transformer in modeling temporal variations \citep{Autoformer, Etsformer} to improve the capacity for temporal-spatial representation.
In \citep{ICMLFedformer}, attention layers are designed that directly function in the frequency domain to enhance spatial or frequency representation.

\noindent\textbf{Modeling Short-Term Variation in Time Series.}
Short-term variations are intrinsic characteristics of time series data and play a crucial role in effective forecasting \citep{app_elec, Non-stationary}. Numerous deep learning-based methods have been proposed to capture these transient patterns \citep{RNN_2, LSTM_2, RNN_3, CNN_2, CNN_1, RobustSTL, Autoformer, CoST, LaST}. Here, we summarize some studies closely aligned with our proposed method. Pyraformer \citep{pyraformer2022} applies a pyramidal attention module with inter-scale and intra-scale connections to capture various temporal dependencies. FEDformer \citep{ICMLFedformer} incorporates a Fourier spectrum within the attention computation to identify pivotal frequency components. Beyond Transformers, TimesNet \citep{Timesnet} employs Inception blocks to capture intra-period and inter-period variations.

\noindent\textbf{Channel-wise Correlation.}
Understanding the cross-channel correlation is also critical for time series forecasting.
Several studies aimed to capture intra-channel temporal variations and model the inter-channel correlations using Graph Neural Networks (GNNs) \citep{MTGNN, StemGNN}. 
Recently, Crossformer \citep{Crossformer} and iTransformer \citep{liu2023itransformer} both adopted channel-wise Transformer-based frameworks,
and extensive experimental results have demonstrated the effectiveness of channel-wise attention for time series forecasting.

\section{Conclusion}\label{sec: 6}

In this paper, we first empirically analyzed frequency bias, delving into its causes and exploring debiasing strategies. 
We then provided a formulation of this bias based on our analytical insights.
We proposed the \method framework with three critical designs to tackle this bias and thus ensure unbiased learning across frequency bands. 
Our extensive experiments across eight datasets confirmed the excellent performance of our proposed method.
Visual analysis confirmed that our approach effectively mitigates frequency bias.
The model analysis further illustrated how our designs aid frequency debiasing and offered preliminary guidelines for future model design. 
Additionally, a lightweight variant of our model addresses computational efficiency, facilitating practical application.

\section{ACKNOWLEDGMENTS}

We thank anonymous reviewers for their insightful comments and discussions. 
This work is supported by
JSPS KAKENHI Grant-in-Aid for Scientific Research Number
JP21H03446,     
JP23K16889,    
JP24K20778,    
NICT JPJ012368C03501, 
JST-AIP JPMJCR21U4,  
JST-CREST JPMJCR23M3,  
JST-RISTEX JPMJRS23L4.  

\bibliographystyle{ACM-Reference-Format}
\bibliography{references}

\clearpage
\newpage
\appendix

\twocolumn[{
\begin{adjustwidth}{1cm}{1cm}

\begin{center}
{\LARGE \textbf{Fredformer: Frequency Debiased Transformer for Time Series Forecasting\\ $\;$ \\ ————Appendix————}}
\end{center}

\vspace{1cm}

\begin{center}
  \hypersetup{hidelinks}
  \Large
  \tableofcontents
  \noindent\hrulefill
\end{center}

\end{adjustwidth}
}]

\newpage
\clearpage

\newpage

\textit{The full appendix can be found at: http://arxiv.org/abs/2406.09009}

\section{Details of the case studies}
\label{app:case_study_details}
Here, we illustrate the details of how we generated the data for case study 2 in Sec.\ref{subsec: CaseStudies.2}:
The generation of data for Case Study 2 from the original time series involves a sequence of steps to emphasize certain frequency components by manipulating their positions in the frequency domain. This process not only constructs a dataset with distinct frequency characteristics but also preserves the inherent noise and instability of the real data, enhancing the robustness and credibility of subsequent analyses.
Specifically, the steps are as follows:

\begin{enumerate}
    \item Apply the Discrete Fourier Transform (DFT) to the original time series data to obtain its frequency components, excluding columns irrelevant for Fourier analysis (e.g., dates).
    \item Select four prominent low-frequency components from the entire frequency spectrum and move them to the mid-frequency part. This modification aims to reduce the impact of frequency bias typically seen between low and high frequencies by placing important components in a non-low and non-high frequency position.
    \item Split the frequency components into three equal parts.
    \item Rearrange these parts according to a predefined order for frequency emphasis, ensuring that the first part is moved to the end while keeping the original second and third parts in their order.
    \item Apply the Inverse Discrete Fourier Transform (IDFT) to the rearranged frequency data to convert it back into the time domain, thereby generating the modified "mid" frequency data.
    \item Reinsert any excluded columns (e.g., dates) to maintain the original structure of the data.
\end{enumerate}

\noindent Through the operations described above, we have constructed a dataset with clearly high amplitude frequency components in the middle of the frequency domain. By moving significant low-frequency components to the mid-frequency section, we aim to mitigate the effects of frequency differences that arise from the dominance of low and high frequencies. The advantage of creating artificial data through these simple modifications to real data lies in its ability to preserve the inherent noise and instability present in the real data, thereby enhancing the robustness and credibility for subsequent analysis.

\begin{algorithm}
\caption{Generation of Modified Frequency Data for Case Study 2}
\begin{algorithmic}[1]
\label{app:case_study_2}
\Procedure{GenerateModifiedFrequencyData}{$data$}
    \State $dateColumn \gets data['date']$ \Comment{Preserve the date column for reinsertion}
    \State $dataValues \gets data.drop(columns=['date'])$ \Comment{Exclude non-relevant columns}
    \State $fftData \gets$ DFT($dataValues$) \Comment{Apply DFT to obtain frequency components}
    \State $selectedComponents \gets$ SelectLowFrequencyComponents($fftData$, 4) \Comment{Select 4 prominent low-frequency components}
    \State $fftDataModified \gets$ MoveComponentsToMid($fftData$, $selectedComponents$) \Comment{Move selected components to mid-frequency}
    \State $parts \gets$ Split($fftDataModified$, 3) \Comment{Split frequency components into three equal parts}
    \State $order \gets$ DefineOrderForFrequencyEmphasis() \Comment{Define a new order for rearrangement}
    \State $rearrangedParts \gets$ Rearrange($parts$, $order$) \Comment{Rearrange parts according to the predefined order}
    \State $modifiedFrequencyData \gets$ IDFT($rearrangedParts$) \Comment{Apply IDFT to generate modified time domain data}
    \State $modifiedFrequencyData.insert(0, 'date', date column)$ \Comment{Reinsert the date column}
    \State \textbf{return} $modifiedFrequencyData$
\EndProcedure
\end{algorithmic}
\end{algorithm}

\section{More Details of the datasets}
\label{appendix:datasets}
Weather contains 21 channels (e.g., temperature and humidity) and is recorded every 10 minutes in 2020. 
ETT \citep{Informer} (Electricity Transformer Temperature) consists of two hourly-level datasets (ETTh1, ETTh2) and two 15-minute-level datasets (ETTm1, ETTm2). 
Electricity \citep{Electricity}, from the UCI Machine Learning Repository and preprocessed by, is composed of the hourly electricity consumption of 321 clients in kWh from 2012 to 2014. 
Solar-Energy \citep{long_and_short_SiGIR18} records the solar power production of 137 PV plants in 2006, sampled every 10 minutes.
Trafﬁc contains hourly road occupancy rates measured by 862 sensors on San Francisco Bay area freeways from January 2015 to December 2016.
More details of these datasets can be found in Table.\ref{tab:dataset_overview}.

\begin{table*}[h!t]
\centering
\caption{Overview of Datasets}
\label{tab:dataset_overview}
\begin{tabular}{@{}lllll@{}}
\toprule
Dataset & Source & Resolution & Channels & Time Range \\ \midrule
Weather & Autoformer\citep{Autoformer} & Every 10 minutes & 21 (e.g., temperature, humidity) & 2020 \\
ETTh1 & Informer\citep{Informer} & Hourly & 7 states of a electrical transformer & 2016-2017 \\
ETTh2 & Informer\citep{Informer} & Hourly & 7 states of a electrical transformer & 2017-2018 \\
ETTm1 & Informer\citep{Informer} & Every 15 minutes & 7 states of a electrical transformer & 2016-2017 \\
ETTm2 & Informer\citep{Informer} & Every 15 minutes & 7 states of a electrical transformer & 2017-2018 \\
Electricity & UCI ML Repository & Hourly & 321 clients' consumption & 2012-2014 \\
Solar-Energy & \citep{long_and_short_SiGIR18} & Every 10 minutes & 137 PV plants' production & 2006 \\
Traffic & Informer\citep{Informer} & Hourly & 862 sensors' occupancy & 2015-2016 \\ \bottomrule
\end{tabular}
\end{table*}

\section{\method Algorithm}
\label{appendix:algorithm1}

The algorithm \ref{algorithm} outlines our overall procedure. 
It includes several parts: 
(i) DFT-to-IDFT Backbone, where the input data is transformed using DFT and segmented into frequency bands; 
(ii \& iii) Frequency Local Independent Learning, where normalization and a Transformer are applied to learn dependencies and features across channels; 
and (iv) Frequency-wise Summarizing, where the processed frequency information is summarized and transformed back to the time domain using IDFT to obtain the forecasting result.

\begin{algorithm}[t]
\caption{\method}
\begin{algorithmic}[1]
\State \textbf{Input:} Historical data $\mathbf{X} \in \mathbb{R}^{C \times L}$, where $C$ is the number of channels and $L$ is the length of the data series.
\State \textbf{Output:} Forecasting result $\mathbf{X}'$

\State \textbf{Procedure:}

\State  \textit{$\rm(\hspace{.18em}i\hspace{.18em})$DFT-to-IDFT Backbone}
\State Perform DFT on $\mathbf{X}$ to obtain $\mathbf{A} \in \mathbb{R}^{C \times L}$
\For{channel $c = 1$ to $C$}
    \For{frequency band $n = 1$ to $N$}
        \State Segment $\mathbf{A}_c$ into $N$ bands: $(\mathbf{W}^{(1:C)}_{1},\ldots,\mathbf{W}^{(1:C)}_{N})$
    \EndFor
\EndFor

\State  \textit{$\rm(\hspace{.18em}ii \& iii\hspace{.18em})$ Frequency Local Independent Learning}
\For{frequency band $n = 1$ to $N$}
    \State Normalize cross-channel amplitude sequences for band $n$: $\quad {\mathbf{W}^*_n} = \sigma({\mathbf{W}_n}) \quad n = 1, 2, \ldots, N$
    \State Apply Transformer to learn channel-wise dependencies and joint features across channels: $\quad {\vw'}^{(1:C)}_n = f_{Trans}({\vw^*}^{(1:C)}_n)$
\EndFor

\State \textit{$\rm(\hspace{.18em}iv\hspace{.18em})$Frequency-wise Summarizing}
\State Abstract overall frequency information to form new $\mathbf{A}'$:
$\quad \mathbf{A}' = \text{Linear}(\mathbf{W}')$
\State Perform IDFT using $\mathbf{A}'$ to generate $\mathbf{X}'$:
$\quad \mathbf{X}' = \text{IDFT}(\mathbf{A}')$
\end{algorithmic}
\label{algorithm}
\end{algorithm}

\section{Look-back window analysis}
\label{appendix:lookbackwindow}

\begin{figure}[t]
\includegraphics[width=\linewidth]{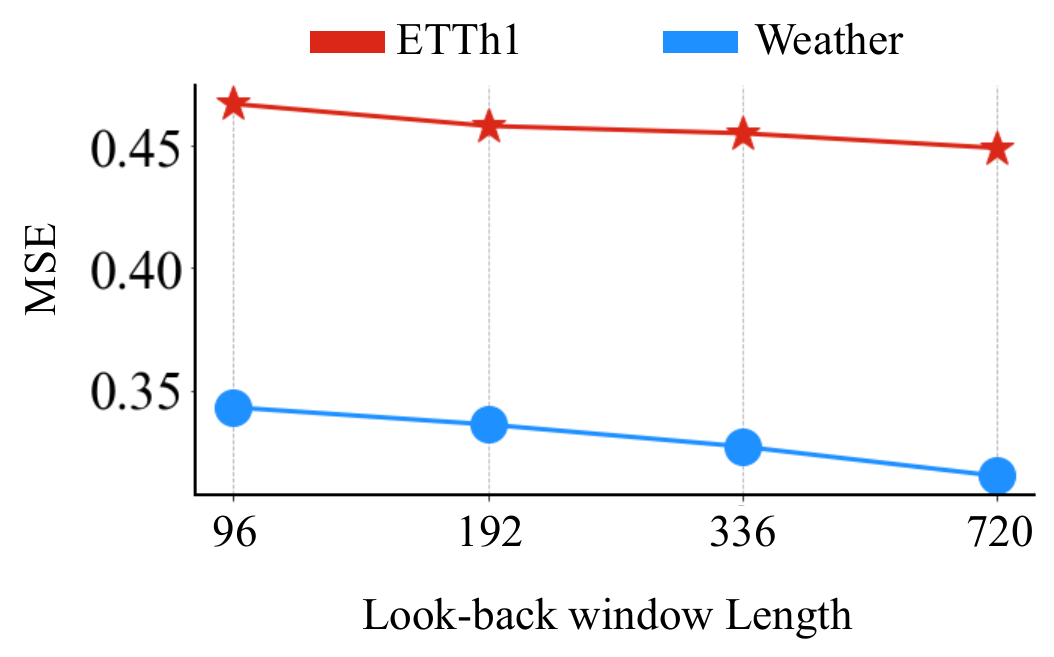}
\caption{The forecasting performance on two datasets, ETTh1 and Weather, across four different look-back window lengths. The x-axis indicates the window length, while the y-axis represents the MSE loss of the forecasting results.
}
\label{fig:lookback}
\end{figure}

We conducted further tests on our method using the ETTh1 and Weather datasets to investigate the impact of different look-back window lengths on forecasting accuracy. Four distinct lengths were chosen: \{96, 192, 336, 720\}, with 96 corresponding to the results presented in the main text and the other three lengths selected to compare the changes in forecasting accuracy with longer input sequences. Figure. \ref{fig:lookback} illustrates the variation in model forecasting accuracy across these input lengths. Overall, as the length of the input sequence increases, so does the model forecasting accuracy, demonstrating that our model is capable of extracting more features from longer input sequences. Specifically, comparing the longest window of 720 to the shortest of 96, the model forecasting accuracy improved by approximately 10\% ($0.343 \to 0.315$ for Weather and $0.467 \to 0.449$ for ETTh1).

\newpage

\section{Hyperparameter sensitivity}
\label{appendix:hyperparameter}

\begin{figure}[t]
\includegraphics[width=\linewidth]{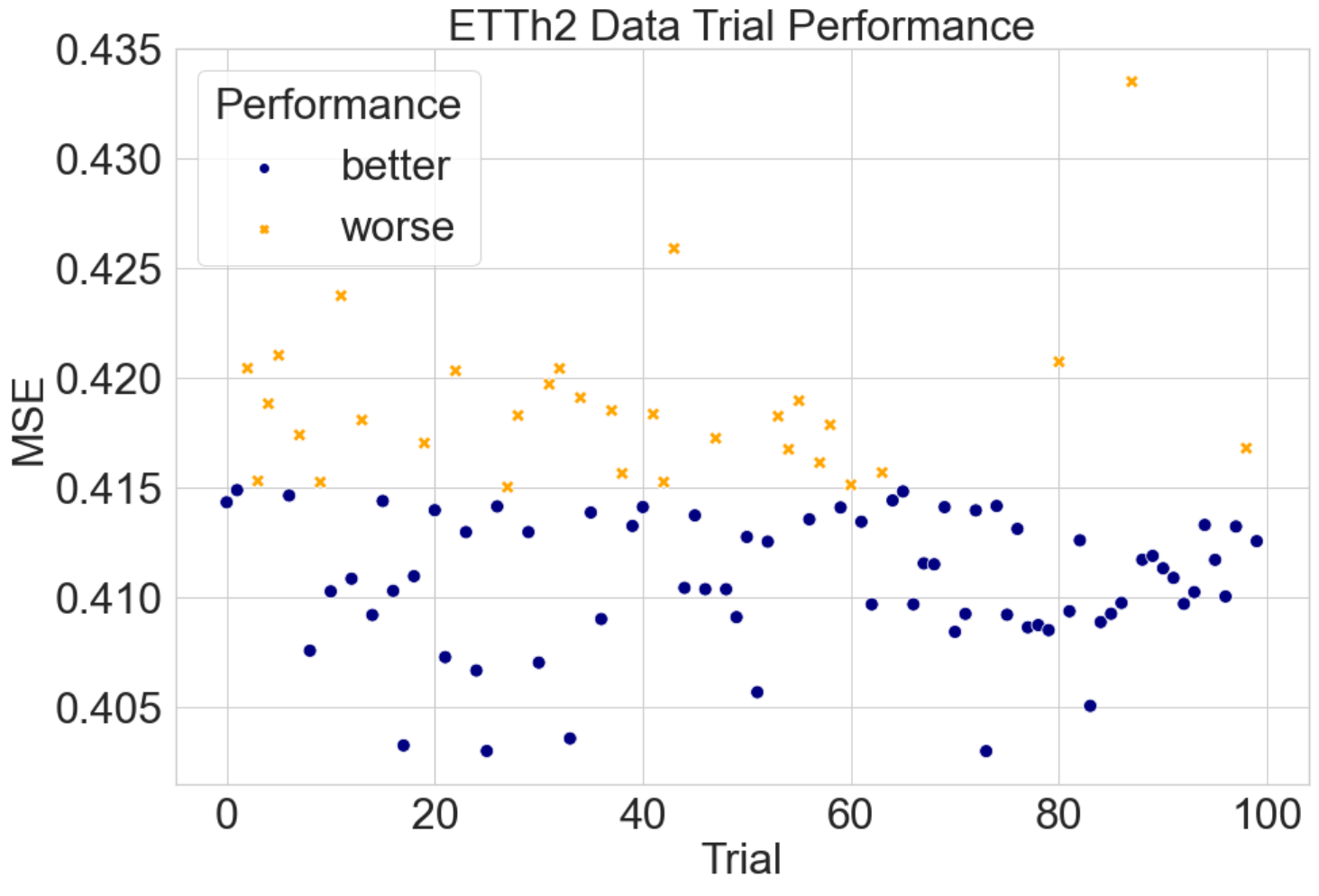}
\caption{Overall view of all trials, plotting different trials on the x-axis against MSE accuracy on the y-axis. Yellow dots represent trials with poor accuracy, falling below our final selection MSE accuracy, while purple dots indicate trials with higher accuracy than our final decision.}

\label{fig:app_trial}
\end{figure}

To evaluate our model robustness across various hyperparameter settings with input/predicting length $L = 96 / H = 720$, we investigated four key hyperparameters: (1) model depth (\texttt{cf depth}), (2) feature dimension of self-attention (\texttt{cf dim}), (3) feature dimension within self-attention multi-heads (\texttt{cf head dim}), (4) number of multi-heads (\texttt{cf heads}), and (5) feature dimension of the feed-forward layer in the Transformer Encoder (\texttt{cf mlp}). 
We tested one hundred hyperparameter combinations, with results shown in Figure E. 
The variation in the model accuracy, ranging from 0.433 to 0.400 with an average value of 0.415, shows our preference for stability in hyperparameter selection over chasing the highest possible accuracy. 
We decided to utilize the averaged accuracy as our benchmark result in Table~\ref{appendix:fullresult}.

\begin{figure}[t]
\includegraphics[width=\linewidth]{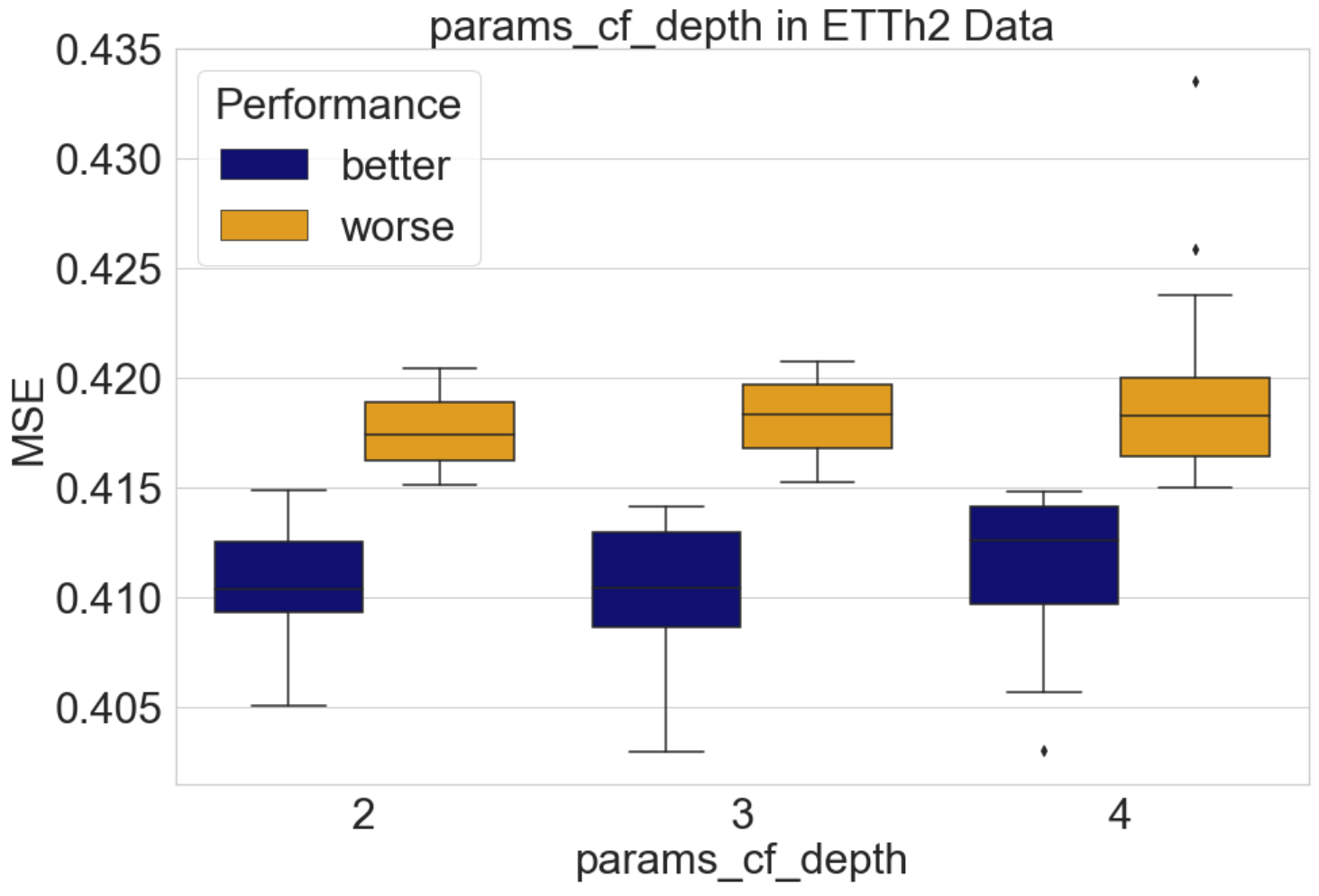}
\caption{Illustration of the \texttt{cf depth} parameter effect across all trials, showcasing MSE accuracy for different values. The x-axis represents the parameter values, while the y-axis displays MSE accuracy. Purple indicates trials with MSE accuracy higher than our final chosen result, and yellow signifies trials with lower accuracy. The box plot details the distribution of MSE accuracy above (in yellow) or below (in purple) our final selection for each value.}
\label{fig:app_depth}
\end{figure}

\begin{figure}[t]
\includegraphics[width=\linewidth]{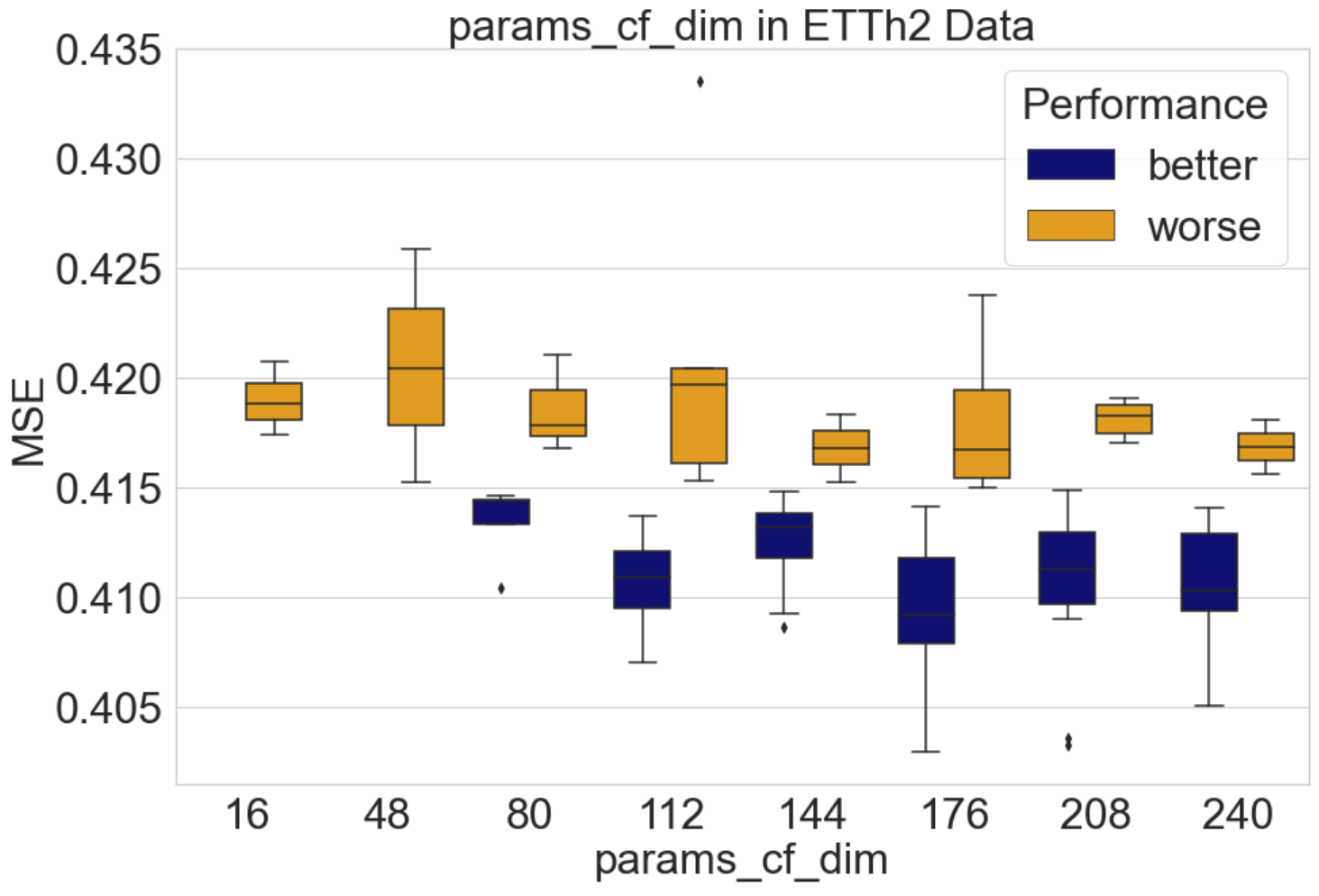}
\caption{Depiction of the \texttt{cf dim} parameter influence, detailing MSE accuracy across various settings. Parameter values are plotted on the x-axis against MSE accuracy on the y-axis. Trials surpassing our final result accuracy are marked in purple, whereas those falling short are in yellow. The distribution of MSE accuracies, differentiated by outcomes exceeding or not meeting our final accuracy, is presented in a box plot for each parameter value.}
\label{fig:app_dim}
\end{figure}

\begin{figure}[t]
\includegraphics[width=\linewidth]{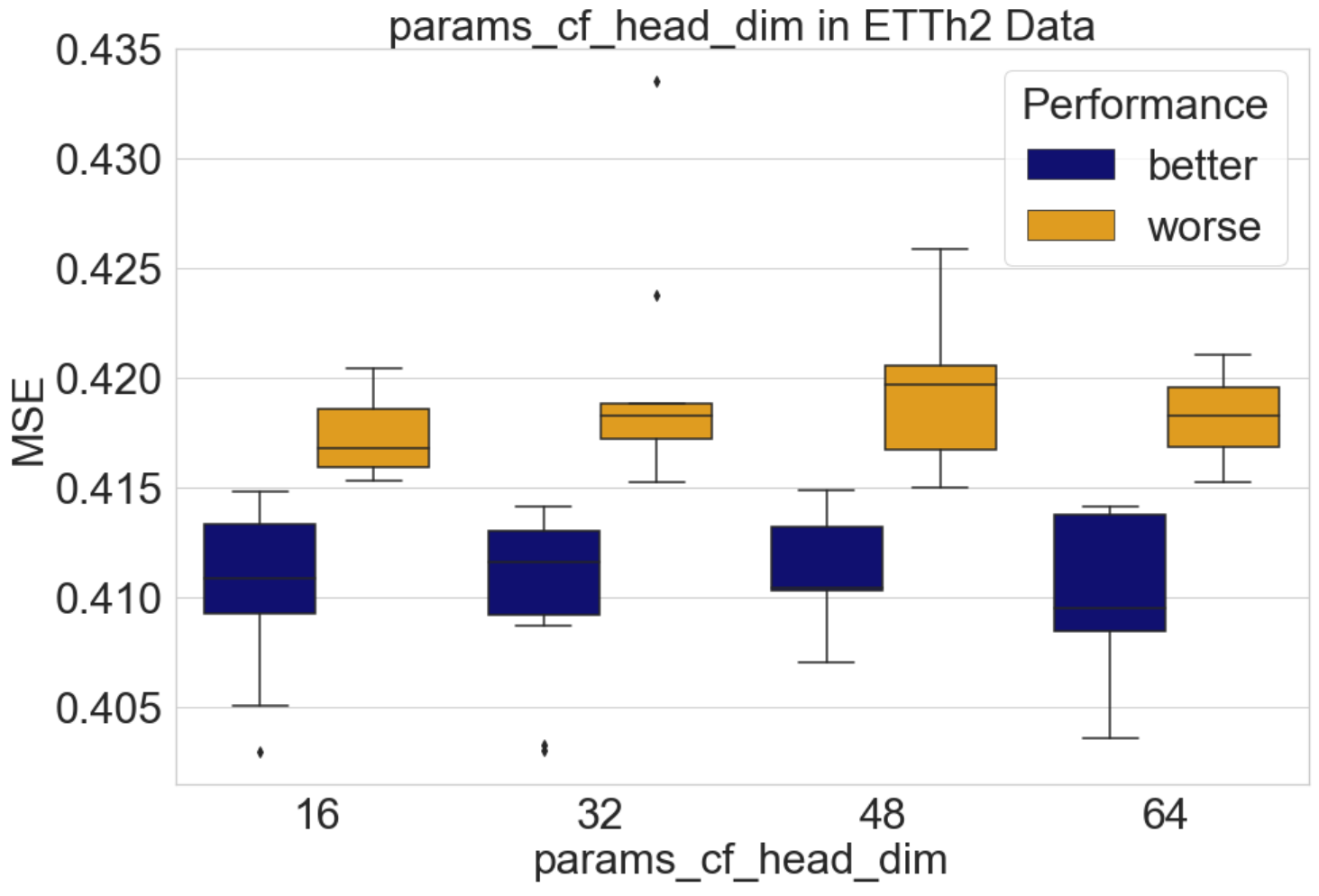}
\caption{Analysis of the \texttt{cf head dim} parameter impact, with the x-axis indicating parameter values and the y-axis MSE accuracy. Purple highlights trials where MSE accuracy is above our finalized result, with yellow showing lower accuracy trials. Each parameter value MSE accuracy distribution, categorized by exceeding or not our final accuracy, is visualized through a box plot.}
\label{fig:app_head_dim}
\end{figure}

\begin{figure}[t]
\includegraphics[width=\linewidth]{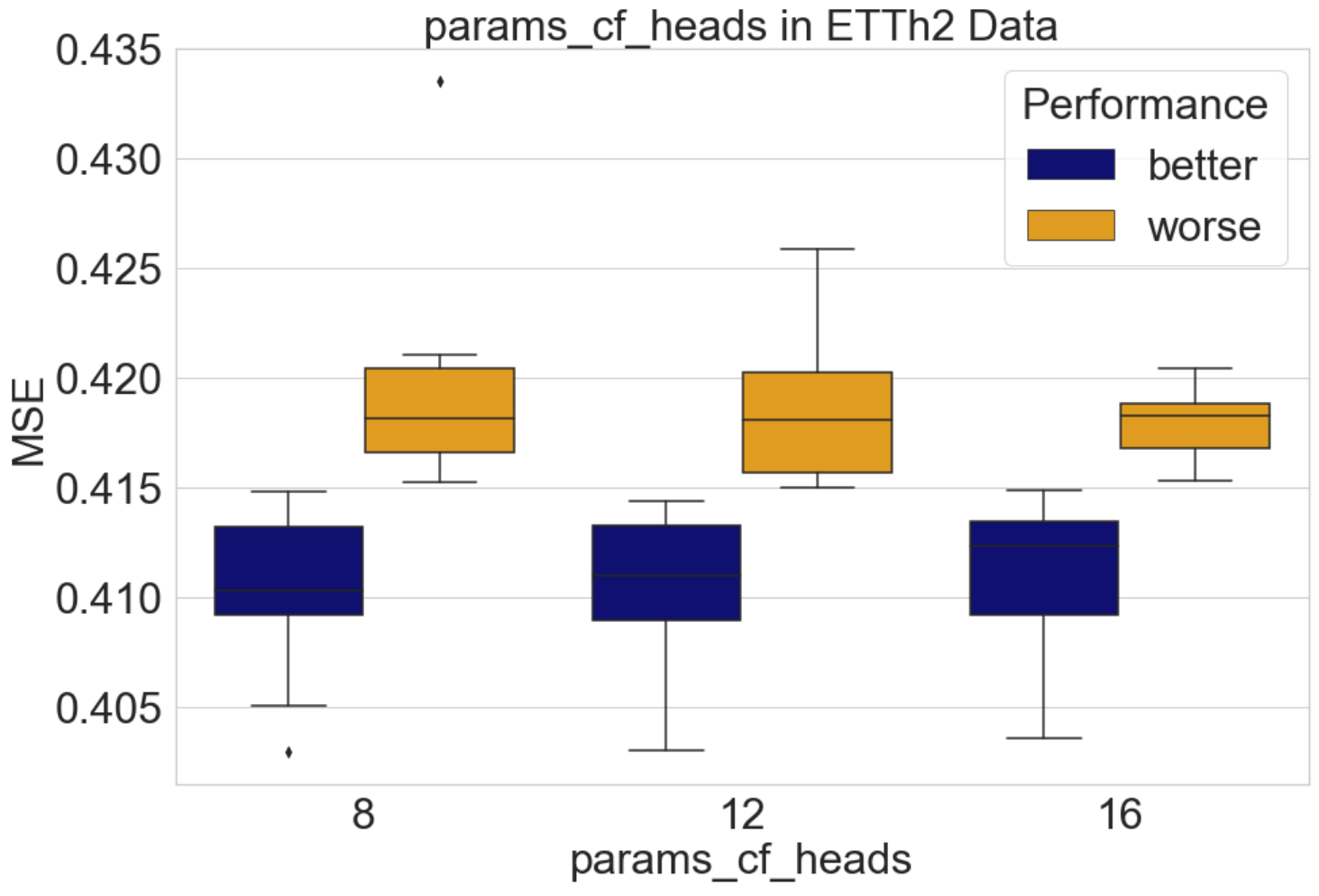}
\caption{Visualization of the \texttt{cf heads} parameter role, plotting parameter values against MSE accuracy. Trials with higher MSE accuracy than our chosen outcome are in purple; those with lower accuracy are in yellow. A box plot represents the spread of MSE accuracies for each value, segmented into results that are above (yellow) or below (purple) our final selection.}
\label{fig:app_heads}
\end{figure}

\begin{figure}[t]
\includegraphics[width=\linewidth]{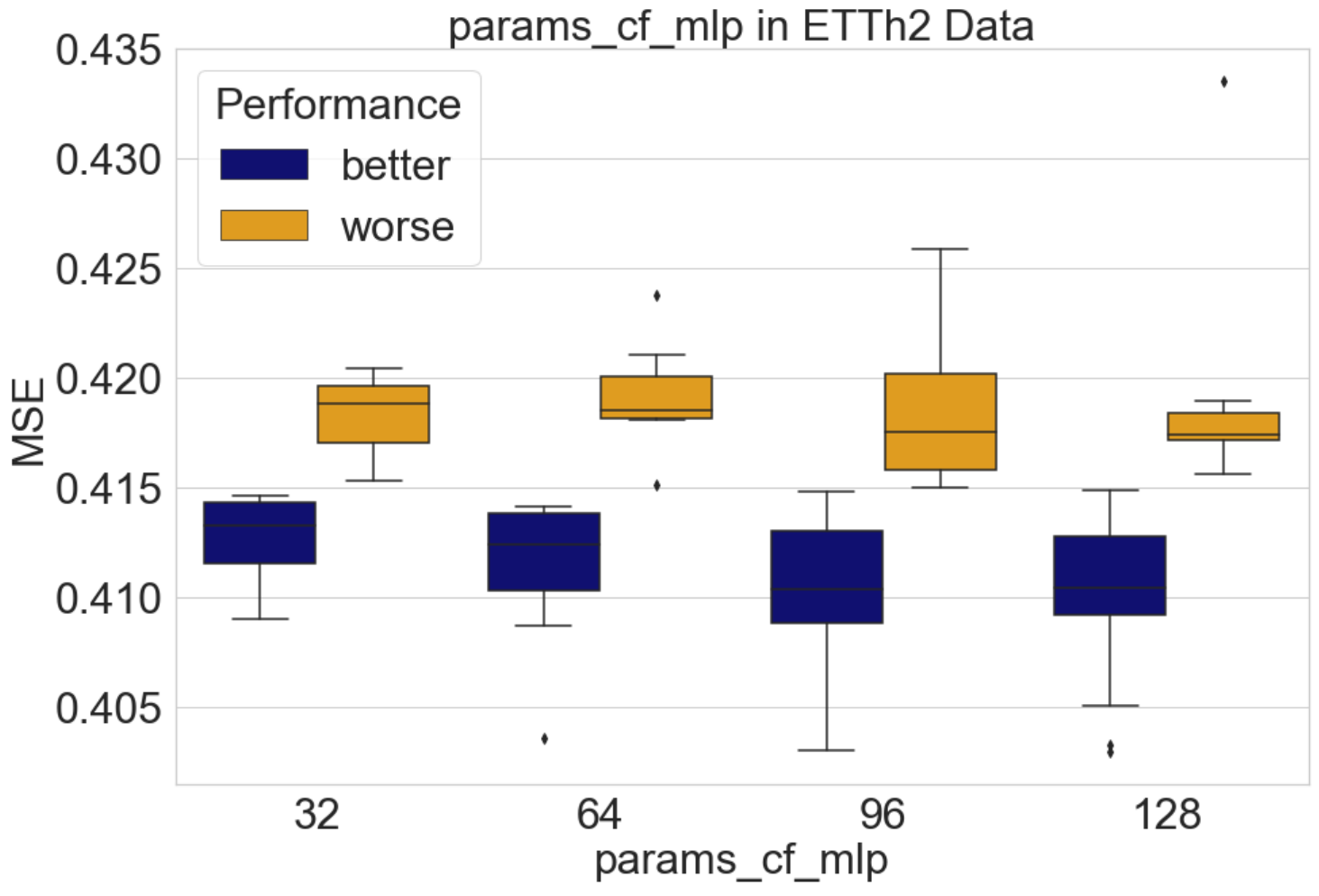}
\caption{Examination of the \texttt{cf mlp} parameter performance, with parameter values on the x-axis and MSE accuracy on the y-axis. Purple represents trials outperforming our final result in MSE accuracy, while Yellow indicates underperformance. The box plot shows the distribution of MSE accuracies for each parameter value, divided into outcomes that surpass or fall short of our final chosen accuracy.}
\label{fig:app_mlp}
\end{figure}

Visual representations of each hyperparameter impact on model robustness are detailed as follows:
\begin{itemize}
    \item Figure \ref{fig:app_depth} illustrates the robustness across different \texttt{cf depth} settings.
    \item Figure \ref{fig:app_dim} showcases the impact of varying \texttt{cf dim} on model performance.
    \item Figure \ref{fig:app_head_dim} presents the model behavior with changes in \texttt{cf head dim}.
    \item Figure \ref{fig:app_heads} depicts the influence of different \texttt{cf heads} counts.
    \item Figure \ref{fig:app_mlp} reveals how adjustments in \texttt{cf mlp} affect accuracy.
\end{itemize}

\section{Visualizations of the forecasting results}
\label{appendix:more_visualizations}

Due to space constraints and the length of the paper, we have omitted a significant number of visualization results in the main text. 
In the appendix, we provide additional visualization results to demonstrate the effectiveness of our method. 
Here, we divide the visualization results into two categories: (1) time-domain visualizations and (2) frequency-domain visualizations. 
These categories highlight two critical aspects of our model effectiveness: (1) the accuracy of predictions in the time domain and (2) the capability to capture important components in the frequency domain.

\subsection{Time domain}
\label{appendix:time_domain}
\begin{figure}[!ht]
\centering
\begin{minipage}{0.98\linewidth}
\includegraphics[width=\linewidth]{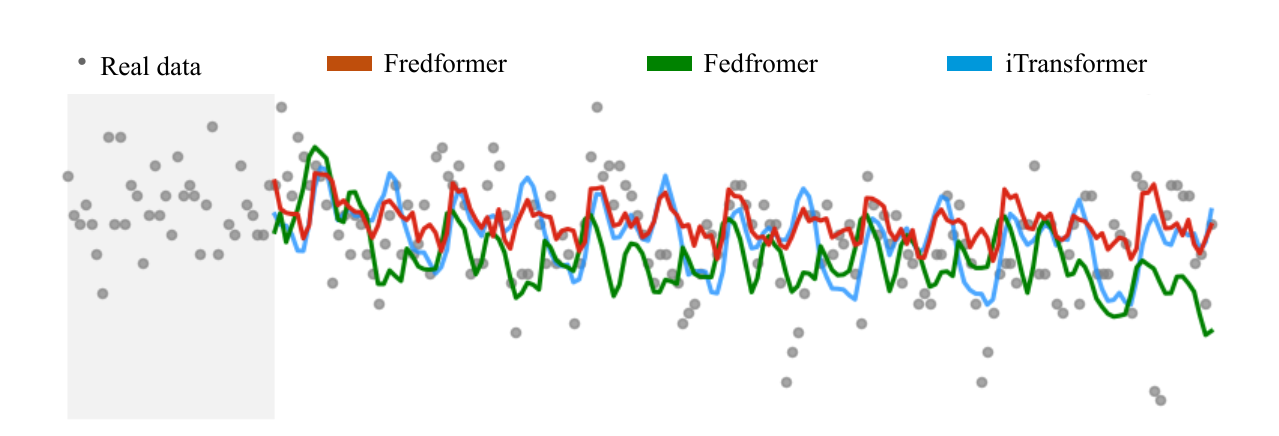}
\caption{Visualization of forecasting results ($H = 336$). Grey dots show the real data, and the grey zone represents the last part of the look-back window. Sample from channel \#5 of the ETTh1 dataset, illustrating the unique data characteristics of this channel.}
\label{fig: app_time1}
\end{minipage}

\vspace{1cm} 

\begin{minipage}{0.98\linewidth}
\includegraphics[width=\linewidth]{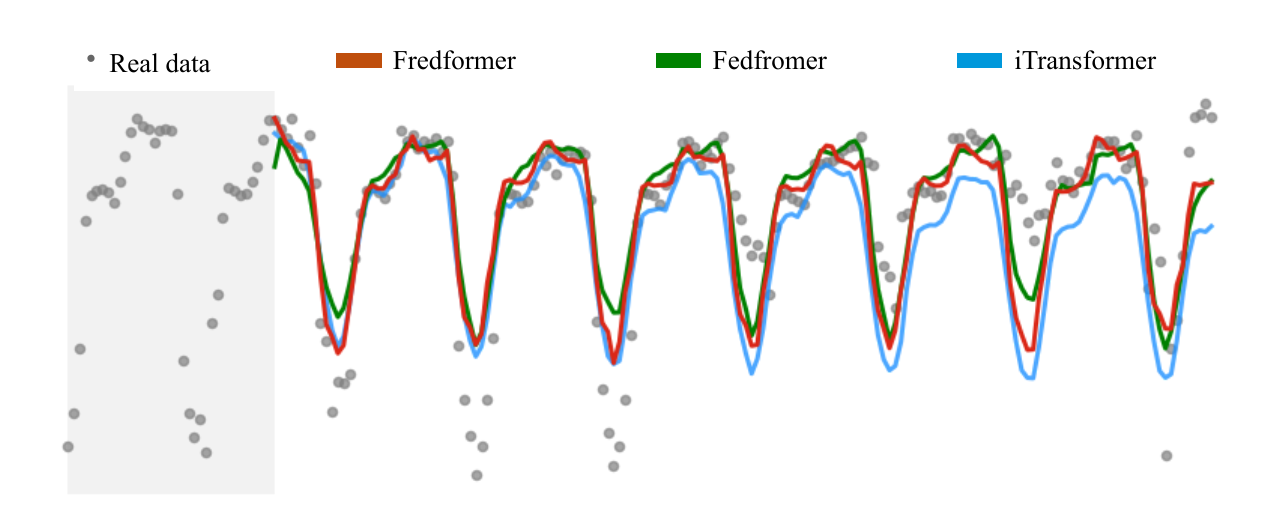}
\caption{Visualization of forecasting results ($H = 336$). Grey dots show the real data, and the grey zone represents the last part of the look-back window. Sample from channel \#2 of the ETTh1 dataset, demonstrating the distinct data features specific to this channel.}
\label{fig: app_time2}
\end{minipage}
\end{figure}

We have included additional samples from two different channels of the ETTh1 dataset. Figure \ref{fig: app_time1} presents a sample from channel \#5, and Figure \ref{fig: app_time2} showcases a sample from channel \#2. The data characteristics across different channels vary. Our model, in contrast to FEDformer, adeptly learns the similarities and differences across various channels, underscoring the significance of channel-wise attention. Compared to iTransformer, our model captures more detailed features, effectively identifying both global and local characteristics, and highlighting the importance of frequency-domain modeling.

\subsection{Frequency domain}
\label{appendix:frequency_domain}

\begin{figure}
\begin{center}
\includegraphics[width=0.98\linewidth]{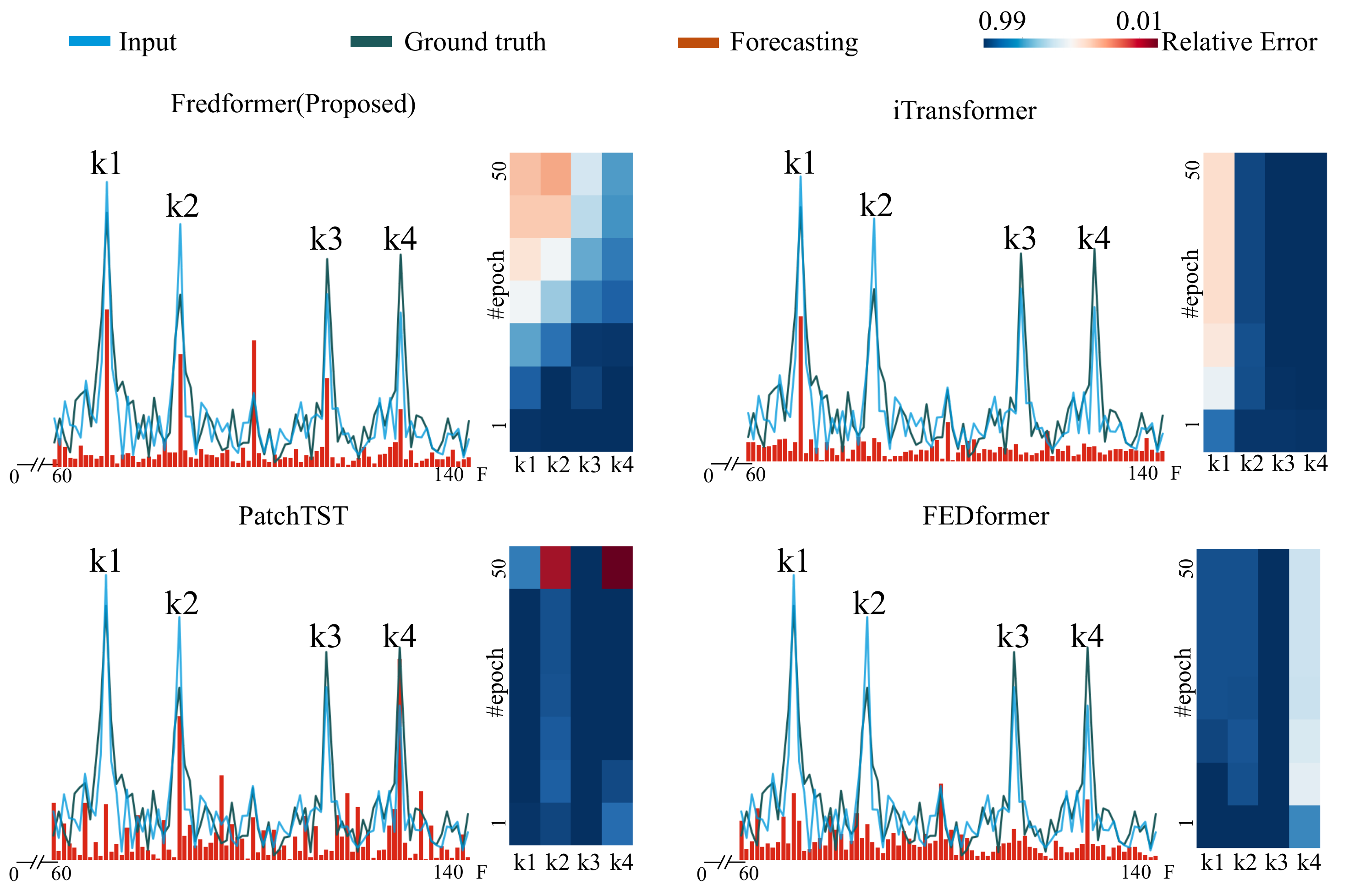}
\end{center}
\caption{
Visualizations of the learning dynamics and results for our method and baselines on the ETTh1 dataset, employing line graphs to illustrate amplitudes in the frequency domain and heatmaps to represent training epoch errors.
}
\label{fig: app_result_fre}
\end{figure}

Figure~\ref{fig: app_result_fre} visualizes another sample output of the model after 50 epochs of training in the frequency domain, with input and ground truth data from ETTh1. Similar to Section~\ref{subsec: CaseStudies}, the line graph displays the frequency amplitudes, and the heat map shows the model relative error for four components over increasing epochs.
We focus on mid-to-high frequency features here, where the amplitudes of these four key components, $k1$, $k2$, $k3$, and $k4$, are significantly lower than low-frequency components, successfully capturing these components indicates the model ability to mitigate frequency bias.
After training, our method accurately identifies $k1$, $k2$, and $k3$, with uniformly decreasing relative errors. Despite a larger learning error for $k4$, $\Delta_{k4}$ consistently diminishes. 
This performance contrasts with all baselines, which demonstrate a lack of effectiveness in capturing these frequency components, with unequal reductions in relative errors.

\begin{table*}[ht]
\centering
\caption{Comparison of GPU Memory Usage and MSE Accuracy on ETTh1 and ECl datasets, the best results are highlighted in \boldres{red}}
\label{tab:comparison_results}
\begin{tabular}{lcc|cc}
\hline
\multirow{2}{*}{Model} & \multicolumn{2}{c|}{ETTh1 Dataset} & \multicolumn{2}{c}{ECl Dataset} \\ \cline{2-5} 
                       & GPU Memory (MB) & MSE Accuracy & GPU Memory (MB) & MSE Accuracy \\ \hline
Ours                   & \boldres{408}             & \boldres{0.453}        & 4338            & 0.213        \\
Ours(Nyström)          & -               & -            & \boldres{3250}            & \boldres{0.212}        \\
iTransformer          & 642             & 0.491        & 3436            & 0.225        \\
PatchTST              & 1872            & 0.454        & 43772           & 0.256        \\
Crossformer           & 2128            & 0.666        & 40066           & 0.280        \\
FEDformer             & 1258            & 0.543        & 3824            & 0.246        \\ \hline
\end{tabular}
\end{table*}

\section{Details of the Lemma \ref{lemma:norm}}
\label{app:lemma}
Here, we give more details of Lemma.\ref{lemma:norm} in Sec.\ref{sec: Fredformer}.
First, we illustrate the full-spectrum normalization as: 

$$
    \mathbf{W}^* = \{ \sigma(\mathbf{W}_1 \dots  \mathbf{W}_N)\}
$$
This shows why normalizing the entire spectrum together does not address the disparity in amplitude across different frequency bands $\mathbf{W}_n$, failing to remove amplitude bias between key frequency components, that is,
\begin{equation}\label{eq:local_norm}
    \max(\mathbf{W}_n^*) > \max(\mathbf{W}_m^*)
\end{equation}
Then, we furthermore illustrate why time domain patching operation and normalization:
\begin{equation}\label{eq:parseval_norm}
\{ \sigma(\mathbf{Wt}_1) \dots  \sigma(\mathbf{Wt}_N)\} = \{ \sigma(\mathbf{W}_1 \dots  \mathbf{W}_N)\}
\end{equation}
where $\mathbf{Wt}$ represents a series of time-domain patches of $\mathbf{X}$. 
According to Parseval's theorem, the equivalence of the energy of the time-domain signal after local normalization and the energy of the frequency-domain signal after global normalization results in similar:
\begin{equation}\label{eq:local_norm}
\max(\mathbf{W}_n^*) > \max(\mathbf{W}_m^*)
\end{equation}
which does not solve the issue of amplitude bias among key frequency components.

Here, we furthermore discuss why this operation does not solve the issue of amplitude bias among key frequency components:
Given the frequency components $\mathbf{W}_n$ in the frequency domain and their corresponding time-domain representations $\mathbf{Wt}_n$, Parseval's theorem provides the foundation for understanding the energy equivalence between the time and frequency domains. Specifically, it states that the total energy in the time domain is equal to the total energy in the frequency domain:

$$
\sum_{n=1}^{N} \int |\sigma(\mathbf{Wt}_n)|^2 dt = \int |\sigma(\mathbf{W}_1, \dots, \mathbf{W}_N)|^2 df,
$$

where $ sigma (cdot) $ denotes the normalization operation. This theorem underscores the equivalence between applying local normalization to patches in the time domain and global normalization across the frequency spectrum.

Each point in the time domain can be expressed as the sum of energies from all frequency components at that point. Therefore, a time-domain patch, consisting of $S$ points, can be represented as the sum of all frequency components over these $S$ points:

$$
\mathbf{Patch}_{\text{time}} = \sum_{s=1}^{S} \sum_{n=1}^{N} \mathbf{W}_n(s),
$$

where $\mathbf{W}_n(s)$ represents the energy contribution of frequency component $n$ at time point $s$. Normalizing this time-domain patch equates to normalizing the weights of all frequency components across these $S$ points. Mathematically, this normalization can be represented as:

$$
\sigma(\mathbf{Patch}_{\text{time}}) = \sigma\left( \sum_{s=1}^{S} \sum_{n=1}^{N} \mathbf{W}_n(s) \right).
$$

However, normalizing within each time-domain patch does not guarantee that the maximum amplitudes across all frequency components $\mathbf{W}_n^*$ and $\mathbf{W}_m^*$ are equalized across different patches. This leads to the critical insight:

$$
\max(\mathbf{W}_n^*) > \max(\mathbf{W}_m^*), \quad \forall n, m,
$$

Indicating that local normalization in the time domain, and by extension, global normalization in the frequency domain, does not effectively address the amplitude bias problem among key frequency components. The inherent limitation is that while normalization can adjust the overall energy levels within patches or across the spectrum, it does not inherently correct for discrepancies in the amplitude distributions among different frequency components. This underscores the necessity for approaches that can specifically target and mitigate amplitude biases to ensure equitable representation and processing of all frequency components.

\section{Nyström Approximation in Transformer Self-Attention Mechanism}
\label{append:Nyström}

\textbf{Overview:} To streamline the attention computation, we select $m$ landmarks by averaging rows or columns of the attention matrix, simplifying the matrices $\mathbf{Q}_n$ and $\mathbf{K}_n$ into $\tilde{\mathbf{Q}_n}$ and $\tilde{\mathbf{K}_n}$.
The Nyström approximation for the $n$-th channel-wise attention \( \mathbf{A}_n \) is then calculated as $\mathbf{A}_n \approx \tilde{\mathbf{A}_n} = \tilde{\mathbf{F}_n} \tilde{\mathbf{A}_n} \tilde{\mathbf{B}_n}$, where
$\tilde{\mathbf{F}_n} = \text{softmax}(\mathbf{Q}_n\tilde{\mathbf{K}_n}^T)$, $\tilde{\mathbf{A}_n} = \text{softmax}(\tilde{\mathbf{Q}_n}\tilde{\mathbf{K}_n}^T)^{+}$, $\tilde{\mathbf{B}_n} = \text{softmax}(\tilde{\mathbf{Q}_n}{\mathbf{K}_n}^T)$.
Here, $\tilde{\mathbf{A}_n}^{+}$ is the Moore-Penrose inverse of $\tilde{\mathbf{A}_n}$ \citep{xiong2021nystromformer}.
This significantly reducing the computational load from $O(\frac{L}{P}C^2)$ to $O(\frac{L}{P}C)$. 
Specifically:

\textbf{Details:} We reduce the computational cost of self-attention in the Transformer encoder using the Nystr\"{o}m method. Following, we describe how to use the Nystr\"{o}m method to approximate the softmax matrix in self-attention by sampling a subset of columns and rows.

Consider the softmax matrix in self-attention, defined as:
\[ S = \text{softmax}\left( \frac{QK^T}{\sqrt{d_q}} \right) \]

This matrix can be partitioned as:
\[ 
S = 
\begin{bmatrix}
    A_S & B_S \\
    F_S & C_S
\end{bmatrix}
\]
Where \( A_S \) is derived by sampling \( m \) columns and rows from \( S \).

By employing the Nyström method, the SVD of \( A_S \) is given by:
\[ A_S = U\Lambda V^T \]

Using this, an approximation \( \hat{S} \) of \( S \) can be constructed:
\[ 
\hat{S} = 
\begin{bmatrix}
    A_S & B_S \\
    F_S & F_SA_S^{+}B_S
\end{bmatrix}
\]
Where \( A_S^{+} \) is the Moore-Penrose inverse of \( A_S \).

To further elaborate on the approximation, given a query \( q_i \) and a key \( k_j \), let:
\[ 
\Kc(q_i, K) = \text{softmax}\left( \frac{q_iK^T}{\sqrt{d_q}}\right)
\]
\[ 
\Kc(Q, k_j) = \text{softmax}\left(\frac{Qk_j^T}{\sqrt{d_q}}\right)
\]

From the above, we can derive:
\[ 
\phi(q_i, K) = \Lambda^{-\frac{1}{2}} V^T \Kc(q_i, K)_{m\times 1}
\]
\[ 
\phi(Q, k_j) = \Lambda ^{-\frac{1}{2}}U^T \Kc(Q, k_j)_{m\times 1}
\]

Thus, the Nyström approximation for a particular entry in \( \hat{S} \) is:
\[ 
\hat{S}_{ij} = \phi(q_i, K)^T \phi(Q, k_j)
\]

In matrix form, \( \hat{S} \) can be represented as:
\[ 
\hat{S} = \text{softmax}\left(\frac{QK^T}{\sqrt{d_q}}\right)_{n\times m} A_S^{+} \text{softmax}\left(\frac{QK^T}{\sqrt{d_q}}\right)_{m\times n}
\]

This method allows for the approximation of the softmax matrix in self-attention, potentially offering computational benefits.


\section{Detailed results of all datasets}
\label{appendix:fullresult}
Here, we show the detailed forecasting results of full datasets in the Table. \ref{appendix:fullresult}.
The \boldres{best} and \secondres{second best} results are highlighted.
With a default look-back window of $L = 96$, our proposal shows leading performance on most datasets and different prediction length settings, with 60 top-1 (29 + 31) cases out of 80 in total.
\begin{table*}[htbp]
  \caption{Full results of the long-term forecasting task}. We compare extensive competitive models under different prediction lengths following the setting of iTransformer \citep{liu2023itransformer}. The input sequence length is set to 96 for all baselines. Avg means the average results from all four prediction lengths.\label{tab:full_baseline_results}
  \vskip -0.0in
  \vspace{3pt}
  \renewcommand{\arraystretch}{0.85} 
  \centering
  \renewcommand{\multirowsetup}{\centering}
  \setlength{\tabcolsep}{1pt}
  \begin{tabular}{c|c|cc|cc|cc|cc|cc|cc|cc|cc|cc|cc|cc|cc}
    \toprule
    \multicolumn{2}{c}{\multirow{2}{*}{Models}} & 
    \multicolumn{2}{c}{\rotatebox{0}{\scalebox{0.8}{\textbf{\method}}}} &
    \multicolumn{2}{c}{\rotatebox{0}{\scalebox{0.8}{iTransformer}}} &
    \multicolumn{2}{c}{\rotatebox{0}{\scalebox{0.8}{RLinear}}} &
    \multicolumn{2}{c}{\rotatebox{0}{\scalebox{0.8}{PatchTST}}} &
    \multicolumn{2}{c}{\rotatebox{0}{\scalebox{0.8}{Crossformer}}}  &
    \multicolumn{2}{c}{\rotatebox{0}{\scalebox{0.8}{TiDE}}} &
    \multicolumn{2}{c}{\rotatebox{0}{\scalebox{0.8}{{TimesNet}}}} &
    \multicolumn{2}{c}{\rotatebox{0}{\scalebox{0.8}{DLinear}}}&
    \multicolumn{2}{c}{\rotatebox{0}{\scalebox{0.8}{SCINet}}} &
    \multicolumn{2}{c}{\rotatebox{0}{\scalebox{0.8}{FEDformer}}} &
    \multicolumn{2}{c}{\rotatebox{0}{\scalebox{0.8}{Stationary}}} &
    \multicolumn{2}{c}{\rotatebox{0}{\scalebox{0.8}{Autoformer}}} \\
    \multicolumn{2}{c}{} &
    \multicolumn{2}{c}{\scalebox{0.8}{\textbf{(Ours)}}} & 
    \multicolumn{2}{c}{\scalebox{0.8}{\citeyearpar{liu2023itransformer}}} & 
    \multicolumn{2}{c}{\scalebox{0.8}{\citeyearpar{li2023RLinear}}} & 
    \multicolumn{2}{c}{\scalebox{0.8}{\citeyearpar{PatchTST}}} & 
    \multicolumn{2}{c}{\scalebox{0.8}{\citeyearpar{Crossformer}}}  & 
    \multicolumn{2}{c}{\scalebox{0.8}{\citeyearpar{das2023TiDE}}} & 
    \multicolumn{2}{c}{\scalebox{0.8}{\citeyearpar{Timesnet}}} & 
    \multicolumn{2}{c}{\scalebox{0.8}{\citeyearpar{DLinear}}}& 
    \multicolumn{2}{c}{\scalebox{0.8}{\citeyearpar{liu2022scinet}}} &
    \multicolumn{2}{c}{\scalebox{0.8}{\citeyearpar{ICMLFedformer}}} &
    \multicolumn{2}{c}{\scalebox{0.8}{\citeyearpar{Non-stationary}}} &
    \multicolumn{2}{c}{\scalebox{0.8}{\citeyearpar{Autoformer}}} \\
    
    \cmidrule(lr){3-4} \cmidrule(lr){5-6}\cmidrule(lr){7-8} \cmidrule(lr){9-10}\cmidrule(lr){11-12}\cmidrule(lr){13-14} \cmidrule(lr){15-16} \cmidrule(lr){17-18} \cmidrule(lr){19-20} \cmidrule(lr){21-22} \cmidrule(lr){23-24}\cmidrule(lr){25-26}
    
    \multicolumn{2}{c}{Metric}  & \scalebox{0.78}{MSE} & \scalebox{0.78}{MAE} & \scalebox{0.78}{MSE} & \scalebox{0.78}{MAE}  & \scalebox{0.78}{MSE} & \scalebox{0.78}{MAE}  & \scalebox{0.78}{MSE} & \scalebox{0.78}{MAE} & \scalebox{0.78}{MSE} & \scalebox{0.78}{MAE}  & \scalebox{0.78}{MSE} & \scalebox{0.78}{MAE}  & \scalebox{0.78}{MSE} & \scalebox{0.78}{MAE} & \scalebox{0.78}{MSE} & \scalebox{0.78}{MAE} & \scalebox{0.78}{MSE} & \scalebox{0.78}{MAE} & \scalebox{0.78}{MSE} & \scalebox{0.78}{MAE} & \scalebox{0.78}{MSE} & \scalebox{0.78}{MAE} & \scalebox{0.78}{MSE} & \scalebox{0.78}{MAE} \\
    
    \toprule
    
    \multirow{5}{*}{\rotatebox{90}{\scalebox{0.95}{ETTm1}}}
    &  \scalebox{0.78}{96} & \boldres{\scalebox{0.78}{0.326}} & \boldres{\scalebox{0.78}{0.361}} & \scalebox{0.78}{0.334} & \scalebox{0.78}{0.368} & \scalebox{0.78}{0.355} & \scalebox{0.78}{0.376} & \secondres{\scalebox{0.78}{0.329}} & \secondres{\scalebox{0.78}{0.367}} & \scalebox{0.78}{0.404} & \scalebox{0.78}{0.426} & \scalebox{0.78}{0.364} & \scalebox{0.78}{0.387} &{\scalebox{0.78}{0.338}} &{\scalebox{0.78}{0.375}} &{\scalebox{0.78}{0.345}} &{\scalebox{0.78}{0.372}} & \scalebox{0.78}{0.418} & \scalebox{0.78}{0.438} &\scalebox{0.78}{0.379} &\scalebox{0.78}{0.419} &\scalebox{0.78}{0.386} &\scalebox{0.78}{0.398} &\scalebox{0.78}{0.505} &\scalebox{0.78}{0.475} \\ 
    
    & \scalebox{0.78}{192} & \boldres{\scalebox{0.78}{0.363}} & \boldres{\scalebox{0.78}{0.380}} & \scalebox{0.78}{0.377} & \scalebox{0.78}{0.391} & \scalebox{0.78}{0.391} & \scalebox{0.78}{0.392} & \secondres{\scalebox{0.78}{0.367}} & \secondres{\scalebox{0.78}{0.385}} & \scalebox{0.78}{0.450} & \scalebox{0.78}{0.451} &\scalebox{0.78}{0.398} & \scalebox{0.78}{0.404} &\scalebox{0.78}{0.374} & \scalebox{0.78}{0.387} &{\scalebox{0.78}{0.380}} &{\scalebox{0.78}{0.389}} & \scalebox{0.78}{0.439} & \scalebox{0.78}{0.450}  &\scalebox{0.78}{0.426} &\scalebox{0.78}{0.441} &\scalebox{0.78}{0.459} &\scalebox{0.78}{0.444} &\scalebox{0.78}{0.553} &\scalebox{0.78}{0.496} \\ 
    
    & \scalebox{0.78}{336} &\boldres{ \scalebox{0.78}{0.395}} & \boldres{\scalebox{0.78}{0.403}} & \scalebox{0.78}{0.426} & \scalebox{0.78}{0.420} & \scalebox{0.78}{0.424} & \scalebox{0.78}{0.415} & \secondres{\scalebox{0.78}{0.399}} & \secondres{\scalebox{0.78}{0.410}} & \scalebox{0.78}{0.532}  &\scalebox{0.78}{0.515} & \scalebox{0.78}{0.428} & \scalebox{0.78}{0.425} & \scalebox{0.78}{0.410} & \scalebox{0.78}{0.411}  &{\scalebox{0.78}{0.413}} &{\scalebox{0.78}{0.413}} & \scalebox{0.78}{0.490} & \scalebox{0.78}{0.485}  &\scalebox{0.78}{0.445} &\scalebox{0.78}{0.459} &\scalebox{0.78}{0.495} &\scalebox{0.78}{0.464} &\scalebox{0.78}{0.621} &\scalebox{0.78}{0.537} \\ 
    
    & \scalebox{0.78}{720} & \boldres{\scalebox{0.78}{0.453}} & \boldres{\scalebox{0.78}{0.438}} & \scalebox{0.78}{0.491} & \scalebox{0.78}{0.459} & \scalebox{0.78}{0.487} & \scalebox{0.78}{0.450} & \secondres{\scalebox{0.78}{0.454}} & \secondres{\scalebox{0.78}{0.439}} & \scalebox{0.78}{0.666} & \scalebox{0.78}{0.589} & \scalebox{0.78}{0.487} & \scalebox{0.78}{0.461} &{\scalebox{0.78}{0.478}} & \scalebox{0.78}{0.450} & \scalebox{0.78}{0.474} &{\scalebox{0.78}{0.453}} & \scalebox{0.78}{0.595} & \scalebox{0.78}{0.550}  &\scalebox{0.78}{0.543} &\scalebox{0.78}{0.490} &\scalebox{0.78}{0.585} &\scalebox{0.78}{0.516} &\scalebox{0.78}{0.671} &\scalebox{0.78}{0.561} \\ 
    
    \cmidrule(lr){2-26}
    
    & \scalebox{0.78}{Avg} & \boldres{\scalebox{0.78}{0.384}} & \boldres{\scalebox{0.78}{0.395}} & \scalebox{0.78}{0.407} & \scalebox{0.78}{0.410} & \scalebox{0.78}{0.414} & \scalebox{0.78}{0.407} & \secondres{\scalebox{0.78}{0.387}} & \secondres{\scalebox{0.78}{0.400}} & \scalebox{0.78}{0.513} & \scalebox{0.78}{0.496} & \scalebox{0.78}{0.419} & \scalebox{0.78}{0.419} & \scalebox{0.78}{0.400} & \scalebox{0.78}{0.406}  &{\scalebox{0.78}{0.403}} &{\scalebox{0.78}{0.407}} & \scalebox{0.78}{0.485} & \scalebox{0.78}{0.481}  &\scalebox{0.78}{0.448} &\scalebox{0.78}{0.452} &\scalebox{0.78}{0.481} &\scalebox{0.78}{0.456} &\scalebox{0.78}{0.588} &\scalebox{0.78}{0.517} \\ 
    
    \midrule
    
    \multirow{5}{*}{\rotatebox{90}{\scalebox{0.95}{ETTm2}}}
    &  \scalebox{0.78}{96} & \secondres{\scalebox{0.78}{0.177}} & \boldres{\scalebox{0.78}{0.259}} & \scalebox{0.78}{0.180} & \scalebox{0.78}{0.264} & \scalebox{0.78}{0.182} & \scalebox{0.78}{0.265} & \boldres{\scalebox{0.78}{0.175}} & \boldres{\scalebox{0.78}{0.259}} & \scalebox{0.78}{0.287} & \scalebox{0.78}{0.366} & \scalebox{0.78}{0.207} & \scalebox{0.78}{0.305} &{\scalebox{0.78}{0.187}} &\scalebox{0.78}{0.267} &\scalebox{0.78}{0.193} &\scalebox{0.78}{0.292} & \scalebox{0.78}{0.286} & \scalebox{0.78}{0.377} &\scalebox{0.78}{0.203} &\scalebox{0.78}{0.287} &{\scalebox{0.78}{0.192}} &\scalebox{0.78}{0.274} &\scalebox{0.78}{0.255} &\scalebox{0.78}{0.339} \\ 
    
    & \scalebox{0.78}{192} & \secondres{\scalebox{0.78}{0.243}} & \boldres{\scalebox{0.78}{0.301}} & \scalebox{0.78}{0.250} & \scalebox{0.78}{0.309} & \scalebox{0.78}{0.246} & \scalebox{0.78}{0.304} & \boldres{\scalebox{0.78}{0.241}} & \secondres{\scalebox{0.78}{0.302}} & \scalebox{0.78}{0.414} & \scalebox{0.78}{0.492} & \scalebox{0.78}{0.290} & \scalebox{0.78}{0.364} &{\scalebox{0.78}{0.249}} &{\scalebox{0.78}{0.309}} &\scalebox{0.78}{0.284} &\scalebox{0.78}{0.362} & \scalebox{0.78}{0.399} & \scalebox{0.78}{0.445} &\scalebox{0.78}{0.269} &\scalebox{0.78}{0.328} &\scalebox{0.78}{0.280} &\scalebox{0.78}{0.339} &\scalebox{0.78}{0.281} &\scalebox{0.78}{0.340} \\ 
    
    & \scalebox{0.78}{336} & \boldres{\scalebox{0.78}{0.302}} & \boldres{\scalebox{0.78}{0.340}} & {\scalebox{0.78}{0.311}} & {\scalebox{0.78}{0.348}} & \scalebox{0.78}{0.307} & \secondres{\scalebox{0.78}{0.342}} & \secondres{\scalebox{0.78}{0.305}} & \scalebox{0.78}{0.343}  & \scalebox{0.78}{0.597} & \scalebox{0.78}{0.542}  & \scalebox{0.78}{0.377} & \scalebox{0.78}{0.422} &{\scalebox{0.78}{0.321}} &{\scalebox{0.78}{0.351}} &\scalebox{0.78}{0.369} &\scalebox{0.78}{0.427} & \scalebox{0.78}{0.637} & \scalebox{0.78}{0.591} &\scalebox{0.78}{0.325} &\scalebox{0.78}{0.366} &\scalebox{0.78}{0.334} &\scalebox{0.78}{0.361} &\scalebox{0.78}{0.339} &\scalebox{0.78}{0.372} \\ 
    
    & \scalebox{0.78}{720} & \boldres{\scalebox{0.78}{0.397}} & \boldres{\scalebox{0.78}{0.396}} & \scalebox{0.78}{0.412} & \scalebox{0.78}{0.407} & \scalebox{0.78}{0.407} & \secondres{\scalebox{0.78}{0.398}} & \secondres{\scalebox{0.78}{0.402}} & \scalebox{0.78}{0.400} & \scalebox{0.78}{1.730} & \scalebox{0.78}{1.042} & \scalebox{0.78}{0.558} & \scalebox{0.78}{0.524} &{\scalebox{0.78}{0.408}} &{\scalebox{0.78}{0.403}} &\scalebox{0.78}{0.554} &\scalebox{0.78}{0.522} & \scalebox{0.78}{0.960} & \scalebox{0.78}{0.735} &\scalebox{0.78}{0.421} &\scalebox{0.78}{0.415} &\scalebox{0.78}{0.417} &\scalebox{0.78}{0.413} &\scalebox{0.78}{0.433} &\scalebox{0.78}{0.432} \\ 
    
    \cmidrule(lr){2-26}
    
    & \scalebox{0.78}{Avg} & \boldres{\scalebox{0.78}{0.279}} & \boldres{\scalebox{0.78}{0.324}} & {\scalebox{0.78}{0.288}} & {\scalebox{0.78}{0.332}} & \scalebox{0.78}{0.286} & \scalebox{0.78}{0.327} & \secondres{\scalebox{0.78}{0.281}} & \secondres{\scalebox{0.78}{0.326}} & \scalebox{0.78}{0.757} & \scalebox{0.78}{0.610} & \scalebox{0.78}{0.358} & \scalebox{0.78}{0.404} &{\scalebox{0.78}{0.291}} &{\scalebox{0.78}{0.333}} &\scalebox{0.78}{0.350} &\scalebox{0.78}{0.401} & \scalebox{0.78}{0.571} & \scalebox{0.78}{0.537} &\scalebox{0.78}{0.305} &\scalebox{0.78}{0.349} &\scalebox{0.78}{0.306} &\scalebox{0.78}{0.347} &\scalebox{0.78}{0.327} &\scalebox{0.78}{0.371} \\ 
    
    \midrule
    
    \multirow{5}{*}{\rotatebox{90}{
    \scalebox{0.95}{ETTh1}}}
    &  \scalebox{0.78}{96} & \boldres{\scalebox{0.78}{0.373}} & \boldres{\scalebox{0.78}{0.392}} & {\scalebox{0.78}{0.386}} & {\scalebox{0.78}{0.405}} & \scalebox{0.78}{0.386} & \secondres{\scalebox{0.78}{0.395}} & \scalebox{0.78}{0.414} & \scalebox{0.78}{0.419} & \scalebox{0.78}{0.423} & \scalebox{0.78}{0.448} & \scalebox{0.78}{0.479}& \scalebox{0.78}{0.464}  & \scalebox{0.78}{0.384} &{\scalebox{0.78}{0.402}} & \scalebox{0.78}{0.386} & \scalebox{0.78}{0.400} & \scalebox{0.78}{0.654} & \scalebox{0.78}{0.599} & \secondres{\scalebox{0.78}{0.376}} &\scalebox{0.78}{0.419} &\scalebox{0.78}{0.513} &\scalebox{0.78}{0.491} &\scalebox{0.78}{0.449} &\scalebox{0.78}{0.459}  \\ 
    
    & \scalebox{0.78}{192} & \secondres{\scalebox{0.78}{0.433}} & \boldres{\scalebox{0.78}{0.420}} & \scalebox{0.78}{0.441} & \scalebox{0.78}{0.436} & {\scalebox{0.78}{0.437}} & \secondres{\scalebox{0.78}{0.424}} & \scalebox{0.78}{0.460} & \scalebox{0.78}{0.445} & \scalebox{0.78}{0.471} & \scalebox{0.78}{0.474}  & \scalebox{0.78}{0.525} & \scalebox{0.78}{0.492} &\scalebox{0.78}{0.436} & \scalebox{0.78}{0.429}  &{\scalebox{0.78}{0.437}} &{\scalebox{0.78}{0.432}} & \scalebox{0.78}{0.719} & \scalebox{0.78}{0.631} &\boldres{\scalebox{0.78}{0.420}} &\scalebox{0.78}{0.448} &\scalebox{0.78}{0.534} &\scalebox{0.78}{0.504} &\scalebox{0.78}{0.500} &\scalebox{0.78}{0.482} \\ 
    
    & \scalebox{0.78}{336} & \secondres{\scalebox{0.78}{0.470}} & \boldres{\scalebox{0.78}{0.437}} & {\scalebox{0.78}{0.487}} & \scalebox{0.78}{0.458} & \scalebox{0.78}{0.479} & \secondres{\scalebox{0.78}{0.446}} & \scalebox{0.78}{0.501} & \scalebox{0.78}{0.466} & \scalebox{0.78}{0.570} & \scalebox{0.78}{0.546} & \scalebox{0.78}{0.565} & \scalebox{0.78}{0.515} &\scalebox{0.78}{0.491} &\scalebox{0.78}{0.469} &{\scalebox{0.78}{0.481}} & {\scalebox{0.78}{0.459}} & \scalebox{0.78}{0.778} & \scalebox{0.78}{0.659} &\boldres{\scalebox{0.78}{0.459}} &{\scalebox{0.78}{0.465}} &\scalebox{0.78}{0.588} &\scalebox{0.78}{0.535} &\scalebox{0.78}{0.521} &\scalebox{0.78}{0.496} \\ 
    
    & \scalebox{0.78}{720} & \boldres{\scalebox{0.78}{0.467}} & \boldres{\scalebox{0.78}{0.456}} & {\scalebox{0.78}{0.503}} & {\scalebox{0.78}{0.491}} & \secondres{\scalebox{0.78}{0.481}} & 
    \secondres{\scalebox{0.78}{0.470}} & \scalebox{0.78}{0.500} & \scalebox{0.78}{0.488} & \scalebox{0.78}{0.653} & \scalebox{0.78}{0.621} & \scalebox{0.78}{0.594} & \scalebox{0.78}{0.558} &\scalebox{0.78}{0.521} &{\scalebox{0.78}{0.500}} &\scalebox{0.78}{0.519} &\scalebox{0.78}{0.516} & \scalebox{0.78}{0.836} & \scalebox{0.78}{0.699} &{\scalebox{0.78}{0.506}} &{\scalebox{0.78}{0.507}} &\scalebox{0.78}{0.643} &\scalebox{0.78}{0.616} &{\scalebox{0.78}{0.514}} &\scalebox{0.78}{0.512}  \\ 
    
    \cmidrule(lr){2-26}
    
    & \scalebox{0.78}{Avg} & \boldres{\scalebox{0.78}{0.435}} & \boldres{\scalebox{0.78}{0.426}} & {\scalebox{0.78}{0.454}} & \scalebox{0.78}{0.447} & \scalebox{0.78}{0.446} & \secondres{\scalebox{0.78}{0.434}} & \scalebox{0.78}{0.469} & \scalebox{0.78}{0.454} & \scalebox{0.78}{0.529} & \scalebox{0.78}{0.522} & \scalebox{0.78}{0.541} & \scalebox{0.78}{0.507} &\scalebox{0.78}{0.458} &{\scalebox{0.78}{0.450}} &{\scalebox{0.78}{0.456}} &{\scalebox{0.78}{0.452}} & \scalebox{0.78}{0.747} & \scalebox{0.78}{0.647} &\secondres{\scalebox{0.78}{0.440}} &\scalebox{0.78}{0.460} &\scalebox{0.78}{0.570} &\scalebox{0.78}{0.537} &\scalebox{0.78}{0.496} &\scalebox{0.78}{0.487}  \\ 
    
    \midrule

    \multirow{5}{*}{\rotatebox{90}{\scalebox{0.95}{ETTh2}}}
    &  \scalebox{0.78}{96} & \secondres{\scalebox{0.78}{0.293}} & \secondres{\scalebox{0.78}{0.342}} & \scalebox{0.78}{0.297} & {\scalebox{0.78}{0.349}} & \boldres{\scalebox{0.78}{0.288}} & \boldres{\scalebox{0.78}{0.338}} & {\scalebox{0.78}{0.302}} & \scalebox{0.78}{0.348} & \scalebox{0.78}{0.745} & \scalebox{0.78}{0.584} &\scalebox{0.78}{0.400} & \scalebox{0.78}{0.440}  & {\scalebox{0.78}{0.340}} & {\scalebox{0.78}{0.374}} &{\scalebox{0.78}{0.333}} &{\scalebox{0.78}{0.387}} & \scalebox{0.78}{0.707} & \scalebox{0.78}{0.621}  &\scalebox{0.78}{0.358} &\scalebox{0.78}{0.397} &\scalebox{0.78}{0.476} &\scalebox{0.78}{0.458} &\scalebox{0.78}{0.346} &\scalebox{0.78}{0.388} \\ 
    
    & \scalebox{0.78}{192} & \boldres{\scalebox{0.78}{0.371}} & \boldres{\scalebox{0.78}{0.389}} & \scalebox{0.78}{0.380} & \scalebox{0.78}{0.400}& \secondres{\scalebox{0.78}{0.374}} & \secondres{\scalebox{0.78}{0.390}} &{\scalebox{0.78}{0.388}} & {\scalebox{0.78}{0.400}} & \scalebox{0.78}{0.877} & \scalebox{0.78}{0.656} & \scalebox{0.78}{0.528} & \scalebox{0.78}{0.509} & {\scalebox{0.78}{0.402}} & {\scalebox{0.78}{0.414}} &\scalebox{0.78}{0.477} &\scalebox{0.78}{0.476} & \scalebox{0.78}{0.860} & \scalebox{0.78}{0.689} &{\scalebox{0.78}{0.429}} &{\scalebox{0.78}{0.439}} &\scalebox{0.78}{0.512} &\scalebox{0.78}{0.493} &\scalebox{0.78}{0.456} &\scalebox{0.78}{0.452} \\ 
    
    & \scalebox{0.78}{336} & \boldres{\scalebox{0.78}{0.382}} & \boldres{\scalebox{0.78}{0.409}} & {\scalebox{0.78}{0.428}} & \scalebox{0.78}{0.432} & \secondres{\scalebox{0.78}{0.415}} & \secondres{\scalebox{0.78}{0.426}} & \scalebox{0.78}{0.426} & {\scalebox{0.78}{0.433}}& \scalebox{0.78}{1.043} & \scalebox{0.78}{0.731} & \scalebox{0.78}{0.643} & \scalebox{0.78}{0.571}  & {\scalebox{0.78}{0.452}} & {\scalebox{0.78}{0.452}} &\scalebox{0.78}{0.594} &\scalebox{0.78}{0.541} & \scalebox{0.78}{1.000} &\scalebox{0.78}{0.744} &\scalebox{0.78}{0.496} &\scalebox{0.78}{0.487} &\scalebox{0.78}{0.552} &\scalebox{0.78}{0.551} &{\scalebox{0.78}{0.482}} &\scalebox{0.78}{0.486}\\ 
    
    & \scalebox{0.78}{720} & \boldres{\scalebox{0.78}{0.415}} & \boldres{\scalebox{0.78}{0.434}} & \scalebox{0.78}{0.427} & \scalebox{0.78}{0.445} & \secondres{\scalebox{0.78}{0.420}} & \secondres{\scalebox{0.78}{0.440}} & {\scalebox{0.78}{0.431}} & {\scalebox{0.78}{0.446}} & \scalebox{0.78}{1.104} & \scalebox{0.78}{0.763} & \scalebox{0.78}{0.874} & \scalebox{0.78}{0.679} & {\scalebox{0.78}{0.462}} & {\scalebox{0.78}{0.468}} &\scalebox{0.78}{0.831} &\scalebox{0.78}{0.657} & \scalebox{0.78}{1.249} & \scalebox{0.78}{0.838} &{\scalebox{0.78}{0.463}} &{\scalebox{0.78}{0.474}} &\scalebox{0.78}{0.562} &\scalebox{0.78}{0.560} &\scalebox{0.78}{0.515} &\scalebox{0.78}{0.511} \\ 
    
    \cmidrule(lr){2-26}
    
    & \scalebox{0.78}{Avg} & \boldres{\scalebox{0.78}{0.365}} & \boldres{\scalebox{0.78}{0.393}} & \scalebox{0.78}{0.383} & \scalebox{0.78}{0.407} & \secondres{\scalebox{0.78}{0.374}} & \secondres{\scalebox{0.78}{0.398}} & {\scalebox{0.78}{0.387}} & {\scalebox{0.78}{0.407}} & \scalebox{0.78}{0.942} & \scalebox{0.78}{0.684} & \scalebox{0.78}{0.611} & \scalebox{0.78}{0.550}  &{\scalebox{0.78}{0.414}} &{\scalebox{0.78}{0.427}} &\scalebox{0.78}{0.559} &\scalebox{0.78}{0.515} & \scalebox{0.78}{0.954} & \scalebox{0.78}{0.723} &\scalebox{0.78}{{0.437}} &\scalebox{0.78}{{0.449}} &\scalebox{0.78}{0.526} &\scalebox{0.78}{0.516} &\scalebox{0.78}{0.450} &\scalebox{0.78}{0.459} \\ 
    
    \midrule
    
    \multirow{5}{*}{\rotatebox{90}{\scalebox{0.95}{ECL}}} 
    &  \scalebox{0.78}{96} & \boldres{\scalebox{0.78}{0.147}} & \secondres{\scalebox{0.78}{0.241}} & \secondres{\scalebox{0.78}{0.148}} & \boldres{\scalebox{0.78}{0.240}} & \scalebox{0.78}{0.201} & \scalebox{0.78}{0.281} & \scalebox{0.78}{0.195} & \scalebox{0.78}{0.285} & \scalebox{0.78}{0.219} & \scalebox{0.78}{0.314} & \scalebox{0.78}{0.237} & \scalebox{0.78}{0.329} &\scalebox{0.78}{0.168} &\scalebox{0.78}{0.272} &\scalebox{0.78}{0.197} &\scalebox{0.78}{0.282} & \scalebox{0.78}{0.247} & \scalebox{0.78}{0.345} &\scalebox{0.78}{0.193} &\scalebox{0.78}{0.308} &{\scalebox{0.78}{0.169}} &{\scalebox{0.78}{0.273}} &\scalebox{0.78}{0.201} &\scalebox{0.78}{0.317}  \\ 
    
    & \scalebox{0.78}{192} & \secondres{\scalebox{0.78}{0.165}} & \secondres{\scalebox{0.78}{0.258}} & \boldres{\scalebox{0.78}{0.162}} & \boldres{\scalebox{0.78}{0.253}} & \scalebox{0.78}{0.201} & \scalebox{0.78}{0.283} & \scalebox{0.78}{0.199} & \scalebox{0.78}{0.289} & \scalebox{0.78}{0.231} & \scalebox{0.78}{0.322} & \scalebox{0.78}{0.236} & \scalebox{0.78}{0.330} &{\scalebox{0.78}{0.184}} &\scalebox{0.78}{0.289} &\scalebox{0.78}{0.196} &{\scalebox{0.78}{0.285}} & \scalebox{0.78}{0.257} & \scalebox{0.78}{0.355} &\scalebox{0.78}{0.201} &\scalebox{0.78}{0.315} &\scalebox{0.78}{0.182} &\scalebox{0.78}{0.286} &\scalebox{0.78}{0.222} &\scalebox{0.78}{0.334} \\ 
    
    & \scalebox{0.78}{336} & \boldres{\scalebox{0.78}{0.177}} & \secondres{\scalebox{0.78}{0.273}} & \secondres{\scalebox{0.78}{0.178}} & \boldres{\scalebox{0.78}{0.269}} & \scalebox{0.78}{0.215} & \scalebox{0.78}{0.298} & \scalebox{0.78}{0.215} & \scalebox{0.78}{0.305} & \scalebox{0.78}{0.246} & \scalebox{0.78}{0.337} & \scalebox{0.78}{0.249} & \scalebox{0.78}{0.344} &\scalebox{0.78}{0.198} &{\scalebox{0.78}{0.300}} &\scalebox{0.78}{0.209} &{\scalebox{0.78}{0.301}} & \scalebox{0.78}{0.269} & \scalebox{0.78}{0.369} &\scalebox{0.78}{0.214} &\scalebox{0.78}{0.329} &{\scalebox{0.78}{0.200}} &\scalebox{0.78}{0.304} &\scalebox{0.78}{0.231} &\scalebox{0.78}{0.338}  \\ 
    
    & \scalebox{0.78}{720} & \boldres{\scalebox{0.78}{0.213}} & \boldres{\scalebox{0.78}{0.304}}  & \scalebox{0.78}{0.225} & \secondres{\scalebox{0.78}{0.317}} & \scalebox{0.78}{0.257} & \scalebox{0.78}{0.331} & \scalebox{0.78}{0.256} & \scalebox{0.78}{0.337} & \scalebox{0.78}{0.280} & \scalebox{0.78}{0.363} & \scalebox{0.78}{0.284} & \scalebox{0.78}{0.373} &\secondres{\scalebox{0.78}{0.220}} &\scalebox{0.78}{0.320}&\scalebox{0.78}{0.245} &\scalebox{0.78}{0.333} & \scalebox{0.78}{0.299} & \scalebox{0.78}{0.390} &\scalebox{0.78}{0.246} &\scalebox{0.78}{0.355} &{\scalebox{0.78}{0.222}} &{\scalebox{0.78}{0.321}} &\scalebox{0.78}{0.254} &\scalebox{0.78}{0.361} \\ 
    
    \cmidrule(lr){2-26}
    
    & \scalebox{0.78}{Avg} & \boldres{\scalebox{0.78}{0.175}} & \boldres{\scalebox{0.78}{0.269}} & \secondres{\scalebox{0.78}{0.178}} & \secondres{\scalebox{0.78}{0.270}} & \scalebox{0.78}{0.219} & \scalebox{0.78}{0.298} & \scalebox{0.78}{0.216} & \scalebox{0.78}{0.304} & \scalebox{0.78}{0.244} & \scalebox{0.78}{0.334} & \scalebox{0.78}{0.251} & \scalebox{0.78}{0.344} &\scalebox{0.78}{0.192} &\scalebox{0.78}{0.295} &\scalebox{0.78}{0.212} &\scalebox{0.78}{0.300} & \scalebox{0.78}{0.268} & \scalebox{0.78}{0.365} &\scalebox{0.78}{0.214} &\scalebox{0.78}{0.327} &{\scalebox{0.78}{0.193}} &{\scalebox{0.78}{0.296}} &\scalebox{0.78}{0.227} &\scalebox{0.78}{0.338} \\ 
    
    \midrule
    
    \multirow{5}{*}{\rotatebox{90}{\scalebox{0.95}{Traffic}}} 
    & \scalebox{0.78}{96} & \secondres{\scalebox{0.78}{0.406}} & \secondres{\scalebox{0.78}{0.277}} & \boldres{\scalebox{0.78}{0.395}} & \boldres{\scalebox{0.78}{0.268}} & \scalebox{0.78}{0.649} & \scalebox{0.78}{0.389} & \scalebox{0.78}{0.544} & \scalebox{0.78}{0.359} & \scalebox{0.78}{0.522} & \scalebox{0.78}{0.290} & \scalebox{0.78}{0.805} & \scalebox{0.78}{0.493} &{\scalebox{0.78}{0.593}} &{\scalebox{0.78}{0.321}} &\scalebox{0.78}{0.650} &\scalebox{0.78}{0.396} & \scalebox{0.78}{0.788} & \scalebox{0.78}{0.499} &{\scalebox{0.78}{0.587}} &\scalebox{0.78}{0.366} &\scalebox{0.78}{0.612} &{\scalebox{0.78}{0.338}} &\scalebox{0.78}{0.613} &\scalebox{0.78}{0.388} \\ 
    
    & \scalebox{0.78}{192} & \secondres{\scalebox{0.78}{0.426}} & \secondres{\scalebox{0.78}{0.290}} & \boldres{\scalebox{0.78}{0.417}} & \boldres{\scalebox{0.78}{0.276}} & \scalebox{0.78}{0.601} & \scalebox{0.78}{0.366} & \scalebox{0.78}{0.540} & \scalebox{0.78}{0.354} & \scalebox{0.78}{0.530} & \scalebox{0.78}{0.293} & \scalebox{0.78}{0.756} & \scalebox{0.78}{0.474} &\scalebox{0.78}{0.617} &{\scalebox{0.78}{0.336}} &{\scalebox{0.78}{0.598}} &\scalebox{0.78}{0.370} & \scalebox{0.78}{0.789} & \scalebox{0.78}{0.505} &\scalebox{0.78}{0.604} &\scalebox{0.78}{0.373} &\scalebox{0.78}{0.613} &{\scalebox{0.78}{0.340}} &\scalebox{0.78}{0.616} &\scalebox{0.78}{0.382}  \\ 
    
    & \scalebox{0.78}{336} & \boldres{\scalebox{0.78}{0.432}} & \boldres{\scalebox{0.78}{0.281}} & \secondres{\scalebox{0.78}{0.433}} & \secondres{\scalebox{0.78}{0.283}} & \scalebox{0.78}{0.609} & \scalebox{0.78}{0.369} & \scalebox{0.78}{0.551} & \scalebox{0.78}{0.358} & \scalebox{0.78}{0.558} & \scalebox{0.78}{0.305}  & \scalebox{0.78}{0.762} & \scalebox{0.78}{0.477} &\scalebox{0.78}{0.629} &{\scalebox{0.78}{0.336}}  &{\scalebox{0.78}{0.605}} &\scalebox{0.78}{0.373} & \scalebox{0.78}{0.797} & \scalebox{0.78}{0.508}&\scalebox{0.78}{0.621} &\scalebox{0.78}{0.383} &\scalebox{0.78}{0.618} &{\scalebox{0.78}{0.328}} &\scalebox{0.78}{0.622} &\scalebox{0.78}{0.337} \\ 
    
    & \scalebox{0.78}{720} & \boldres{\scalebox{0.78}{0.463}} & \boldres{\scalebox{0.78}{0.300}} & \secondres{\scalebox{0.78}{0.467}} & \secondres{\scalebox{0.78}{0.302}} & \scalebox{0.78}{0.647} & \scalebox{0.78}{0.387} & \scalebox{0.78}{0.586} & \scalebox{0.78}{0.375} & \scalebox{0.78}{0.589} & \scalebox{0.78}{0.328}  & \scalebox{0.78}{0.719} & \scalebox{0.78}{0.449} &\scalebox{0.78}{0.640} &{\scalebox{0.78}{0.350}} &\scalebox{0.78}{0.645} &\scalebox{0.78}{0.394} & \scalebox{0.78}{0.841} & \scalebox{0.78}{0.523} &{\scalebox{0.78}{0.626}} &\scalebox{0.78}{0.382} &\scalebox{0.78}{0.653} &{\scalebox{0.78}{0.355}} &\scalebox{0.78}{0.660} &\scalebox{0.78}{0.408} \\ 
    
    \cmidrule(lr){2-26}
    
    & \scalebox{0.78}{Avg} & \secondres{\scalebox{0.78}{0.431}} & \secondres{\scalebox{0.78}{0.287}} & \boldres{\scalebox{0.78}{0.428}} & \boldres{\scalebox{0.78}{0.282}} & \scalebox{0.78}{0.626} & \scalebox{0.78}{0.378} & \scalebox{0.78}{0.555} & \scalebox{0.78}{0.362} & \scalebox{0.78}{0.550} & \scalebox{0.78}{0.304} & \scalebox{0.78}{0.760} & \scalebox{0.78}{0.473} &{\scalebox{0.78}{0.620}} &{\scalebox{0.78}{0.336}} &\scalebox{0.78}{0.625} &\scalebox{0.78}{0.383} & \scalebox{0.78}{0.804} & \scalebox{0.78}{0.509} &{\scalebox{0.78}{0.610}} &\scalebox{0.78}{0.376} &\scalebox{0.78}{0.624} &{\scalebox{0.78}{0.340}} &\scalebox{0.78}{0.628} &\scalebox{0.78}{0.379} \\ 
    
    \midrule
    
    \multirow{5}{*}{\rotatebox{90}{\scalebox{0.95}{Weather}}} 
    &  \scalebox{0.78}{96} & \secondres{\scalebox{0.78}{0.163}} & \boldres{\scalebox{0.78}{0.207}} & \scalebox{0.78}{0.174} & \secondres{\scalebox{0.78}{0.214}} & \scalebox{0.78}{0.192} & \scalebox{0.78}{0.232} & \scalebox{0.78}{0.177} & \scalebox{0.78}{0.218} & \boldres{\scalebox{0.78}{0.158}} & \scalebox{0.78}{0.230}  & \scalebox{0.78}{0.202} & \scalebox{0.78}{0.261} &\scalebox{0.78}{0.172} &{\scalebox{0.78}{0.220}} & \scalebox{0.78}{0.196} &\scalebox{0.78}{0.255} & \scalebox{0.78}{0.221} & \scalebox{0.78}{0.306} & \scalebox{0.78}{0.217} &\scalebox{0.78}{0.296} & {\scalebox{0.78}{0.173}} &{\scalebox{0.78}{0.223}} & \scalebox{0.78}{0.266} &\scalebox{0.78}{0.336} \\ 
    
    & \scalebox{0.78}{192} & \secondres{\scalebox{0.78}{0.211}} & \boldres{\scalebox{0.78}{0.251}} & \scalebox{0.78}{0.221} & \secondres{\scalebox{0.78}{0.254}} & \scalebox{0.78}{0.240} & \scalebox{0.78}{0.271} & \scalebox{0.78}{0.225} & \scalebox{0.78}{0.259} & \boldres{\scalebox{0.78}{0.206}} & \scalebox{0.78}{0.277} & \scalebox{0.78}{0.242} & \scalebox{0.78}{0.298} &\scalebox{0.78}{0.219} &\scalebox{0.78}{0.261} & \scalebox{0.78}{0.237} &\scalebox{0.78}{0.296} & \scalebox{0.78}{0.261} & \scalebox{0.78}{0.340} & \scalebox{0.78}{0.276} &\scalebox{0.78}{0.336} & \scalebox{0.78}{0.245} &\scalebox{0.78}{0.285} & \scalebox{0.78}{0.307} &\scalebox{0.78}{0.367} \\ 
    
    & \scalebox{0.78}{336} & \boldres{\scalebox{0.78}{0.267}} & \boldres{\scalebox{0.78}{0.292}} & \scalebox{0.78}{0.278} & \secondres{\scalebox{0.78}{0.296}} & \scalebox{0.78}{0.292} & \scalebox{0.78}{0.307} & \scalebox{0.78}{0.278} & \scalebox{0.78}{0.297} & \secondres{\scalebox{0.78}{0.272}} & \scalebox{0.78}{0.335} & \scalebox{0.78}{0.287} & \scalebox{0.78}{0.335} &{\scalebox{0.78}{0.280}} &{\scalebox{0.78}{0.306}} & \scalebox{0.78}{0.283} &\scalebox{0.78}{0.335} & \scalebox{0.78}{0.309} & \scalebox{0.78}{0.378} & \scalebox{0.78}{0.339} &\scalebox{0.78}{0.380} & \scalebox{0.78}{0.321} &\scalebox{0.78}{0.338} & \scalebox{0.78}{0.359} &\scalebox{0.78}{0.395}\\ 
    
    & \scalebox{0.78}{720} & \boldres{\scalebox{0.78}{0.343}} & \boldres{\scalebox{0.78}{0.341}} & \scalebox{0.78}{0.358} & \secondres{\scalebox{0.78}{0.349}} & \scalebox{0.78}{0.364} & \scalebox{0.78}{0.353} & \scalebox{0.78}{0.354} & \scalebox{0.78}{0.348} & \scalebox{0.78}{0.398} & \scalebox{0.78}{0.418} & \secondres{\scalebox{0.78}{0.351}} & \scalebox{0.78}{0.386} &\scalebox{0.78}{0.365} &{\scalebox{0.78}{0.359}} & {\scalebox{0.78}{0.345}} &{\scalebox{0.78}{0.381}} & \scalebox{0.78}{0.377} & \scalebox{0.78}{0.427} & \scalebox{0.78}{0.403} &\scalebox{0.78}{0.428} & \scalebox{0.78}{0.414} &\scalebox{0.78}{0.410} & \scalebox{0.78}{0.419} &\scalebox{0.78}{0.428} \\ 
    
    \cmidrule(lr){2-26}
    
    & \scalebox{0.78}{Avg} & \boldres{\scalebox{0.78}{0.246}} & \boldres{\scalebox{0.78}{0.272}} & \secondres{\scalebox{0.78}{0.258}} & \secondres{\scalebox{0.78}{0.279}} & \scalebox{0.78}{0.272} & \scalebox{0.78}{0.291} & \scalebox{0.78}{0.259} & \scalebox{0.78}{0.281} & \scalebox{0.78}{0.259} & \scalebox{0.78}{0.315} & \scalebox{0.78}{0.271} & \scalebox{0.78}{0.320} &{\scalebox{0.78}{0.259}} &{\scalebox{0.78}{0.287}} &\scalebox{0.78}{0.265} &\scalebox{0.78}{0.317} & \scalebox{0.78}{0.292} & \scalebox{0.78}{0.363} &\scalebox{0.78}{0.309} &\scalebox{0.78}{0.360} &\scalebox{0.78}{0.288} &\scalebox{0.78}{0.314} &\scalebox{0.78}{0.338} &\scalebox{0.78}{0.382} \\ 
    
    \midrule
    
    \multirow{5}{*}{\rotatebox{90}{\scalebox{0.95}{Solar-Energy}}} 
    &  \scalebox{0.78}{96} & \boldres{\scalebox{0.78}{0.185}} & \boldres{\scalebox{0.78}{0.233}} &\secondres{\scalebox{0.78}{0.203}} & \secondres{\scalebox{0.78}{0.237}} & \scalebox{0.78}{0.322} & \scalebox{0.78}{0.339} & \scalebox{0.78}{0.234} & \scalebox{0.78}{0.286} &\scalebox{0.78}{0.310} &\scalebox{0.78}{0.331} &\scalebox{0.78}{0.312} &\scalebox{0.78}{0.399} &\scalebox{0.78}{0.250} &\scalebox{0.78}{0.292} &\scalebox{0.78}{0.290} &\scalebox{0.78}{0.378} &\scalebox{0.78}{0.237} &\scalebox{0.78}{0.344} &\scalebox{0.78}{0.242} &\scalebox{0.78}{0.342} &\scalebox{0.78}{0.215} &\scalebox{0.78}{0.249} &\scalebox{0.78}{0.884} &\scalebox{0.78}{0.711}\\ 
    
    & \scalebox{0.78}{192} & \boldres{\scalebox{0.78}{0.227}} & \boldres{\scalebox{0.78}{0.253}} &\secondres{\scalebox{0.78}{0.233}} &\secondres{\scalebox{0.78}{0.261}} & \scalebox{0.78}{0.359} & \scalebox{0.78}{0.356}& \scalebox{0.78}{0.267} & \scalebox{0.78}{0.310} &\scalebox{0.78}{0.734} &\scalebox{0.78}{0.725} &\scalebox{0.78}{0.339} &\scalebox{0.78}{0.416} &\scalebox{0.78}{0.296} &\scalebox{0.78}{0.318} &\scalebox{0.78}{0.320} &\scalebox{0.78}{0.398} &\scalebox{0.78}{0.280} &\scalebox{0.78}{0.380} &\scalebox{0.78}{0.285} &\scalebox{0.78}{0.380} &\scalebox{0.78}{0.254} &\scalebox{0.78}{0.272} &\scalebox{0.78}{0.834} &\scalebox{0.78}{0.692} \\ 
    
    & \scalebox{0.78}{336} & \boldres{\scalebox{0.78}{0.246}} & \secondres{\scalebox{0.78}{0.284}} &\secondres{\scalebox{0.78}{0.248}} &\boldres{\scalebox{0.78}{0.273}} & \scalebox{0.78}{0.397} & \scalebox{0.78}{0.369}& \scalebox{0.78}{0.290} & \scalebox{0.78}{0.315} &\scalebox{0.78}{0.750} &\scalebox{0.78}{0.735} &\scalebox{0.78}{0.368} &\scalebox{0.78}{0.430} &\scalebox{0.78}{0.319} &\scalebox{0.78}{0.330} &\scalebox{0.78}{0.353} &\scalebox{0.78}{0.415} &\scalebox{0.78}{0.304} &\scalebox{0.78}{0.389} &\scalebox{0.78}{0.282} &\scalebox{0.78}{0.376} &\scalebox{0.78}{0.290} &\scalebox{0.78}{0.296} &\scalebox{0.78}{0.941} &\scalebox{0.78}{0.723} \\ 
    
    & \scalebox{0.78}{720} & \boldres{\scalebox{0.78}{0.247}} & \secondres{\scalebox{0.78}{0.276}} &\secondres{\scalebox{0.78}{0.249}} &\boldres{\scalebox{0.78}{0.275}} & \scalebox{0.78}{0.397} & \scalebox{0.78}{0.356} &\scalebox{0.78}{0.289} &\scalebox{0.78}{0.317} &\scalebox{0.78}{0.769} &\scalebox{0.78}{0.765} &\scalebox{0.78}{0.370} &\scalebox{0.78}{0.425} &\scalebox{0.78}{0.338} &\scalebox{0.78}{0.337} &\scalebox{0.78}{0.356} &\scalebox{0.78}{0.413} &\scalebox{0.78}{0.308} &\scalebox{0.78}{0.388} &\scalebox{0.78}{0.357} &\scalebox{0.78}{0.427} &\scalebox{0.78}{0.285} &\scalebox{0.78}{0.295} &\scalebox{0.78}{0.882} &\scalebox{0.78}{0.717} \\ 
    
    \cmidrule(lr){2-26}
    
    & \scalebox{0.78}{Avg} & \boldres{\scalebox{0.78}{0.226}} & \boldres{\scalebox{0.78}{0.261}} &\secondres{\scalebox{0.78}{0.233}} &\secondres{\scalebox{0.78}{0.262}} & \scalebox{0.78}{0.369} & \scalebox{0.78}{0.356} &\scalebox{0.78}{0.270} &\scalebox{0.78}{0.307} &\scalebox{0.78}{0.641} &\scalebox{0.78}{0.639} &\scalebox{0.78}{0.347} &\scalebox{0.78}{0.417} &\scalebox{0.78}{0.301} &\scalebox{0.78}{0.319} &\scalebox{0.78}{0.330} &\scalebox{0.78}{0.401} &\scalebox{0.78}{0.282} &\scalebox{0.78}{0.375} &\scalebox{0.78}{0.291} &\scalebox{0.78}{0.381} &\scalebox{0.78}{0.261} &\scalebox{0.78}{0.381} &\scalebox{0.78}{0.885} &\scalebox{0.78}{0.711} \\ 
    
    \midrule
    
     \multicolumn{2}{c|}{\scalebox{0.78}{{$1^{\text{st}}$ Count}}} & \scalebox{0.78}{\boldres{29}} & \scalebox{0.78}{\boldres{31}} & \scalebox{0.78}{\secondres{4}} & \scalebox{0.78}{\secondres{8}} & \scalebox{0.78}{1} & \scalebox{0.78}{1} & \scalebox{0.78}{2} & \scalebox{0.78}{1} & \scalebox{0.78}{2} & \scalebox{0.78}{0} & \scalebox{0.78}{0} & \scalebox{0.78}{0} & \scalebox{0.78}{0} & \scalebox{0.78}{0} & \scalebox{0.78}{0} & \scalebox{0.78}{0} & \scalebox{0.78}{0} & \scalebox{0.78}{0} & \scalebox{0.78}{2} & \scalebox{0.78}{0} & \scalebox{0.78}{0} & \scalebox{0.78}{0} & \scalebox{0.78}{0} & \scalebox{0.78}{0} \\ 
    \bottomrule
  \end{tabular}
\end{table*}

\end{document}

%% file: math_commands.tex
\usepackage{amsmath, amsfonts, bm}

\def\eqref#1{equation~\ref{#1}}

\def\1{\bm{1}}

\def\va{{\bm{a}}}

\def\vw{{\bm{w}}}
\def\vx{{\bm{x}}}

\DeclareMathAlphabet{\mathsfit}{\encodingdefault}{\sfdefault}{m}{sl}
\SetMathAlphabet{\mathsfit}{bold}{\encodingdefault}{\sfdefault}{bx}{n}

